\pdfoutput=1
\documentclass[11pt]{article}

\usepackage[preprint]{ARR}

\usepackage{times}
\usepackage{latexsym}

\usepackage[T1]{fontenc}
\usepackage[utf8]{inputenc}

\usepackage{microtype}
\usepackage{amsmath}

\usepackage{inconsolata}

\usepackage{graphicx}
\usepackage{subcaption}

\title{Small Changes, Large Consequences:\\Analyzing the Allocational Fairness of LLMs in Hiring Contexts}

\author{Preethi Seshadri \\
  UC Irvine \\
  \texttt{preethis@uci.edu} \\\And
  Hongyu Chen \\
  Cohere \\
  \texttt{charlie@cohere.com}
  \\\And
  Sameer Singh \\
  UC Irvine \\
  \texttt{sameer@uci.edu} 
  \\\And
  Seraphina Goldfarb-Tarrant \\
  Cohere \\
\texttt{seraphina@cohere.com} 
  \\}

\begin{document}
\maketitle

\begin{abstract}
Large language models (LLMs) are increasingly being deployed in high-stakes applications like hiring, yet their potential for unfair decision-making  remains understudied in generative and retrieval settings. 
In this work, we examine the allocational fairness of LLM-based hiring systems through two tasks that reflect actual HR usage: resume summarization and applicant ranking. 
By constructing a synthetic resume dataset with controlled perturbations and curating job postings, we investigate whether model behavior differs across demographic groups. 
Our findings reveal that generated summaries exhibit meaningful differences more frequently for race than for gender perturbations.
Models also display non-uniform retrieval selection patterns across demographic groups and exhibit high ranking sensitivity to both gender and race perturbations. 
Surprisingly, retrieval models can show comparable sensitivity to both demographic and non-demographic changes, suggesting that fairness issues may stem from broader model brittleness. 
Overall, our results indicate that LLM-based hiring systems, especially in the retrieval stage, can exhibit notable biases that lead to discriminatory outcomes in real-world contexts.
\end{abstract}

\section{Introduction}
\label{introduction}

Large language models (LLMs) are increasingly being adopted in real-world, high-stakes domains such as hiring \citep{bcg2025_ai_recruitment}, where they assist HR teams with tasks like resume screening and candidate matching.
As LLMs are incorporated into critical decision-making processes, ensuring fair and responsible deployment is essential, especially when the outcomes can profoundly impact individuals' career prospects  \citep{dastin2018amazon, raghavan2020, sanchez-monedero20, suresh2021}.

A key aspect of developing responsible LLM systems includes anticipating and preventing specific risks and harms, such as \textit{allocational harms} (i.e., allocating resources or opportunities unfairly to different social groups, also called \textit{allocational fairness}) \citep{barocas2017problem, blodgett-etal-2020-language}. 
This is especially important in automated hiring pipelines, since models may produce unfair outcomes and reinforce systemic inequalities.
While there is a substantial body of work that analyzes representational harms (i.e., representing certain social groups negatively, demeaning them, or erasing their existence) in LLMs \citep{zhao-etal-2018-gender, Abid2021LargeLM, kirk2021, cheng-etal-2023-marked, gadiraju2023}, allocational harms---which are the primary harm at play in high-stakes situations---remain understudied beyond discriminative systems.

The few studies that evaluate allocational harms of LLMs \citep{tamkin2023evaluatingmitigatingdiscriminationlanguage, an-etal-2024-large, haim2024whatsnameauditinglarge, nghiem-etal-2024-gotta} have primarily cast their investigations as discrete classification tasks (e.g., yes/no decisions) or quantitative predictions (e.g., determining salary levels), which do not capture how LLMs are deployed in applications like hiring \citep{kelly2023ai}. 
As a result, these highly simplified setups may inadequately predict real-world outcomes and assess harms.
Investigations of LLM harms must ensure \textit{ecological validity} \citep{blodgett-etal-2021-stereotyping, goldfarb-tarrant-etal-2021-intrinsic, cao-etal-2022-intrinsic}; they should be grounded in realistic scenarios and tasks that match how these systems are used in practice, or use a proxy that is predictive of real world outcomes. 
Yet there is limited work on allocational harms in generative settings without adding a simplification layer, with \citet{wan-etal-2023-kelly} being a notable exception, since measuring how generated text might create disparities is more open-ended and complex than analyzing classification predictions.

\begin{figure}[t]
    \centering
    \includegraphics[width=0.95\columnwidth]{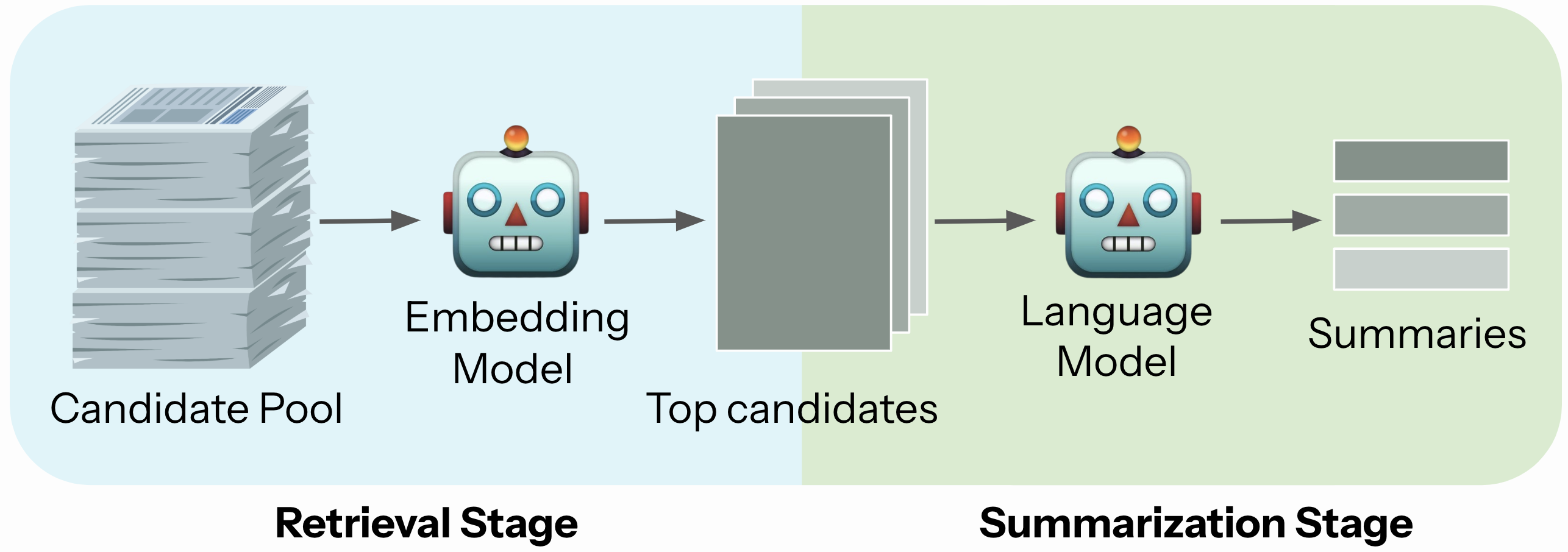}
    \caption{We investigate the fairness of an LLM hiring pipeline with a \textbf{retrieval stage} (ranks/filters the top-$n$ candidates with respect to a job post) and a \textbf{summarization stage} (generates resume summaries for filtered candidates). We assess fairness at each stage separately.\protect\footnotemark[1]}
    \label{fig:pipeline}
    \vspace{-0.4em}
\end{figure}

In this work, we examine whether LLMs behave fairly in real-world hiring contexts. 
We focus on two critical tasks that mirror how LLMs are integrated into hiring workflows \citep{herman2024ways, humanly2024using}:  (1) ranking candidates with respect to a job posting and (2) summarizing resumes, as illustrated in Figure \ref{fig:pipeline}. 
These tasks represent key stages where automation can influence which candidates are surfaced and considered for a role.
To evaluate fairness, we examine whether models are sensitive to gender and race perturbations in resumes.
We investigate the following questions:
\begin{itemize}
\item \textbf{RQ1:} Do generated summaries differ meaningfully across demographic groups?\footnotemark
\vspace{-0.4em}
\item \textbf{RQ2:} Do models disparately select resumes across demographic groups?
\vspace{-0.4em}
\item \textbf{RQ3:} How sensitive are model rankings to demographic and non-demographic perturbations in resumes?
\end{itemize}

\footnotetext{\label{fn:ordering}We study summarization first, since it is less explored from an allocational harms perspective.}

To this end, we: 
(1) construct a new benchmark consisting of a synthetic resume dataset with controlled demographic perturbations (varying names and extracurricular content) and curated job postings,
(2) design an evaluation framework with fairness metrics tailored to both generative and retrieval settings, validated by an expert human preference study,
(3) conduct a comprehensive fairness analysis of 10 large language models (6 generative, 4 retrieval) based on real-world hiring tasks.
We will make all data and code publicly available.

Our results demonstrate that an LLM hiring system with automated resume retrieval and summarization exhibits considerable bias, primarily stemming from the retrieval stage.
In the summarization setting, we observe meaningful differences in generated summaries up to 20\% of the time between racial groups, compared to 3\% for gender (\textbf{RQ1}).
For retrieval, models non-uniformly select resumes across demographic groups up to 55\% of the time (\textbf{RQ2}), and produce rankings that are highly sensitive to gender and race, with up to 74\% of candidates being filtered out after demographic perturbation (\textbf{RQ3}). 
We also find that models exhibit high sensitivity to non-demographic changes, sometimes on par with demographic changes (\textbf{RQ3}), suggesting that fairness issues can stem from general model brittleness rather than demographic bias alone.
Overall, our analysis reveals that even seemingly minor changes can lead to considerable disparities, raising concerns about the fairness and robustness of LLMs in hiring.

\section{Methodology}
To study fairness in hiring, we consider an LLM-based pipeline with two components: resume retrieval with respect to a job post (using an embedding model) and resume summarization (using an LLM).
This pipeline is informed by interviews with several corporations that actively deploy LLMs for hiring;\footnote{We cannot share details due to non-disclosure agreements.} both components reflect real-world usage of automation to streamline hiring processes. 
We focus on summarization first because it is more neglected in research, though in a pipeline it would come after retrieval, as shown in Figure \ref{fig:pipeline} (as summarization would be of retrieved resumes).

We propose two metrics, invariance violations (summarization) and exclusion (retrieval), to investigate allocational fairness in hiring.
% for both generative and retrieval settings. 
Specifically, these metrics quantify: (1) systematic differences in generated resume summaries\footnote{Since recruiters may rely on summaries rather than full resumes, meaningful differences in summaries could directly affect hiring outcomes, leading to allocational harms.} 
and (2) changes in the similarity of a resume to a job posting, and as a result its ranking in a resume set. We additionally benchmark the distribution metric of \citet{wilson2024intersectionalbias} to study fairness in hiring, but note that their approach does not directly capture how perturbing a resume impacts resume screening outcomes, and includes only retrieval and not generative settings.

Let $D$ represent a set of resumes, where each resume  $d \in D$ has demographic label $l(d)$.
We denote $d'$ as any perturbed version of $d$ where $l(d') \neq l(d)$.
For each $d$, there can be several demographically perturbed versions $d'$ (e.g., gender or race perturbations). 
The resume content remains largely unchanged except for the demographic perturbation.

\subsection{Summarization}

Perturbed resumes are by design highly similar to original resumes, so we expect generated summaries for original and perturbed resumes to also maintain high similarity.
In other words, we are testing for \textit{invariance}; we expect the output to change minimally after perturbing the input \citep{ribeiro-etal-2020-beyond}.
To assess invariance, we need a way to measure whether the original and perturbed summaries differ meaningfully in the context of hiring. 
% one could collect human preferences to see whether generated summaries from a specific group are systematically preferred. 
For cost and scalability reasons, we rely on an automated approach---human preferences are expensive and cannot be collected quickly enough to be used in model development, especially if these preferences are sourced from HR staff.
%Additionally, collecting human preferences for a specific dataset does not extend to other analyses with new datasets. 
% However, this approach requires human annotations for each summary, and does not extend to other analyses with new summaries.

Automated measures can evaluate whether specific properties of generated summaries differ, in a way that could affect a human reader's opinion. %where differences in these properties would be undesirable in practice.
For instance, summaries for a specific demographic group should not be written more positively than summaries for other demographic groups; this could lead to disparities in hiring outcomes.
% We opt for the latter approach of using automated measures for evaluation due to its scalability. Furthermore, this approach enables us to test whether generated summaries differ along targeted dimensions.
We use the following measures as proxies for undesirable variation that could influence the decision of HR staff reading the summary: reading ease, reading time, polarity, subjectivity, and regard (Appendix \ref{appendix:proxy_measures}).
We verify that these measures capture meaningful differences in practice by conducting a preference task annotated by HR staff. 
This study shows that these are good proxies for human preferences (Appendix \ref{appendix:human_preferences}).

\paragraph{Fairness Metric} 
% We compute an invariance metric that utilizes the proposed measures. 
For each measure, we perform a paired t-test between scores for original and perturbed summaries. 
We then calculate how often the null hypothesis (that the mean difference between paired summaries is 0) is rejected, i.e., how often invariance is violated.
We choose a significance level of $\alpha=0.05$, and apply Benjamini-Hochberg correction to account for Type 1 errors with multiple comparisons 
\citep{benjamini1995controlling}.
\vspace{0.1em}
\[
\text{\shortstack{invariance \\ violations}} = \frac{\shortstack{\textit{\# t-tests for which null hypothesis} \\ \textit{is rejected}}}{\textit{total \# of t-tests}}
\]

\subsection{Retrieval}
In contrast to summarization, which relies only on the resume, retrieval uses both a query (job posting) and candidates to select (resumes). 
The top-$n$ of the resulting resumes sorted by similarity then make it to the next stage of the pipeline.
%By retrieving resumes that are most relevant to a job posting and ranking them according to their relevance, we simulate automated resume screening. 
We assume that demographically perturbing a resume should have minimal impact on its relevance to a job posting. % which we formalize below

Given an embedding model \( \mathcal{M} \) and a similarity measure $sim$, we compute the similarity between embeddings for resume $d$ and job posting $p$. In practice, we use cosine similarity following \citet{wilson2024intersectionalbias}.
We define the set $S(p)$, which represents the set of similarity values between each resume \( d \in D \) and a given job posting \( p \): $S(p) = \{ sim(\mathcal{M}(d), \mathcal{M}(p)) \mid d \in D \}$.
% \[
% S(p) = \{ sim(\mathcal{M}(d), \mathcal{M}(p)) \mid d \in D \}
% \]

We transform the similarity values between each resume and job posting, $s_i$, into a rank such that lower ranks indicate higher similarity: $\text{rank}_p(s_i) = |\{ s_j \in S(p) \mid s_j > s_i \}| + 1$. Let $D_n(p)$ represent the set of top-$n$ resumes from $D$, which are the resumes with the $n$ lowest ranks (i.e., highest similarities) for job posting $p$: $ D_n(p) = \{ d_i \in D \mid \text{rank}(s_i) \leq n \}$.
% \vspace{-0.1em}
% \[
% \text{rank}_p(s_i) = |\{ s_j \in S(p) \mid s_j > s_i \}| + 1
% \]

% \[
% D_n(p) = \{ d_i \in D \mid \text{rank}(s_i) \leq n \}
% \]

\paragraph{Fairness Metrics} 
We compute two fairness metrics for retrieval, \textit{non-uniformity}, proposed by \citet{wilson2024intersectionalbias}, and \textit{exclusion}, which we introduce below.\footnote{Further intuition and differentiation of these metrics are provided in Appendix \ref{appendix:intuition_retrieval}.}
\vspace{-0.3em}

\paragraph{Non-uniformity} assesses whether the top resumes are uniformly distributed across demographic groups. 
First, the set of top-$x$\% of resumes ($x$ being a percentage rather than a fixed number $n$) is retrieved from the combined pool of all demographically perturbed versions, denoted as $D_x'(p)$. 
A chi-squared goodness-of-fit test is then used to check if the demographic composition of $D_x'(p)$ deviates from the uniform distribution.
% \paragraph{Non-uniformity} evaluates whether the top-$x$\% of resumes ($x$ is a percentage instead of a fixed number $n$) are uniformly distributed with respect to demographic groups. 
% To compute this, we consider the set of top-$x$\% of resumes retrieved from the combined pool of all demographically perturbed versions, which we denote as $D_x'(p)$.
% Following the approach from \citet{wilson2024intersectionalbias}, we perform a chi-squared goodness-of-fit test to determine whether the demographic makeup of $D_x'(p)$ deviates from the uniform distribution.

\paragraph{Exclusion} evaluates how often resumes in the set of top-$n$ resumes are excluded (i.e., the ranking falls outside top-$n$) after perturbation. 
Ideally, \( \mathcal{M} \) should be robust to demographic perturbations, yielding nearly identical similarity scores and rankings for both resume $d$ and demographically perturbed version $d'$. Exclusion directly assesses allocational fairness by measuring how much the set of top-$n$ resumes differs after perturbation.

\vspace{-0.3em}
\[
\text{exclusion}_n(p) = \frac{|{\text{rank}_p(d') > n \mid d \in D_n(p)}|}{|D_n(p)|}
\]

\section{Experimental Setup}
\label{sec:setup}
In this section, we describe the data, perturbations, and models used for our evaluation.
 \vspace{-0.3em}
 
\subsection{Data}
\paragraph{Resumes} Resumes are sourced through social media platforms (LinkedIn, Slack, X). 
Given the authors' professional circles, the sample skews heavily toward tech and academic professionals.
For privacy reasons, we anonymize resumes using Presidio to mask PII entities \citep{microsoft_presidio_anonymizer}.
To further mitigate privacy concerns and enable dataset release, we use the collected resumes as examples to generate synthetic resumes. % instead of using original resumes in our analysis.
We generate 525 resumes across 22 professions using Cohere's Command-R model \citep{cohere2024command}. 
All synthetic resumes are free of explicit demographic information, until added during experimentation. 

In addition, we use a publicly available resume dataset from Kaggle \citep{bhawal2021resume} to increase coverage and generalization. 
These resumes differ in two notable ways: (1) they are less structured and formatted than generated ones, and (2) they include a more diverse set of fields (e.g., construction, fitness, etc.). 
We sample 1175 Kaggle resumes across 24 fields.
More details about dataset curation and statistics are provided in Appendix \ref{appendix:dataset-stats} - \ref{appendix:prompts}.

\paragraph{Job Posts}
Our resume dataset consists of two types: synthetic resumes generated for specific roles (e.g., data analyst) and actual resumes labeled with broader field categories (e.g., construction). 
For each profession/field (we choose 11 each from generated and Kaggle resumes), we carefully select 7 detailed LinkedIn job postings, resulting in 154 job postings. 

\subsection{Demographic Perturbations} 
We use names as a proxy for gender and racial information.
All resumes are initially free of names; we add them using the curated set from \citet{yin24-recruiting-bias}.\footnote{Uses voter registration data from North Carolina to identify demographically-distinct names.}
We consider four demographic groups, each with 100 unique names: Black female (FB), White female (FW), Black male (MB), and White male (MW).
Following \citet{wilson2024intersectionalbias}, we only vary the first name and fix ``Williams'' as the last name for all groups.

In actual resumes, demographic information can be encoded in more than just names. 
Therefore %, for synthetically generated resumes, 
we perform an additional augmentation step that adds extracurricular information using Command-R\footnote{Awards, clubs and leadership, and mentorship and volunteering experiences that are reflective of the individual's background and identity (see Appendix \ref{appendix:prompts} and \ref{appendix:extra-example}).} to the resumes (similar to \citet{glazko2024})
Adding this information can reinforce demographic signal by providing both explicit and implicit cues.

\subsection{Non-Demographic Perturbations}
We conduct two non-demographic perturbation experiments for retrieval to assess the baseline sensitivity of embedding models that is not due to demographics.

\paragraph{Within-Group Name Perturbations}
As a baseline comparison to performing name perturbations between different demographic groups (e.g., White female $\rightarrow$ Black Female), we assess whether models are sensitive to within-group demographic perturbations (e.g., White female $\rightarrow$ White Female).
By doing so, we disentangle how much bias is due to demographics vs. model sensitivity to name changes.
To control for the effects of frequency \citep{ethayarajh-etal-2019-understanding}, we bin names in each demographic group according to their frequency in the Pile dataset \citep{pile}, and match names based on the bin.\footnote{We use the What's in My Big Data tool \citep{elazar2023s} to obtain frequencies.}

\paragraph{Non-Name Perturbations}
We assess whether model rankings are sensitive to non-name perturbations. 
This allows us to examine whether models lack robustness more broadly. We test two perturbation types: (1) random character swapping, which does not impact readability or comprehension in a meaningful way\footnote{We choose 10 random characters in the resume and swap with neighboring keys to simulate typos.} and (2) replacing new lines in the resume with a single space instead, which targets formatting without modifying content.

\begin{figure*}[htbp]
    \centering
    % Set the width of the entire figure to \textwidth
    \begin{subfigure}[b]{0.43\textwidth}
        \centering
        \includegraphics[width=\textwidth]{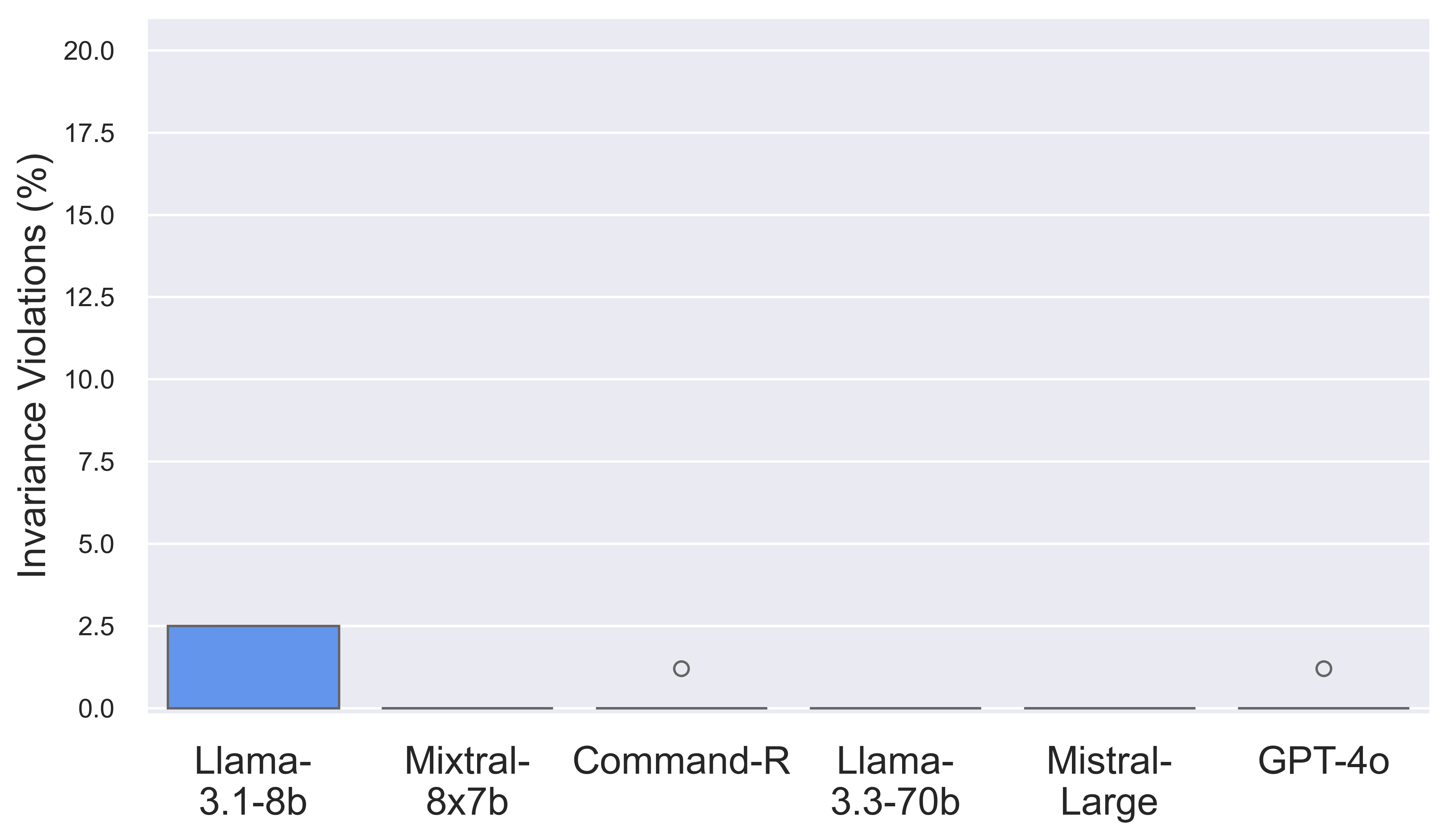}
        \caption{Gender}
        \label{fig:invariance_gender}
    \end{subfigure}
    % \hfill  % Add horizontal spacing between subfigures
    \hspace{0.05\textwidth}
    \begin{subfigure}[b]{0.43\textwidth}
        \centering
        \includegraphics[width=\textwidth]{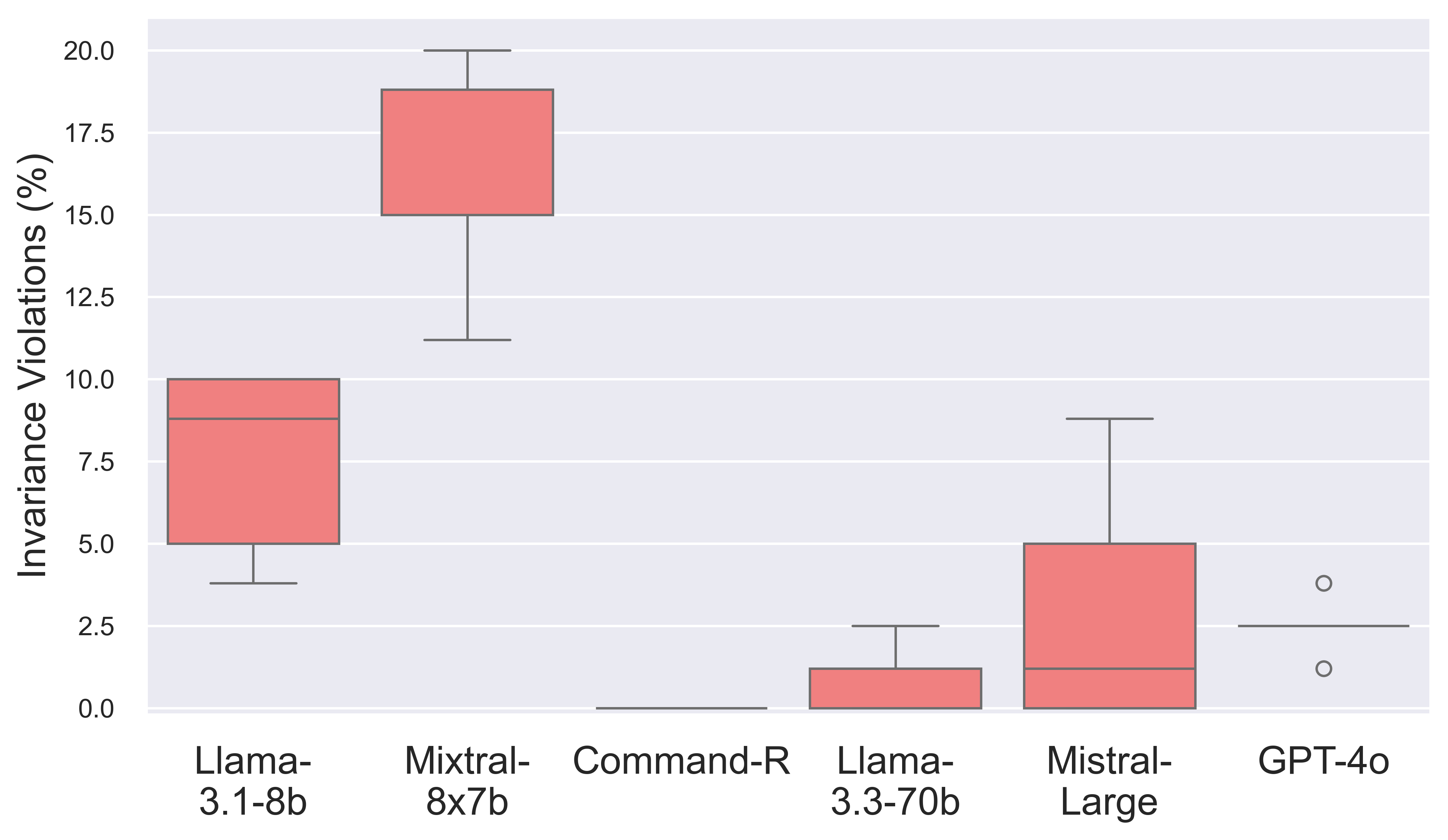}
        \caption{Race}
        \label{fig:invariance_race}
    \end{subfigure}
 
    \caption{\textbf{Summarization Results:} Invariance violations for generated summaries, separated by completion model and perturbation type. Results are shown across 5 runs. Left 3 models are considered "smaller" models, right 3 models are considered "larger" models.}
   \vspace{-0.5em}
\label{fig:inv_violations}
\end{figure*}

\subsection{Models}
\paragraph{Summarization} We generate summaries using closed and open state-of-the-art LLMs: GPT-4o \citep{openai2024gpt4}, Command-R \citep{cohere2024command}, Mixtral 8x7B and Mistral Large  \citep{jiang2024mixtralexperts}, and Llama 3.1 8B and Llama 3.3 70B \citep{meta2024llama3}. 
For summary instructions, we vary the generation length (100, 200 words) as well as the point of view (first, third person) specified in the prompt. 
We generate summaries with temperatures of 0.0 and 0.3.
To account for stochasticity in generations, we generate each summary five times.

\paragraph{Retrieval} For retrieval, we select four popular dense embedding models used in retrieval augmented generation (RAG) systems \citep{lewis2020}: OpenAI's text-embedding-3-small and text-embedding-3-large, Cohere's embed-english-v3.0, and Mistral's mistral-embed.

\section{Results}
In this section, we evaluate the use of LLMs in two real-world hiring tasks: resume summarization and retrieval. 
Unless otherwise mentioned, we present results for generated resumes below, and include results for Kaggle resumes in the Appendix.
Similar trends and findings hold for both datasets.

\subsection{Summarization}
We analyze whether generated summaries differ meaningfully when applying gender and race perturbations (\textbf{RQ1}) by examining invariance violations, i.e., the percentage of t-tests that yield significant differences in our automated measures. 
We measure violations separately for summaries with different characteristics (length, point of view, and temperature).
Figure \ref{fig:inv_violations} displays results grouped by completion model and perturbation type.

All models violate invariance much more for resumes that differ by race as opposed to gender.
In fact, gender invariance violations are zero or near zero for all models.
In contrast, all models except Command-R exhibit invariance violations with respect to race, with Mixtral 8x7B exhibiting violations 16.76\% of the time on average. 
Our results also provide some indication that smaller models are more susceptible to violations.
In summary, we observe that models exhibit some but not considerable discrepancies between generated summaries for different demographic groups, with minimal differences for gender perturbations. 

\subsection{Retrieval}
Moving to retrieval, we ask: Do models exhibit fairness issues in selecting resumes?
Our analysis tackles this question from distributional (\textbf{RQ2}) and robustness (\textbf{RQ3}) perspectives. 

\subsubsection{Non-Uniformity}
\begin{figure*}[htbp]
    \centering
    \begin{minipage}{0.38\textwidth}
        \centering
        \includegraphics[width=\linewidth]{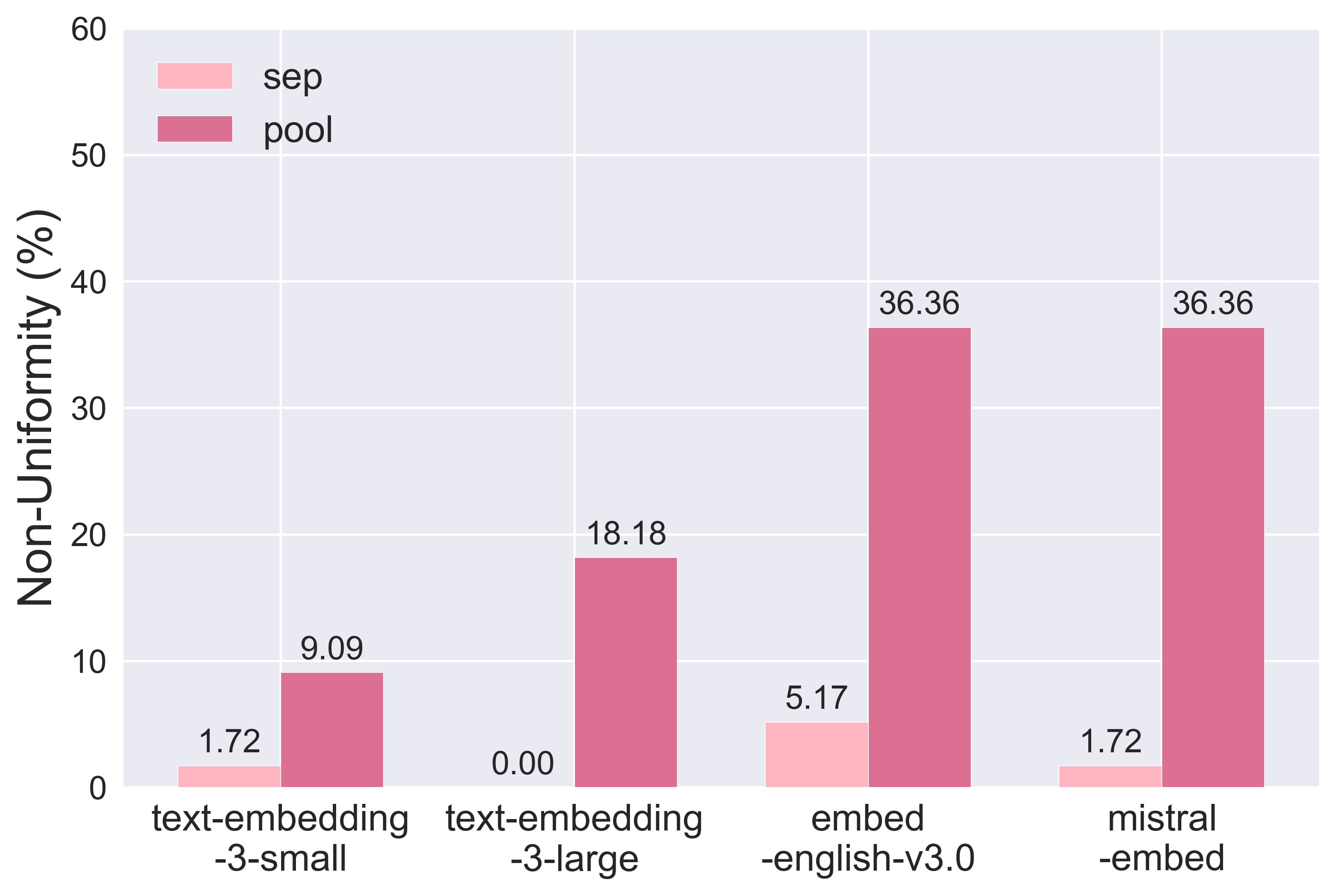}
        \subcaption{Top-5\%}
        \label{fig:nonuniform_plot1}
    \end{minipage}
    \hspace{0.05\textwidth}
    \begin{minipage}{0.38\textwidth}
        \centering
        \includegraphics[width=\linewidth]{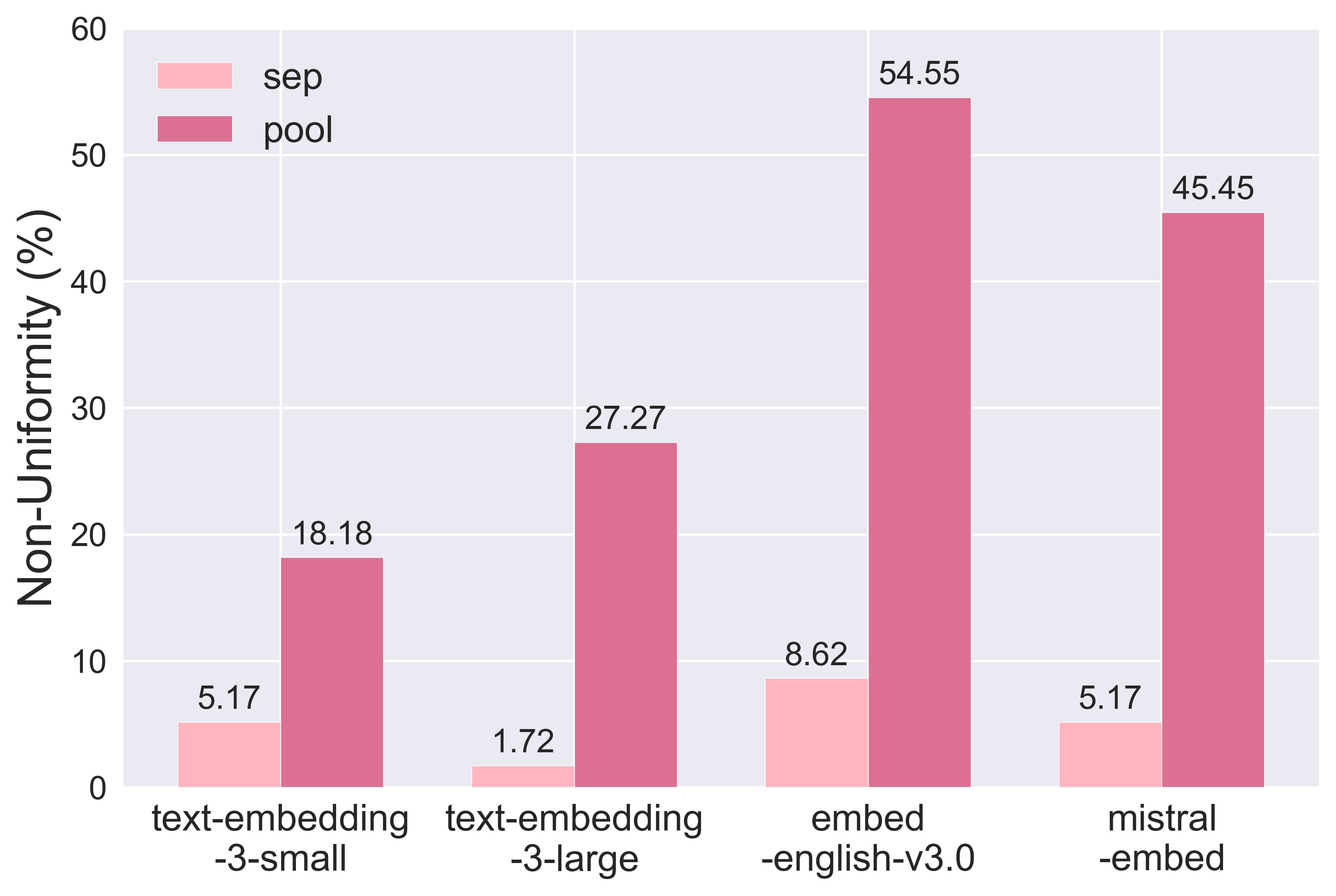}
        \subcaption{Top-10\%}
        \label{fig:nonuniform_plot2}
    \end{minipage}
    \caption{Non-uniformity metric for top-5 and top-10\% of retrieved resumes. Separated (sep) measures non-uniformity at a job post level, while pooled (pool) measures it at an occupation level by pooling results across job posts for a given occupation.}
    \label{fig:nonuniformity}
    \vspace{-0.5em}
\end{figure*}

\paragraph{Do models disparately select resumes across demographic groups?} 
To answer this question, we compute \textit{non-uniformity} (i.e., how often top retrieved resumes have non-uniform demographic distributions). 
All models disparately retrieve resumes across demographic groups, consistent with the findings of \citet{wilson2024intersectionalbias}.
That being said, non-uniformity differs considerably across models, choice of top-$x$ percent, and pooling of resumes across occupations (Figure \ref{fig:nonuniformity}).

We observe that embed-english-v3.0 exhibits the highest non-uniformity on average, with 6.90\% of job posts and 45.45\% of occupations having non-uniformly distributed resumes.
Increasing top-$x$ from 5\% to 10\% and pooling resumes both yield higher non-uniformity across all models. 
In particular, pooling resumes by occupation can produce massive changes; on average across models, non-uniformity goes from 3.66\% $\rightarrow$ 30.68\%.
This reflects sensitivity in the metric itself more than a change in the shape of the distribution.\footnote{Increasing top-$x$ and pooling both increase sample size, which can lead to rejecting the null hypothesis in cases where the null hypothesis previously failed to be rejected.}

\begin{figure*}[htbp]
    \centering
    \begin{subfigure}[b]{0.31\textwidth}
        \centering
        \includegraphics[width=\linewidth]{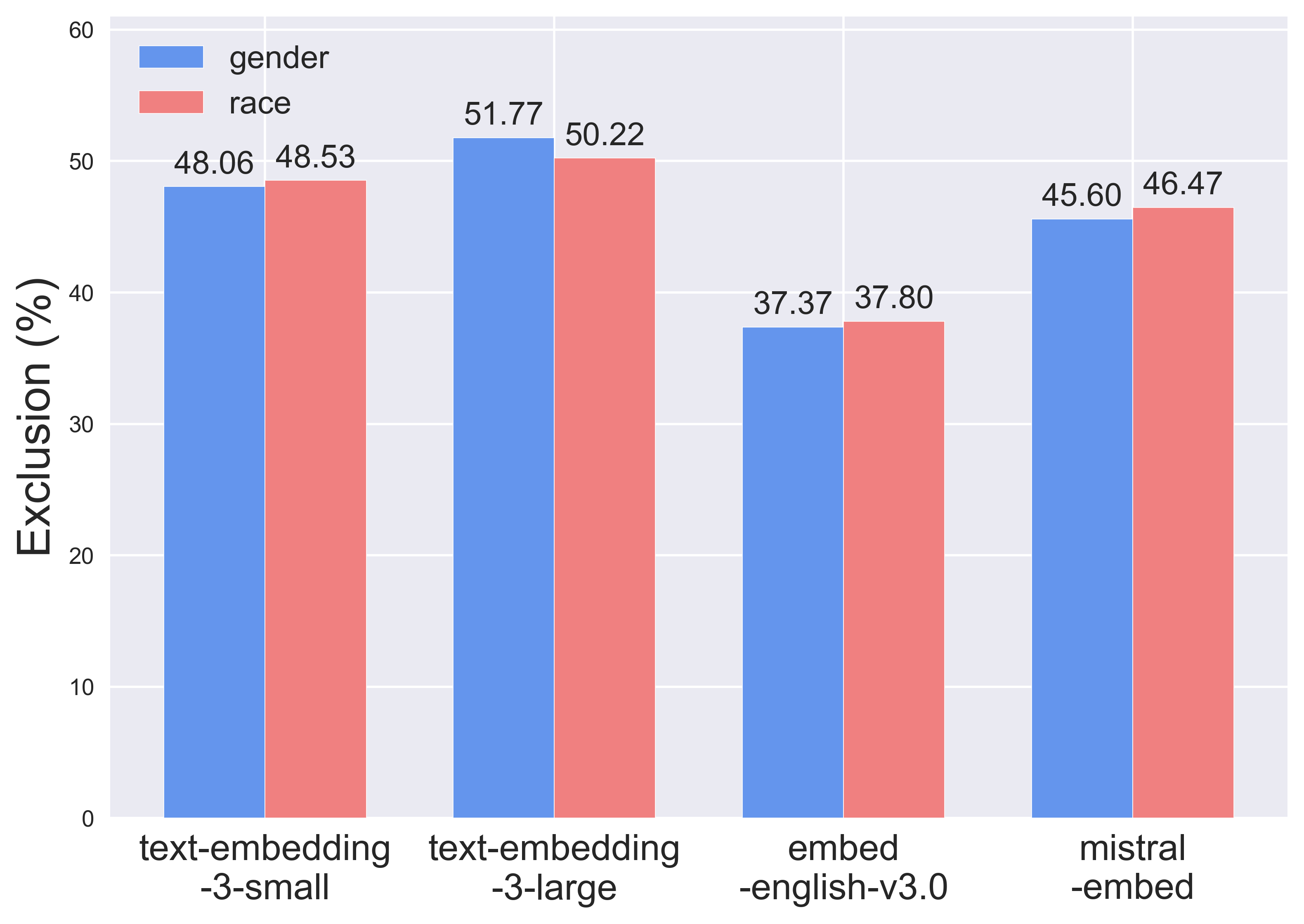}
        \caption{$n=5$}
        \label{fig:exclusion_plot1}
    \end{subfigure}
    \hspace{0.02\textwidth}
    \begin{subfigure}[b]{0.31\textwidth}
        \centering
        \includegraphics[width=\linewidth]{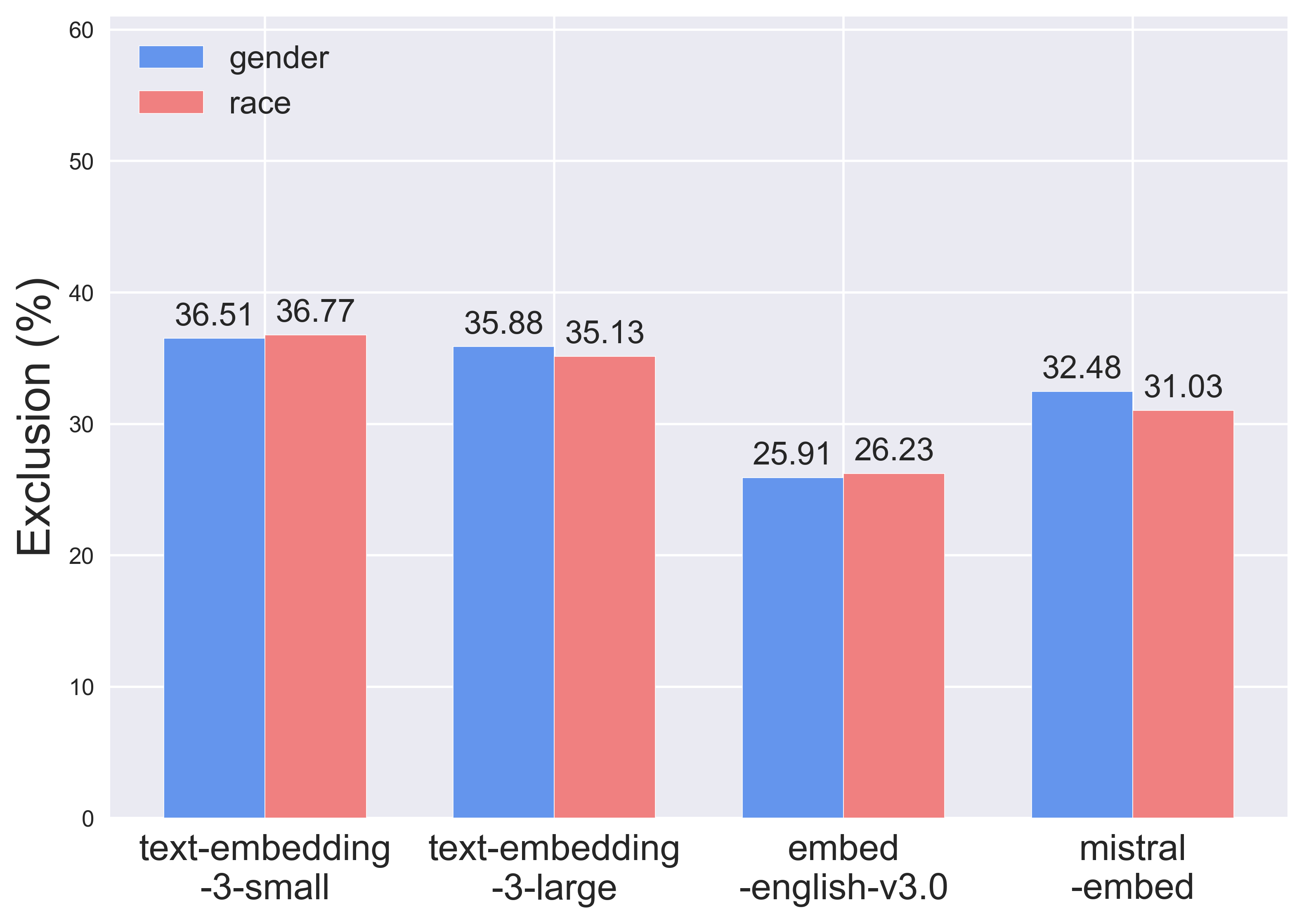}
        \caption{$n=10$}
        \label{fig:exclusion_plot2}
    \end{subfigure}
    \hspace{0.02\textwidth}
    \begin{subfigure}[b]{0.31 \textwidth}
        \centering
        \includegraphics[width=\linewidth]{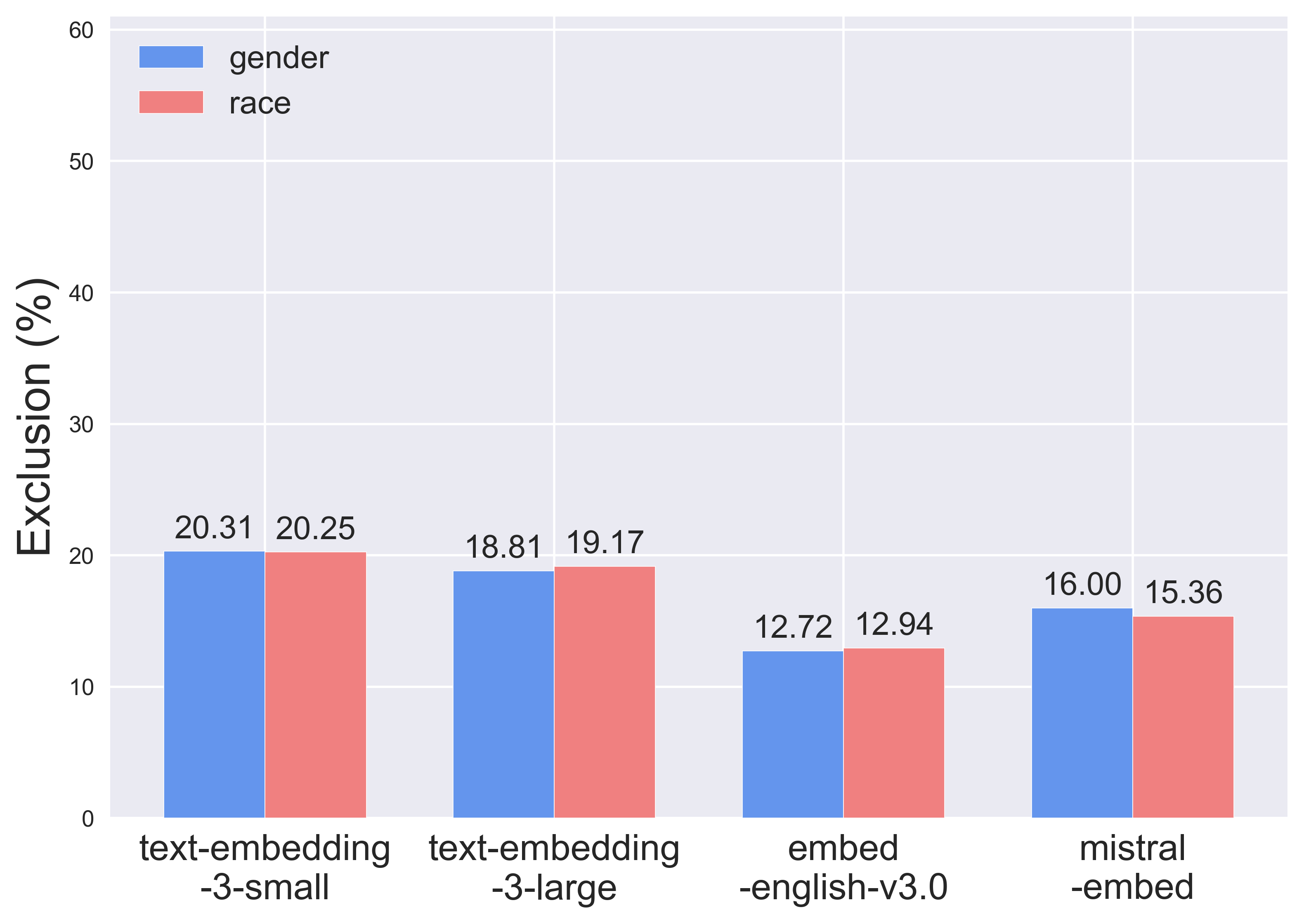}
        \caption{$n=100$}
        \label{fig:exclusion_plot3}
    \end{subfigure}
    \caption{\textbf{Exclusion metric for retrieval} after performing gender and race name perturbations for the top-5, top-10, and top-100 retrieved resumes. Lower values indicate models are less sensitive to demographic perturbations.}
    \label{fig:exclusion-initial}
\end{figure*}

Different models show distinct patterns of bias: the non-uniformity privileges different demographic groups.
For example, in the top-10\% of resumes from embed-english-v3.0, White females are the top group 48.05\% of the time, compared to 3.90\% for Black males.
In contrast, for mistral-embed, White males are the top group 72.73\% of the time, compared to 5.19\% for White females.
Reasons for these differences are unclear without access to dataset and training details but it is notable and surprising that models do not consistently favor the same demographic group.

\subsubsection{Exclusion}
\paragraph{How sensitive are models to gender and race perturbations?} We compute our proposed metric, exclusion (i.e., how often top retrieved resumes are excluded from the set of top-$n$ resumes after perturbations), and find that all models display notable sensitivity to gender and race name perturbations (Figure \ref{fig:exclusion-initial}). 
When considering the top-5 resumes, we find that models tend to exclude perturbed resumes nearly half the time (45.75\% on average).

Across both gender and race name perturbations, and different $n$ values, text-embedding-3-small and text-embedding-3-large have the highest exclusion, while embed-english-v3.0 consistently has the lowest exclusion.
As expected, exclusion lowers as $n$ increases, since larger $n$ values are less restrictive and consider a larger set of retrieved resumes. 
That being said, exclusion for $n=100$ is still considerable, as all models have exclusion $>$ 12\%.\footnote{In practice we expect $n$ to be low for filtering candidates.}

\begin{figure*}[t]
    \centering
    \begin{subfigure}[b]{0.38\textwidth}
        \centering
        \includegraphics[width=\textwidth]{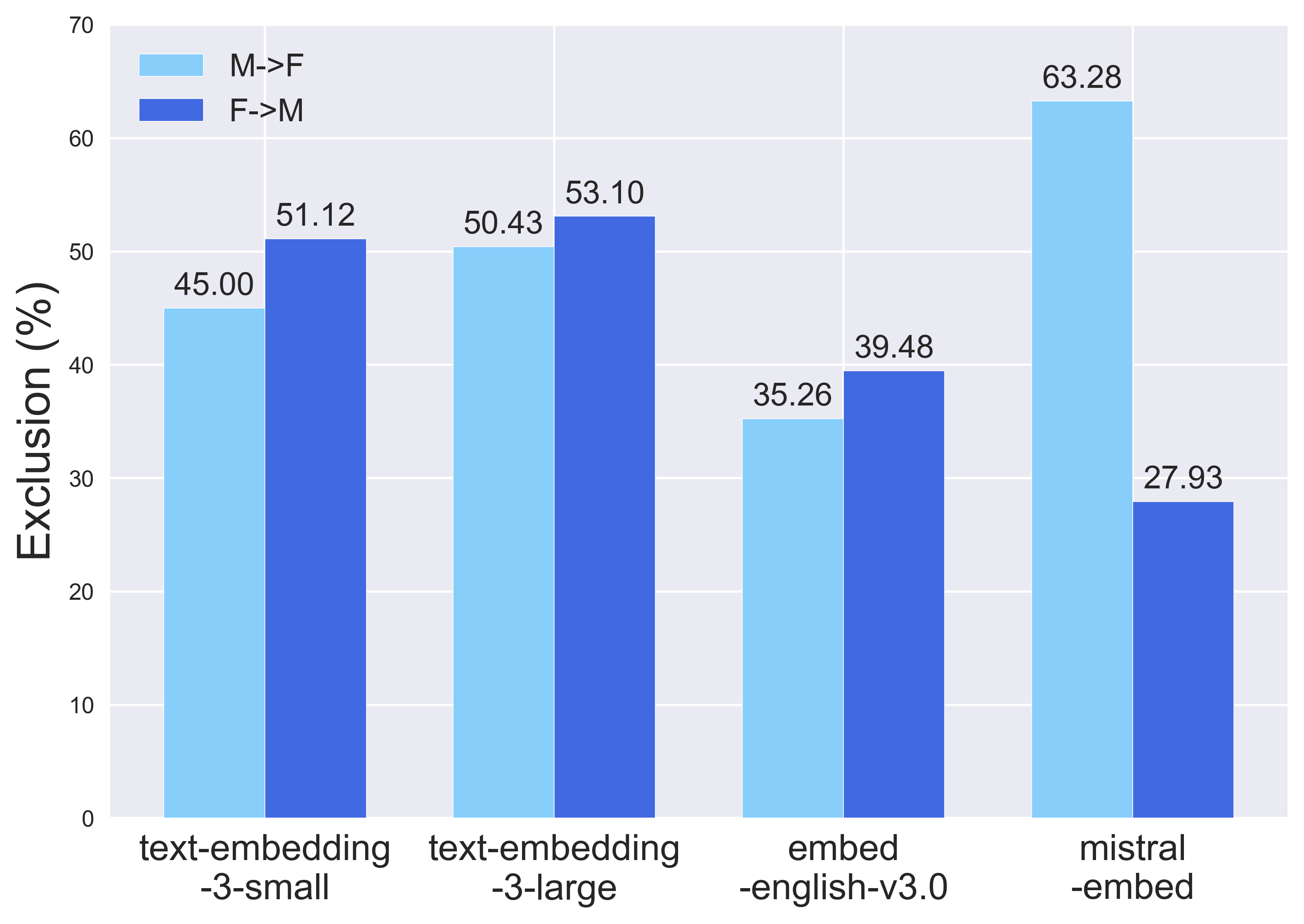}
        \caption{Gender, $n=5$}
        \label{fig:exclusion_nondemo_plot1}
    \end{subfigure}%
    \hspace{0.05\textwidth}  % Explicit spacing between plots
    \begin{subfigure}[b]{0.38\textwidth}
        \centering
        \includegraphics[width=\textwidth]{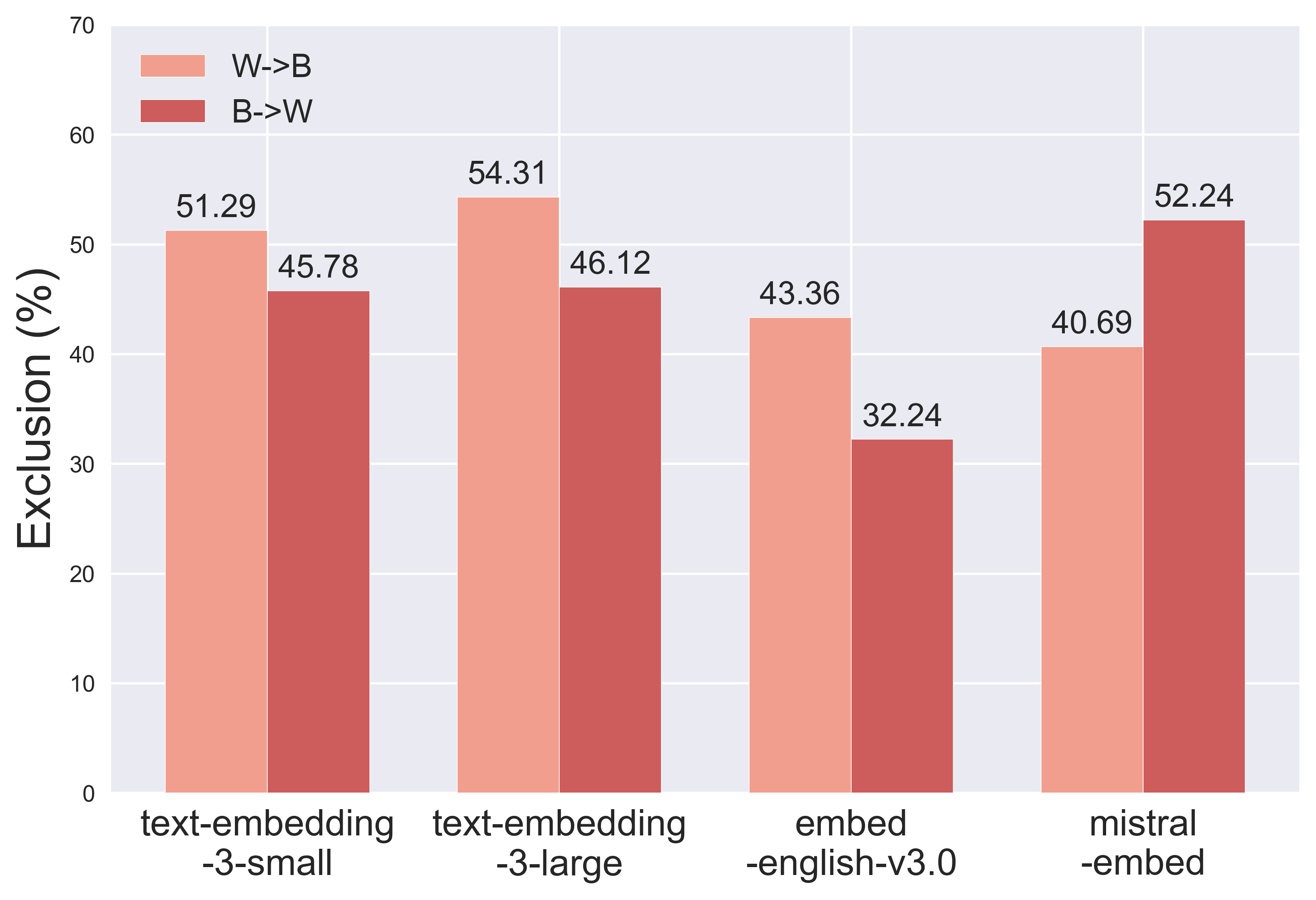}
        \caption{Race, $n=5$}
        \label{fig:exclusion_nondemo_plot2}
    \end{subfigure}
    \caption{\textbf{Directional differences in exclusion metric for retrieval after applying name perturbations} (i.e., separating based on perturbation direction). M→F perturbs male to female names and F→M perturbs female to male names, while W→B perturbs White to Black names and B→W perturbs Black to White names.}
    \label{fig:exclusion-dir}
\end{figure*}

In contrast to our summarization findings, where models show greater invariance violations for race vs. gender perturbations, models have similar sensitivity to gender vs. race perturbations for exclusion. 
Overall, average exclusion for gender is 31.78\% on generated resumes (25.47\% on Kaggle resumes) vs. 31.66\% on generated resumes (25.40\% on Kaggle resumes) for race.\footnote{Kaggle resumes exhibit similar patterns to generated resumes, but are lower in exclusion magnitude. This is likely because generated resumes are tech-focused and more overlapping in content.}
Our analysis reveals that the set of top retrieved resumes with respect to a job posting is highly brittle, as merely altering the demographic with names often results in otherwise identical resumes dropping out of the top-$n$ results.

\begin{figure*}[t]
    \centering
    \begin{subfigure}[b]{0.38\textwidth}
        \centering
        \includegraphics[width=\textwidth]{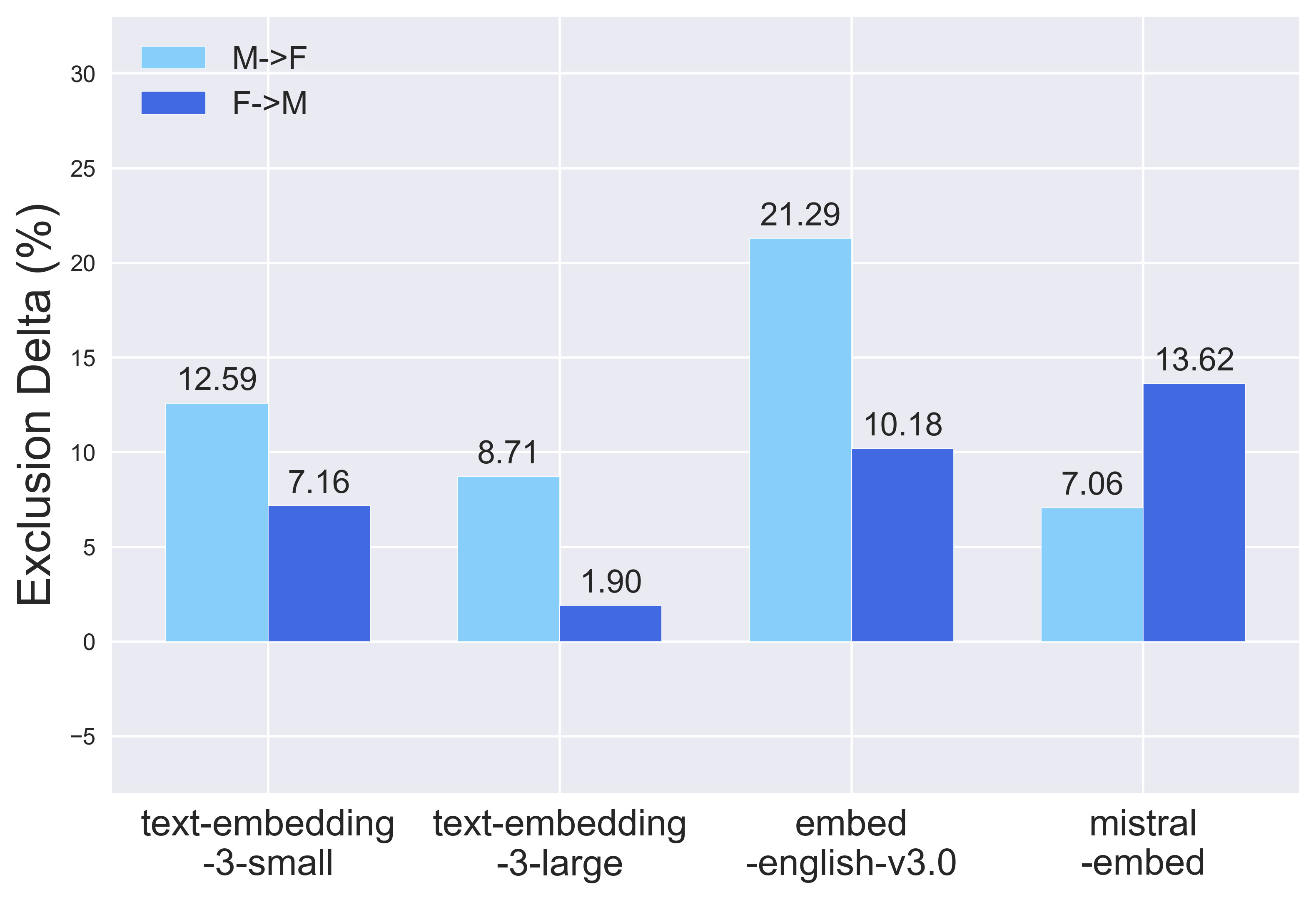}
        \caption{Gender, $n=5$}
        \label{fig:exclusion_extra_gen_plot1}
    \end{subfigure}%
    \hspace{0.05\textwidth}  % Explicit spacing between plots
    \begin{subfigure}[b]{0.38\textwidth}
        \centering
        \includegraphics[width=\textwidth]{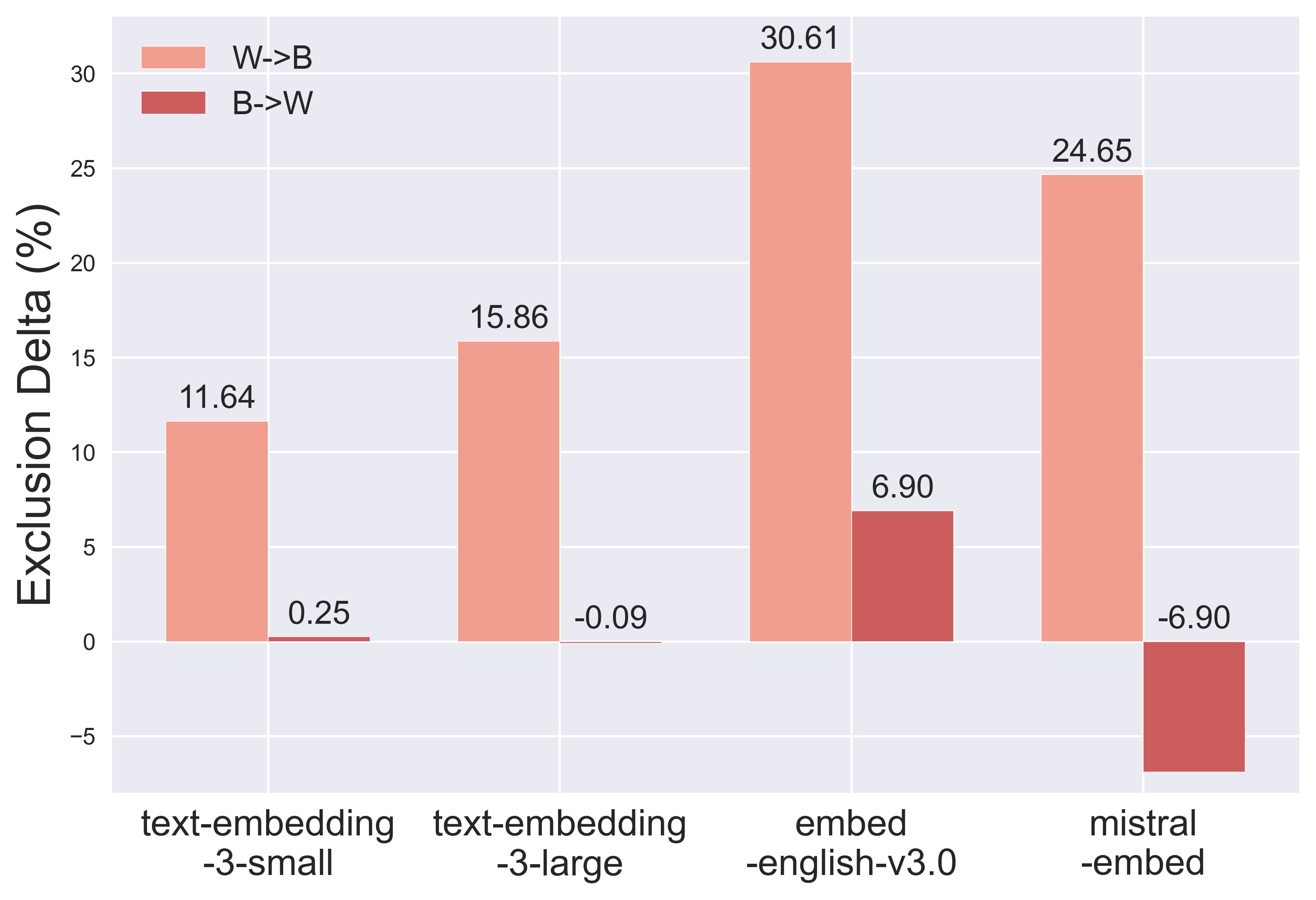}
        \caption{Race, $n=5$}
        \label{fig:exclusion_extra_race_plot1}
    \end{subfigure}
    \caption{\textbf{Deltas (differences) in exclusion metric for retrieval} after performing demographic perturbations with names + extracurricular information vs. names only. As expected, adding extracurricular information increases sensitivity to perturbations.}
    \label{fig:exclusion-extra}
    \vspace{-0.4em}
\end{figure*}
\paragraph{Does model sensitivity to perturbations differ based on the direction of perturbation?}
We partition the results based on the perturbation direction (Figure \ref{fig:exclusion-dir}), and find that models often exhibit higher sensitivity to one direction of perturbation over the other. 
In particular, the gender directional difference is notable for mistral-embed, going from 63.28\% for M $\rightarrow$ F to 27.93\% for F $\rightarrow$ M, for generated resumes with $n=5$.
We also observe that models exhibit opposite directional trends for gender and race. 
For gender, all models except mistral-embed are more sensitive when perturbing female names (marginalized) to male names (non-marginalized).
On the other hand, for race, all models except mistral-embed are more sensitive when perturbing White names (non-marginalized) to Black names (marginalized). 
These results highlight an asymmetry in how models handle various demographic changes.

\paragraph{Are models more sensitive when perturbing both names and extracurricular information, as opposed to names only?}
Figure \ref{fig:exclusion-extra} shows that models tend to be more sensitive when perturbing extracurricular information in addition to names. 
On average, we observe the following increases in exclusion: 9.35\% for M $\rightarrow$ F, 8.06\% for F $\rightarrow$ M, 16.41\% for W $\rightarrow$ B, and 2.90\% for B $\rightarrow$ W. 

For gender, adding extracurricular information results in comparable increases in exclusion for both directions.
In contrast, adding extracurricular information for race results in highly asymmetric increases. W $\rightarrow$ B averages more than 5x the increase of B $\rightarrow$ W changes.
We observe that adding extracurricular information results in non-uniform increases to exclusion, which suggests that models may encode and utilize various types of demographic signal differently.
This finding is notable given that prior work often examines a single way of encoding demographics, overlooking how various signals interact and compound.

\begin{figure*}[t]
    \centering
    \begin{subfigure}[b]{0.38\textwidth}
        \centering
        \includegraphics[width=\textwidth]{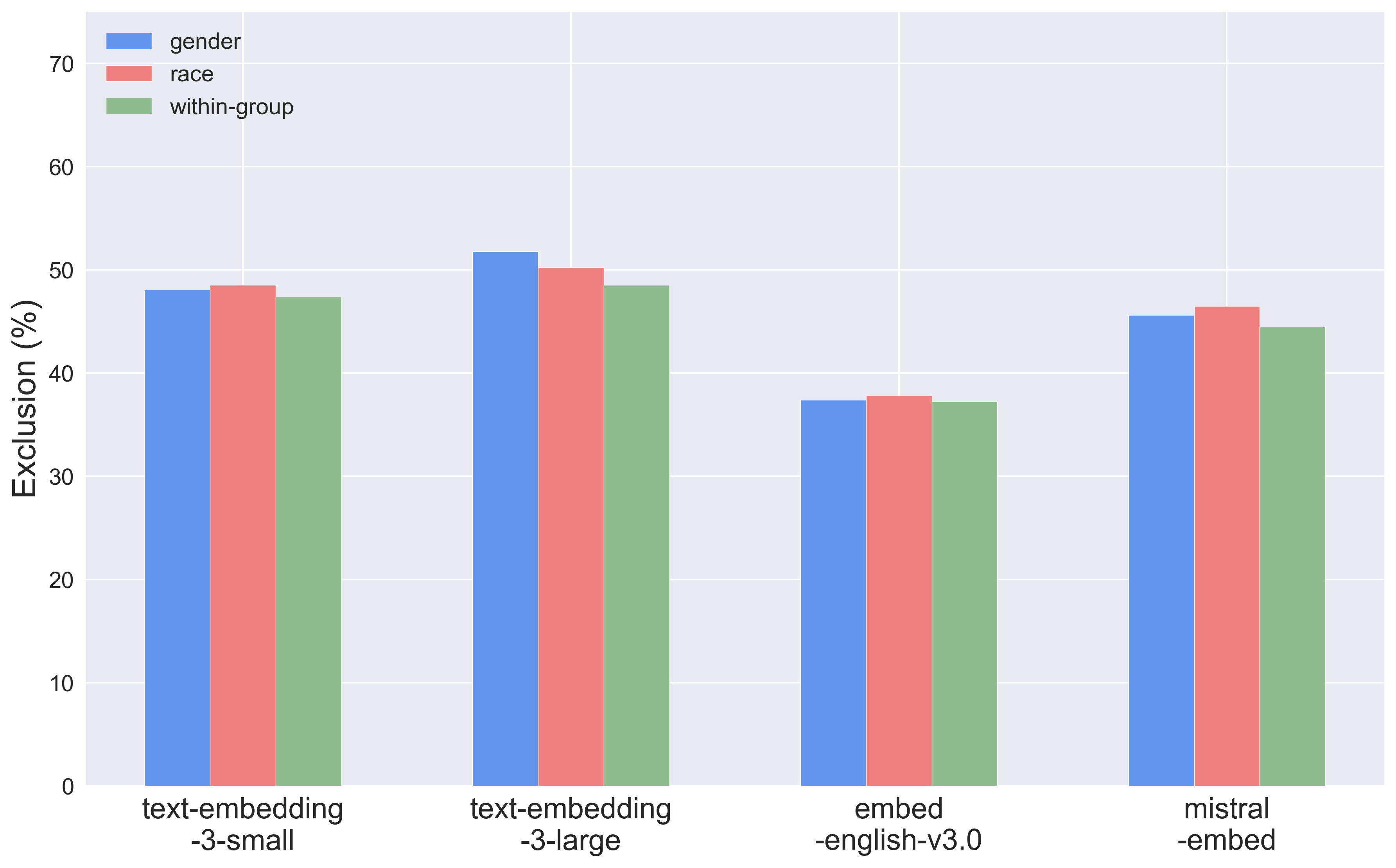}
        \caption{Within-Group Name, $n=5$}
        \label{fig:exclusion_nondemo_plot1}
    \end{subfigure}%
    \hspace{0.05\textwidth}  % Explicit spacing between plots
    \begin{subfigure}[b]{0.38\textwidth}
        \centering
        \includegraphics[width=\textwidth]{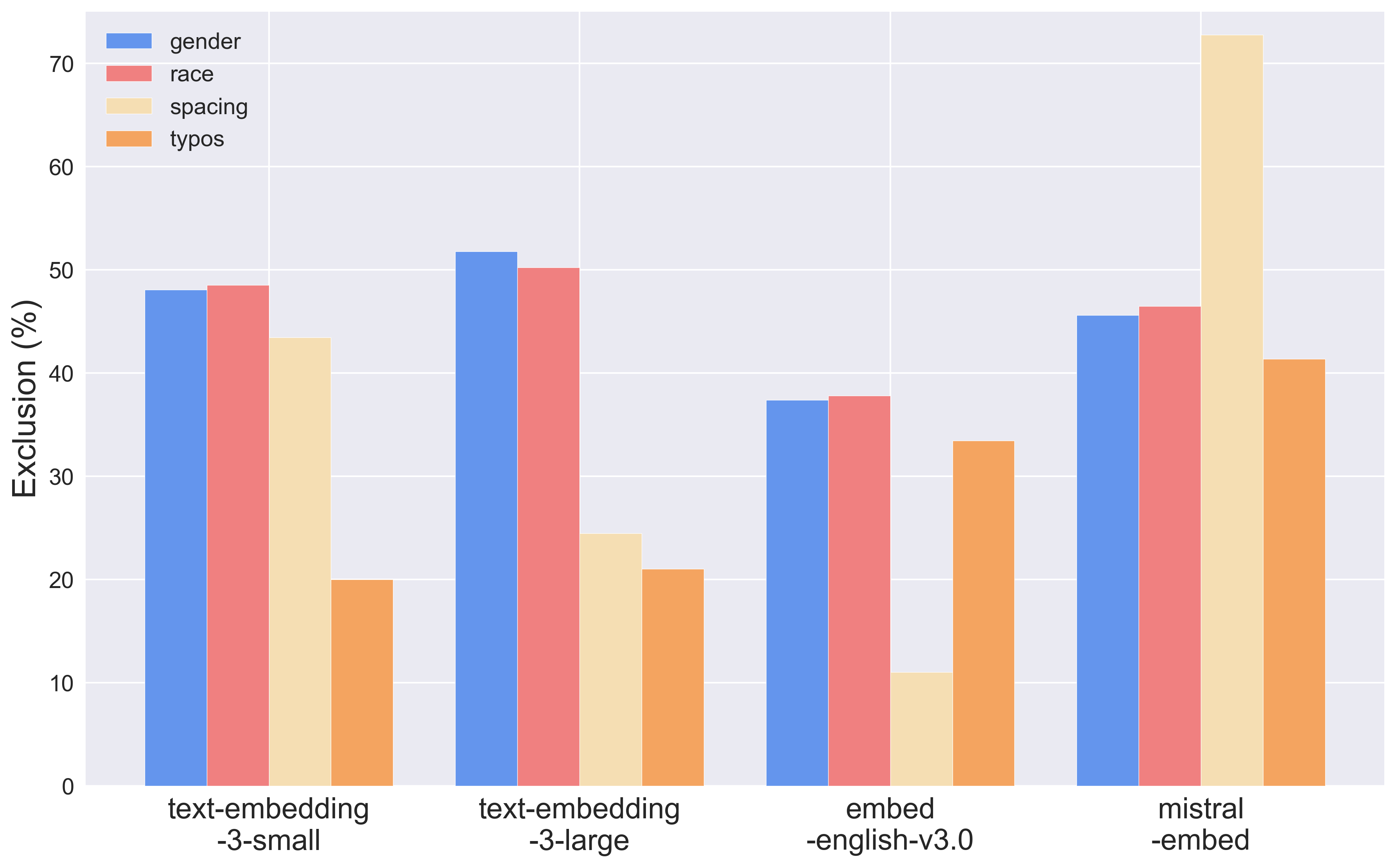}
        \caption{Non-name, $n=5$}
        \label{fig:exclusion_nondemo_plot2}
    \end{subfigure}
    \caption{\textbf{Exclusion metric for retrieval after performing non-demographic perturbations} (i.e., within group name changes - left, and modifying spacing and adding typos - right).}
    \label{fig:exclusion-nondemo}
    \vspace{-0.8em}
\end{figure*}

\paragraph{More broadly, do models exhibit brittleness to non-demographic perturbations?} 
To disentangle fairness from robustness issues, we consider two sets of perturbation analyses that are non-demographic: (1) How sensitive are models to within-group name perturbations? and (2) How sensitive are models to non-name perturbations? 
Even when perturbing names within the same demographic group, models surprisingly exhibit highly similar levels of sensitivity to those observed with gender and race name perturbations (Figure \ref{fig:exclusion_nondemo_plot1}). 

We find that models are extremely sensitive to both spacing and typos, but to a lesser extent than names.
As shown in Figure \ref{fig:exclusion_nondemo_plot2}, most models demonstrate higher sensitivity to spacing than typos, though there is surprising sensitivity to both.
In particular, mistral-embed excludes resumes from the top-5 set 72.76\% of the time solely based on spacing, which indicates that formatting can have a massive impact on fairness (in this case, much more than names).
Generated resumes display nearly twice the sensitivity to spacing changes compared to web-scraped Kaggle resumes (27.98\% vs. 15.35\% averaged across models and $n$ values), likely due to their structured formatting. 
In summary, we observe that retrieval models lack overall robustness, which has fairness implications.

\section{Discussion}

Our results highlight failures of both fairness and robustness in LLMs in hiring contexts. 
We differ from prior work on LLM fairness in summarization \citep{zhang-etal-2024-fair} and ranking \citep{xu-etal-2024-study} in that our evaluations are grounded in real-world applications, and this reveals novel insights that have ecological validity.
% Without this grounding, we cannot adequately assess fairness issues or derive conclusions about impact in actual deployment settings.
First, we observe that model rankings in a retrieval setting are impacted considerably by both subtle demographic and non-demographic changes.
In practice, these differences would lead to unintended exclusion, with candidates being eliminated from consideration during initial screening. 
We also observe that subtle demographic differences in resumes can alter the way candidates are discussed in generated summaries.
As a result, candidates who make it past the initial resume screening stage may be portrayed differently based on demographic attributes, which can impact downstream decision-making.

Additionally, it is important to consider compositional effects when combining components, since biases can compound due to the sequential nature of these tasks. 
The resulting candidate pool may (1) leave out qualified candidates through retrieval bias, and (2) differentially represent candidates through summarization bias. 
Applying this to our results, in the worst case, we find that modifying racial indicators on resumes (using names + extracurricular information) can result in roughly (1) 70\% of candidates being filtered out at the resume screening stage, and (2) 20\% of remaining candidates being depicted in less preferable ways.

Our analysis also reveals that all models are sensitive to non-demographic perturbations, suggesting that model unfairness may partially stem from more general robustness issues, rather than encoded biases alone. 
These perturbations still result in disparate outcomes, but with different underlying causes.
While these insights do not change discriminatory impact, understanding that disparate treatment can arise from small input changes, both demographic and otherwise, provides a more complete picture for addressing fairness issues.
% Moreover, given that retrieval models are commonly used in RAG systems, these issues may extend to various applications beyond HR.

\section{Conclusion}
We examine allocational fairness in LLM-based hiring systems by analyzing two key components: applicant ranking and summary generation.
To support systematic measurement and mitigation of fairness issues, we release a benchmark dataset and introduce an evaluation framework with new metrics.\footnote{The data and code can be found at: \url{https://github.com/preethisesh/hiring_fairness}.}
We find that a hiring pipeline consisting of these two stages produces biased outcomes, particularly during the retrieval phase. 
In addition, models show unexpected sensitivity to minor non-demographic changes, revealing a lack of overall robustness that may contribute to unfair outcomes.
These findings underscore the need for targeted strategies to improve the fairness of LLM-based hiring, and the importance of realistic, application-grounded evaluations of LLM harms.

\section*{Limitations}
Our analysis focuses exclusively on English resumes and job posts.
Future research should investigate fairness considerations in multilingual settings and examine whether our conclusions hold across various languages.
Additionally, cultural norms likely influence how candidates present themselves and describe their professional experience, qualifications, and achievements. 
Understanding these nuances is crucial for evaluating and developing hiring systems that serve diverse global talent pools.
Since we are releasing our code and datasets, researchers in other regions will be able to expand our work as well.

While our analysis examines whether hiring systems behave differently for various gender (male and female) and racial (White and Black) groups, it is meant to be illustrative rather than exhaustive and only covers a subset of gender and racial identities.
We only consider binary gender biases, and exclude non-binary gender biases from our analysis, since this information cannot be inferred from a name.
While candidates may explicitly declare pronouns on resumes, we do not observe this in the resumes we collect, so we do not vary them.
In addition, we only focus on Black and White racial groups, since this is a common emphasis in fairness studies, and only to do so in the context of US names. 
We hope future work expands beyond these commonly investigated biases and analyzes the extent to which other types of demographic information (e.g., age and nationality) impact LLM fairness in hiring.
% We focus on names as the primary demographic signal, since they are always present in resumes.
% We also go beyond names and vary extracurricular information, since this is commonly found in resumes and often contains implicit and explicit demographic information.

Moreover, although the way we handle name perturbations is standard practice in NLP fairness literature, we acknowledge that names can encode demographic axes beyond gender and race, including age, class, and region.
These signals are more subtle and challenging to isolate, making it difficult in practice to vary only a single dimension at a time.  
It is worth noting that we control for other factors such as name frequency to reduce potential confounds. 

\bibliography{custom}

\begin{thebibliography}{54}
\providecommand{\natexlab}[1]{#1}

\bibitem[{Abid et~al.(2021)Abid, Farooqi, and Zou}]{Abid2021LargeLM}
Abubakar Abid, Maheen Farooqi, and James Zou. 2021.
\newblock \href {https://api.semanticscholar.org/CorpusID:236384212} {Large language models associate muslims with violence}.
\newblock \emph{Nature Machine Intelligence}, 3:461 -- 463.

\bibitem[{An et~al.(2024)An, Acquaye, Wang, Li, and Rudinger}]{an-etal-2024-large}
Haozhe An, Christabel Acquaye, Colin Wang, Zongxia Li, and Rachel Rudinger. 2024.
\newblock \href {https://doi.org/10.18653/v1/2024.acl-short.37} {Do large language models discriminate in hiring decisions on the basis of race, ethnicity, and gender?}
\newblock In \emph{Proceedings of the 62nd Annual Meeting of the Association for Computational Linguistics (Volume 2: Short Papers)}, pages 386--397, Bangkok, Thailand. Association for Computational Linguistics.

\bibitem[{An and Rudinger(2023)}]{an-rudinger-2023-nichelle}
Haozhe An and Rachel Rudinger. 2023.
\newblock \href {https://doi.org/10.18653/v1/2023.acl-short.34} {Nichelle and nancy: The influence of demographic attributes and tokenization length on first name biases}.
\newblock In \emph{Proceedings of the 61st Annual Meeting of the Association for Computational Linguistics (Volume 2: Short Papers)}, pages 388--401, Toronto, Canada. Association for Computational Linguistics.

\bibitem[{Barocas et~al.(2017)Barocas, Crawford, Shapiro, and Wallach}]{barocas2017problem}
Solon Barocas, Kate Crawford, Aaron Shapiro, and Hanna Wallach. 2017.
\newblock The problem with bias: Allocative versus representational harms in machine learning.
\newblock In \emph{9th Annual conference of the special interest group for computing, information and society}, page~1. New York, NY.

\bibitem[{Benjamini and Hochberg(1995)}]{benjamini1995controlling}
Yoav Benjamini and Yosef Hochberg. 1995.
\newblock Controlling the false discovery rate: a practical and powerful approach to multiple testing.
\newblock \emph{Journal of the Royal statistical society: series B (Methodological)}, 57(1):289--300.

\bibitem[{Bhawal(2021)}]{bhawal2021resume}
Snehaan Bhawal. 2021.
\newblock Resume dataset.
\newblock \url{https://www.kaggle.com/datasets/snehaanbhawal/resume-dataset}.

\bibitem[{Bland and Altman(1995)}]{bland1995multiple}
J~Martin Bland and Douglas~G Altman. 1995.
\newblock Multiple significance tests: the bonferroni method.
\newblock \emph{Bmj}, 310(6973):170.

\bibitem[{Blodgett et~al.(2020)Blodgett, Barocas, Daum{\'e}~III, and Wallach}]{blodgett-etal-2020-language}
Su~Lin Blodgett, Solon Barocas, Hal Daum{\'e}~III, and Hanna Wallach. 2020.
\newblock \href {https://doi.org/10.18653/v1/2020.acl-main.485} {Language (technology) is power: A critical survey of {``}bias{''} in {NLP}}.
\newblock In \emph{Proceedings of the 58th Annual Meeting of the Association for Computational Linguistics}, pages 5454--5476, Online. Association for Computational Linguistics.

\bibitem[{Blodgett et~al.(2021)Blodgett, Lopez, Olteanu, Sim, and Wallach}]{blodgett-etal-2021-stereotyping}
Su~Lin Blodgett, Gilsinia Lopez, Alexandra Olteanu, Robert Sim, and Hanna Wallach. 2021.
\newblock \href {https://doi.org/10.18653/v1/2021.acl-long.81} {Stereotyping {N}orwegian salmon: An inventory of pitfalls in fairness benchmark datasets}.
\newblock In \emph{Proceedings of the 59th Annual Meeting of the Association for Computational Linguistics and the 11th International Joint Conference on Natural Language Processing (Volume 1: Long Papers)}, pages 1004--1015, Online. Association for Computational Linguistics.

\bibitem[{{Boston Consulting Group}(2025)}]{bcg2025_ai_recruitment}
{Boston Consulting Group}. 2025.
\newblock \href {https://www.bcg.com/publications/2025/ai-changing-recruitment} {How ai is changing recruitment}.

\bibitem[{Cao et~al.(2022)Cao, Pruksachatkun, Chang, Gupta, Kumar, Dhamala, and Galstyan}]{cao-etal-2022-intrinsic}
Yang~Trista Cao, Yada Pruksachatkun, Kai-Wei Chang, Rahul Gupta, Varun Kumar, Jwala Dhamala, and Aram Galstyan. 2022.
\newblock \href {https://doi.org/10.18653/v1/2022.acl-short.62} {On the intrinsic and extrinsic fairness evaluation metrics for contextualized language representations}.
\newblock In \emph{Proceedings of the 60th Annual Meeting of the Association for Computational Linguistics (Volume 2: Short Papers)}, pages 561--570, Dublin, Ireland. Association for Computational Linguistics.

\bibitem[{Cheng et~al.(2023)Cheng, Durmus, and Jurafsky}]{cheng-etal-2023-marked}
Myra Cheng, Esin Durmus, and Dan Jurafsky. 2023.
\newblock \href {https://doi.org/10.18653/v1/2023.acl-long.84} {Marked personas: Using natural language prompts to measure stereotypes in language models}.
\newblock In \emph{Proceedings of the 61st Annual Meeting of the Association for Computational Linguistics (Volume 1: Long Papers)}, pages 1504--1532, Toronto, Canada. Association for Computational Linguistics.

\bibitem[{{Cohere}(2024)}]{cohere2024command}
{Cohere}. 2024.
\newblock Command-r.
\newblock \url{https://docs.cohere.com/v2/docs/command-r}.

\bibitem[{Dastin(2018)}]{dastin2018amazon}
Jeffrey Dastin. 2018.
\newblock \href {https://www.reuters.com/article/world/insight-amazon-scraps-secret-ai-recruiting-tool-that-showed-bias-against-women-idUSKCN1MK0AG} {Insight - {Amazon} scraps secret {AI} recruiting tool that showed bias against women}.
\newblock \emph{Reuters}.

\bibitem[{Dwork et~al.(2012)Dwork, Hardt, Pitassi, Reingold, and Zemel}]{dwork-etal-2012}
Cynthia Dwork, Moritz Hardt, Toniann Pitassi, Omer Reingold, and Richard Zemel. 2012.
\newblock \href {https://doi.org/10.1145/2090236.2090255} {Fairness through awareness}.
\newblock In \emph{Proceedings of the 3rd Innovations in Theoretical Computer Science Conference}, ITCS '12, page 214–226, New York, NY, USA. Association for Computing Machinery.

\bibitem[{Elazar et~al.(2023)Elazar, Bhagia, Magnusson, Ravichander, Schwenk, Suhr, Walsh, Groeneveld, Soldaini, Singh, Hajishirzi, Smith, and Dodge}]{elazar2023s}
Yanai Elazar, Akshita Bhagia, Ian~Helgi Magnusson, Abhilasha Ravichander, Dustin Schwenk, Alane Suhr, Evan~Pete Walsh, Dirk Groeneveld, Luca Soldaini, Sameer Singh, Hanna Hajishirzi, Noah~A. Smith, and Jesse Dodge. 2023.
\newblock What's in my big data?
\newblock In \emph{The Twelfth International Conference on Learning Representations}.

\bibitem[{Ethayarajh et~al.(2019)Ethayarajh, Duvenaud, and Hirst}]{ethayarajh-etal-2019-understanding}
Kawin Ethayarajh, David Duvenaud, and Graeme Hirst. 2019.
\newblock \href {https://doi.org/10.18653/v1/P19-1166} {Understanding undesirable word embedding associations}.
\newblock In \emph{Proceedings of the 57th Annual Meeting of the Association for Computational Linguistics}, pages 1696--1705, Florence, Italy. Association for Computational Linguistics.

\bibitem[{Gadiraju et~al.(2023)Gadiraju, Kane, Dev, Taylor, Wang, Denton, and Brewer}]{gadiraju2023}
Vinitha Gadiraju, Shaun Kane, Sunipa Dev, Alex Taylor, Ding Wang, Emily Denton, and Robin Brewer. 2023.
\newblock \href {https://doi.org/10.1145/3593013.3593989} {"i wouldn’t say offensive but...": Disability-centered perspectives on large language models}.
\newblock In \emph{Proceedings of the 2023 ACM Conference on Fairness, Accountability, and Transparency}, FAccT '23, page 205–216, New York, NY, USA. Association for Computing Machinery.

\bibitem[{Gao et~al.(2020)Gao, Biderman, Black, Golding, Hoppe, Foster, Phang, He, Thite, Nabeshima, Presser, and Leahy}]{pile}
Leo Gao, Stella Biderman, Sid Black, Laurence Golding, Travis Hoppe, Charles Foster, Jason Phang, Horace He, Anish Thite, Noa Nabeshima, Shawn Presser, and Connor Leahy. 2020.
\newblock \href {https://arxiv.org/abs/2101.00027} {The pile: An 800gb dataset of diverse text for language modeling}.
\newblock \emph{Preprint}, arXiv:2101.00027.

\bibitem[{Gautam et~al.(2024)Gautam, Subramonian, Lauscher, and Keyes}]{gautam-etal-2024-stop}
Vagrant Gautam, Arjun Subramonian, Anne Lauscher, and Os~Keyes. 2024.
\newblock \href {https://doi.org/10.18653/v1/2024.gebnlp-1.20} {Stop! in the name of flaws: Disentangling personal names and sociodemographic attributes in {NLP}}.
\newblock In \emph{Proceedings of the 5th Workshop on Gender Bias in Natural Language Processing (GeBNLP)}, pages 323--337, Bangkok, Thailand. Association for Computational Linguistics.

\bibitem[{Glazko et~al.(2024)Glazko, Mohammed, Kosa, Potluri, and Mankoff}]{glazko2024}
Kate Glazko, Yusuf Mohammed, Ben Kosa, Venkatesh Potluri, and Jennifer Mankoff. 2024.
\newblock \href {https://doi.org/10.1145/3630106.3658933} {Identifying and improving disability bias in gpt-based resume screening}.
\newblock In \emph{Proceedings of the 2024 ACM Conference on Fairness, Accountability, and Transparency}, FAccT '24, page 687–700, New York, NY, USA. Association for Computing Machinery.

\bibitem[{Goldfarb-Tarrant et~al.(2021)Goldfarb-Tarrant, Marchant, Mu{\~n}oz~S{\'a}nchez, Pandya, and Lopez}]{goldfarb-tarrant-etal-2021-intrinsic}
Seraphina Goldfarb-Tarrant, Rebecca Marchant, Ricardo Mu{\~n}oz~S{\'a}nchez, Mugdha Pandya, and Adam Lopez. 2021.
\newblock \href {https://doi.org/10.18653/v1/2021.acl-long.150} {Intrinsic bias metrics do not correlate with application bias}.
\newblock In \emph{Proceedings of the 59th Annual Meeting of the Association for Computational Linguistics and the 11th International Joint Conference on Natural Language Processing (Volume 1: Long Papers)}, pages 1926--1940, Online. Association for Computational Linguistics.

\bibitem[{Guo et~al.(2023)Guo, Jin, Liu, Huang, Shi, Supryadi, Yu, Liu, Li, Xiong, and Xiong}]{guo2023evaluatinglargelanguagemodels}
Zishan Guo, Renren Jin, Chuang Liu, Yufei Huang, Dan Shi, Supryadi, Linhao Yu, Yan Liu, Jiaxuan Li, Bojian Xiong, and Deyi Xiong. 2023.
\newblock \href {https://arxiv.org/abs/2310.19736} {Evaluating large language models: A comprehensive survey}.
\newblock \emph{Preprint}, arXiv:2310.19736.

\bibitem[{Haim et~al.(2024)Haim, Salinas, and Nyarko}]{haim2024whatsnameauditinglarge}
Amit Haim, Alejandro Salinas, and Julian Nyarko. 2024.
\newblock \href {https://arxiv.org/abs/2402.14875} {What's in a name? auditing large language models for race and gender bias}.
\newblock \emph{Preprint}, arXiv:2402.14875.

\bibitem[{Herman(2024)}]{herman2024ways}
Hannah Herman. 2024.
\newblock \href {https://zapier.com/blog/automate-recruit-crm} {6 ways to automate {Recruit CRM} with {Zapier}}.
\newblock \emph{Zapier Blog}.

\bibitem[{Humanly(2024)}]{humanly2024using}
Humanly. 2024.
\newblock \href {https://www.humanly.io/resource-library/blog/using-ai-to-streamline-the-recruiting-process} {Using {AI} to streamline the recruiting process}.
\newblock \emph{Humanly.io Blog}.

\bibitem[{Jiang et~al.(2024)Jiang, Sablayrolles, Roux, Mensch, Savary, Bamford, Chaplot, de~las Casas, Hanna, Bressand, Lengyel, Bour, Lample, Lavaud, Saulnier, Lachaux, Stock, Subramanian, Yang, Antoniak, Scao, Gervet, Lavril, Wang, Lacroix, and Sayed}]{jiang2024mixtralexperts}
Albert~Q. Jiang, Alexandre Sablayrolles, Antoine Roux, Arthur Mensch, Blanche Savary, Chris Bamford, Devendra~Singh Chaplot, Diego de~las Casas, Emma~Bou Hanna, Florian Bressand, Gianna Lengyel, Guillaume Bour, Guillaume Lample, Lélio~Renard Lavaud, Lucile Saulnier, Marie-Anne Lachaux, Pierre Stock, Sandeep Subramanian, Sophia Yang, Szymon Antoniak, Teven~Le Scao, Théophile Gervet, Thibaut Lavril, Thomas Wang, Timothée Lacroix, and William~El Sayed. 2024.
\newblock \href {https://arxiv.org/abs/2401.04088} {Mixtral of experts}.
\newblock \emph{Preprint}, arXiv:2401.04088.

\bibitem[{Kelly(2023)}]{kelly2023ai}
Jack Kelly. 2023.
\newblock \href {https://www.forbes.com/sites/jackkelly/2023/03/15/how-ai-powered-tech-can-help-recruiters-and-hiring-managers-find-candidates-quicker-and-more-efficiently} {How {AI}-powered tech can help recruiters and hiring managers find candidates quicker and more efficiently}.
\newblock \emph{Forbes}.

\bibitem[{Kincaid(1975)}]{kincaid1975derivation}
JP~Kincaid. 1975.
\newblock Derivation of new readability formulas (automated readability index, fog count and flesch reading ease formula) for navy enlisted personnel.
\newblock \emph{Chief of Naval Technical Training}.

\bibitem[{Kirk et~al.(2021)Kirk, Jun, Volpin, Iqbal, Benussi, Dreyer, Shtedritski, and Asano}]{kirk2021}
Hannah~Rose Kirk, Yennie Jun, Filippo Volpin, Haider Iqbal, Elias Benussi, Frederic Dreyer, Aleksandar Shtedritski, and Yuki Asano. 2021.
\newblock \href {https://proceedings.neurips.cc/paper_files/paper/2021/file/1531beb762df4029513ebf9295e0d34f-Paper.pdf} {Bias out-of-the-box: An empirical analysis of intersectional occupational biases in popular generative language models}.
\newblock In \emph{Advances in Neural Information Processing Systems}, volume~34, pages 2611--2624. Curran Associates, Inc.

\bibitem[{Kusner et~al.(2017)Kusner, Loftus, Russell, and Silva}]{kusner-etal-2017}
Matt Kusner, Joshua Loftus, Chris Russell, and Ricardo Silva. 2017.
\newblock Counterfactual fairness.
\newblock In \emph{Proceedings of the 31st International Conference on Neural Information Processing Systems}, NIPS'17, page 4069–4079, Red Hook, NY, USA. Curran Associates Inc.

\bibitem[{Lewis et~al.(2020)Lewis, Perez, Piktus, Petroni, Karpukhin, Goyal, K\"{u}ttler, Lewis, Yih, Rockt\"{a}schel, Riedel, and Kiela}]{lewis2020}
Patrick Lewis, Ethan Perez, Aleksandra Piktus, Fabio Petroni, Vladimir Karpukhin, Naman Goyal, Heinrich K\"{u}ttler, Mike Lewis, Wen-tau Yih, Tim Rockt\"{a}schel, Sebastian Riedel, and Douwe Kiela. 2020.
\newblock \href {https://proceedings.neurips.cc/paper_files/paper/2020/file/6b493230205f780e1bc26945df7481e5-Paper.pdf} {Retrieval-augmented generation for knowledge-intensive nlp tasks}.
\newblock In \emph{Advances in Neural Information Processing Systems}, volume~33, pages 9459--9474. Curran Associates, Inc.

\bibitem[{Li et~al.(2025)Li, Zhang, and Chaturvedi}]{li-etal-2025-improving}
Haoyuan Li, Rui Zhang, and Snigdha Chaturvedi. 2025.
\newblock \href {https://aclanthology.org/2025.acl-short.90/} {Improving fairness of large language models in multi-document summarization}.
\newblock In \emph{Proceedings of the 63rd Annual Meeting of the Association for Computational Linguistics (Volume 2: Short Papers)}, pages 1143--1154, Vienna, Austria. Association for Computational Linguistics.

\bibitem[{Mehrabi et~al.(2021)Mehrabi, Morstatter, Saxena, Lerman, and Galstyan}]{mehrabi-etal-2021}
Ninareh Mehrabi, Fred Morstatter, Nripsuta Saxena, Kristina Lerman, and Aram Galstyan. 2021.
\newblock \href {https://doi.org/10.1145/3457607} {A survey on bias and fairness in machine learning}.
\newblock \emph{ACM Comput. Surv.}, 54(6).

\bibitem[{{Meta AI}(2024)}]{meta2024llama3}
{Meta AI}. 2024.
\newblock Meta llama 3.
\newblock \url{https://ai.meta.com/blog/meta-llama-3/}.
\newblock Blog post.

\bibitem[{{Microsoft}()}]{microsoft_presidio_anonymizer}
{Microsoft}.
\newblock {Presidio Anonymizer}.
\newblock \url{https://microsoft.github.io/presidio/anonymizer/}.

\bibitem[{Nghiem et~al.(2024)Nghiem, Prindle, Zhao, and Daum{\'e}~Iii}]{nghiem-etal-2024-gotta}
Huy Nghiem, John Prindle, Jieyu Zhao, and Hal Daum{\'e}~Iii. 2024.
\newblock \href {https://aclanthology.org/2024.emnlp-main.413} {{``}you gotta be a doctor, lin{''} : An investigation of name-based bias of large language models in employment recommendations}.
\newblock In \emph{Proceedings of the 2024 Conference on Empirical Methods in Natural Language Processing}, pages 7268--7287, Miami, Florida, USA. Association for Computational Linguistics.

\bibitem[{{OpenAI}(2024)}]{openai2024gpt4}
{OpenAI}. 2024.
\newblock Gpt-4o.
\newblock \url{https://platform.openai.com/docs/models/gpt-4o}.

\bibitem[{Raghavan et~al.(2020)Raghavan, Barocas, Kleinberg, and Levy}]{raghavan2020}
Manish Raghavan, Solon Barocas, Jon Kleinberg, and Karen Levy. 2020.
\newblock \href {https://doi.org/10.1145/3351095.3372828} {Mitigating bias in algorithmic hiring: evaluating claims and practices}.
\newblock In \emph{Proceedings of the 2020 Conference on Fairness, Accountability, and Transparency}, FAT* '20, page 469–481, New York, NY, USA. Association for Computing Machinery.

\bibitem[{Ribeiro et~al.(2020)Ribeiro, Wu, Guestrin, and Singh}]{ribeiro-etal-2020-beyond}
Marco~Tulio Ribeiro, Tongshuang Wu, Carlos Guestrin, and Sameer Singh. 2020.
\newblock \href {https://doi.org/10.18653/v1/2020.acl-main.442} {Beyond accuracy: Behavioral testing of {NLP} models with {C}heck{L}ist}.
\newblock In \emph{Proceedings of the 58th Annual Meeting of the Association for Computational Linguistics}, pages 4902--4912, Online. Association for Computational Linguistics.

\bibitem[{S\'{a}nchez-Monedero et~al.(2020)S\'{a}nchez-Monedero, Dencik, and Edwards}]{sanchez-monedero20}
Javier S\'{a}nchez-Monedero, Lina Dencik, and Lilian Edwards. 2020.
\newblock \href {https://doi.org/10.1145/3351095.3372849} {What does it mean to 'solve' the problem of discrimination in hiring? social, technical and legal perspectives from the uk on automated hiring systems}.
\newblock In \emph{Proceedings of the 2020 Conference on Fairness, Accountability, and Transparency}, FAT* '20, page 458–468, New York, NY, USA. Association for Computing Machinery.

\bibitem[{Shandilya et~al.(2018)Shandilya, Ghosh, and Ghosh}]{shandilya-etal-2018}
Anurag Shandilya, Kripabandhu Ghosh, and Saptarshi Ghosh. 2018.
\newblock \href {https://doi.org/10.1145/3184558.3186947} {Fairness of extractive text summarization}.
\newblock In \emph{Companion Proceedings of the The Web Conference 2018}, WWW '18, page 97–98, Republic and Canton of Geneva, CHE. International World Wide Web Conferences Steering Committee.

\bibitem[{Sheng et~al.(2019)Sheng, Chang, Natarajan, and Peng}]{sheng-etal-2019-woman}
Emily Sheng, Kai-Wei Chang, Premkumar Natarajan, and Nanyun Peng. 2019.
\newblock \href {https://doi.org/10.18653/v1/D19-1339} {The woman worked as a babysitter: On biases in language generation}.
\newblock In \emph{Proceedings of the 2019 Conference on Empirical Methods in Natural Language Processing and the 9th International Joint Conference on Natural Language Processing (EMNLP-IJCNLP)}, pages 3407--3412, Hong Kong, China. Association for Computational Linguistics.

\bibitem[{Steen and Markert(2024)}]{steen2024biasnewssummarizationmeasures}
Julius Steen and Katja Markert. 2024.
\newblock \href {https://arxiv.org/abs/2309.08047} {Bias in news summarization: Measures, pitfalls and corpora}.
\newblock \emph{Preprint}, arXiv:2309.08047.

\bibitem[{Suresh and Guttag(2021)}]{suresh2021}
Harini Suresh and John Guttag. 2021.
\newblock \href {https://doi.org/10.1145/3465416.3483305} {A framework for understanding sources of harm throughout the machine learning life cycle}.
\newblock In \emph{Proceedings of the 1st ACM Conference on Equity and Access in Algorithms, Mechanisms, and Optimization}, EAAMO '21, New York, NY, USA. Association for Computing Machinery.

\bibitem[{Tamkin et~al.(2023)Tamkin, Askell, Lovitt, Durmus, Joseph, Kravec, Nguyen, Kaplan, and Ganguli}]{tamkin2023evaluatingmitigatingdiscriminationlanguage}
Alex Tamkin, Amanda Askell, Liane Lovitt, Esin Durmus, Nicholas Joseph, Shauna Kravec, Karina Nguyen, Jared Kaplan, and Deep Ganguli. 2023.
\newblock \href {https://arxiv.org/abs/2312.03689} {Evaluating and mitigating discrimination in language model decisions}.
\newblock \emph{Preprint}, arXiv:2312.03689.

\bibitem[{Wan et~al.(2023)Wan, Pu, Sun, Garimella, Chang, and Peng}]{wan-etal-2023-kelly}
Yixin Wan, George Pu, Jiao Sun, Aparna Garimella, Kai-Wei Chang, and Nanyun Peng. 2023.
\newblock \href {https://doi.org/10.18653/v1/2023.findings-emnlp.243} {{``}kelly is a warm person, joseph is a role model{''}: Gender biases in {LLM}-generated reference letters}.
\newblock In \emph{Findings of the Association for Computational Linguistics: EMNLP 2023}, pages 3730--3748, Singapore. Association for Computational Linguistics.

\bibitem[{Wang et~al.(2024)Wang, Wu, Wu, Tao, and Fang}]{wang-etal-2024-large}
Yuan Wang, Xuyang Wu, Hsin-Tai Wu, Zhiqiang Tao, and Yi~Fang. 2024.
\newblock \href {https://doi.org/10.18653/v1/2024.naacl-long.319} {Do large language models rank fairly? an empirical study on the fairness of {LLM}s as rankers}.
\newblock In \emph{Proceedings of the 2024 Conference of the North American Chapter of the Association for Computational Linguistics: Human Language Technologies (Volume 1: Long Papers)}, pages 5712--5724, Mexico City, Mexico. Association for Computational Linguistics.

\bibitem[{Webster et~al.(2021)Webster, Wang, Tenney, Beutel, Pitler, Pavlick, Chen, Chi, and Petrov}]{webster2021measuringreducinggenderedcorrelations}
Kellie Webster, Xuezhi Wang, Ian Tenney, Alex Beutel, Emily Pitler, Ellie Pavlick, Jilin Chen, Ed~Chi, and Slav Petrov. 2021.
\newblock \href {https://arxiv.org/abs/2010.06032} {Measuring and reducing gendered correlations in pre-trained models}.
\newblock \emph{Preprint}, arXiv:2010.06032.

\bibitem[{Wilson and Caliskan(2024)}]{wilson2024intersectionalbias}
Kyra Wilson and Aylin Caliskan. 2024.
\newblock \href {https://arxiv.org/abs/2407.20371} {Gender, race, and intersectional bias in resume screening via language model retrieval}.
\newblock \emph{Preprint}, arXiv:2407.20371.

\bibitem[{Xu et~al.(2024)Xu, Wang, Li, Pang, Xu, and Chua}]{xu-etal-2024-study}
Chen Xu, Wenjie Wang, Yuxin Li, Liang Pang, Jun Xu, and Tat-Seng Chua. 2024.
\newblock \href {https://doi.org/10.18653/v1/2024.findings-emnlp.467} {A study of implicit ranking unfairness in large language models}.
\newblock In \emph{Findings of the Association for Computational Linguistics: EMNLP 2024}, pages 7957--7970, Miami, Florida, USA. Association for Computational Linguistics.

\bibitem[{Yin et~al.(2024)Yin, Alba, and Nicoletti}]{yin24-recruiting-bias}
Leon Yin, Davey Alba, and Leonardo Nicoletti. 2024.
\newblock \href {https://www.bloomberg.com/graphics/2024-openai-gpt-hiring-racial-discrimination/} {Openai’s gpt is a recruiter’s dream tool. tests show there’s racial bias}.
\newblock \emph{Bloomberg}.

\bibitem[{Zhang et~al.(2024)Zhang, Zhang, Liu, Fabbri, Liu, Kamoi, Lu, Xiong, Zhao, Radev, McKeown, and Zhang}]{zhang-etal-2024-fair}
Yusen Zhang, Nan Zhang, Yixin Liu, Alexander Fabbri, Junru Liu, Ryo Kamoi, Xiaoxin Lu, Caiming Xiong, Jieyu Zhao, Dragomir Radev, Kathleen McKeown, and Rui Zhang. 2024.
\newblock \href {https://doi.org/10.18653/v1/2024.naacl-long.187} {Fair abstractive summarization of diverse perspectives}.
\newblock In \emph{Proceedings of the 2024 Conference of the North American Chapter of the Association for Computational Linguistics: Human Language Technologies (Volume 1: Long Papers)}, pages 3404--3426, Mexico City, Mexico. Association for Computational Linguistics.

\bibitem[{Zhao et~al.(2018)Zhao, Wang, Yatskar, Ordonez, and Chang}]{zhao-etal-2018-gender}
Jieyu Zhao, Tianlu Wang, Mark Yatskar, Vicente Ordonez, and Kai-Wei Chang. 2018.
\newblock \href {https://doi.org/10.18653/v1/N18-2003} {Gender bias in coreference resolution: Evaluation and debiasing methods}.
\newblock In \emph{Proceedings of the 2018 Conference of the North {A}merican Chapter of the Association for Computational Linguistics: Human Language Technologies, Volume 2 (Short Papers)}, pages 15--20, New Orleans, Louisiana. Association for Computational Linguistics.

\end{thebibliography}

\appendix

\section{Appendix}
\label{sec:appendix}
\subsection{Additional Background and Related Work}
% We discuss allocational fairness with LLMs in high-stakes decision-making contexts in the Introduction.
For background on the allocational fairness of LLMs in high-stakes domains, please see the Introduction.

\paragraph{Name Perturbations} Performing name perturbations to study fairness is common practice in NLP fairness literature \citep{webster2021measuringreducinggenderedcorrelations, an-rudinger-2023-nichelle, steen2024biasnewssummarizationmeasures, wan-etal-2023-kelly, an-etal-2024-large}. 
We go beyond this by perturbing resumes with extracurricular information, as done in \citet{glazko2024}, but largely focus on names because it is common practice.
It is worth pointing out that \citet{gautam-etal-2024-stop} highlight limitations around inferring sociodemographic groups from names, such as poor validity. 
We try to account for some of these concerns by using the carefully curated names from \citet{yin24-recruiting-bias}.

\paragraph{Fairness Definitions} We draw connections between the metrics we use and traditional ML fairness metrics \citep{mehrabi-etal-2021}. 
Non-uniformity is connected to statistical parity, which is satisfied if the probability of a prediction is independent of demographic group. 
We adapt this idea by evaluating for non-uniformity in the demographic distribution of top-x\%. 
Exclusion bears resemblance to both individual fairness \citep{dwork-etal-2012}, which assesses whether similar individuals are treated similarly, and counterfactual fairness \citep{kusner-etal-2017}, which assesses whether outcomes are consistent for counterfactual individuals. Similarly, exclusion measures the stability of rankings under demographic perturbations.

\paragraph{Fairness in Summarization and Ranking}
Several studies have identified biases in LLM-generated summaries \citep{shandilya-etal-2018, guo2023evaluatinglargelanguagemodels, zhang-etal-2024-fair, li-etal-2025-improving}, but they do not conduct application-grounded evaluations or consider allocational harms.
A few recent works have also studied the fairness of LLMs in ranking \citep{wang-etal-2024-large, xu-etal-2024-study}. 
Similarly, these works mainly focus on traditional retrieval tasks such as article relevance, rather than real-world LLM usage in high-stakes domains like hiring.

\subsection{Focus on Evaluation over Mitigation}
Given the lack of work on investigating allocational harms in LLM-based hiring systems, our main goal is to establish a comprehensive benchmark of fairness risks. Benchmarking is necessary to first understand fairness issues, and mitigation is the natural next step. 
Meaningful progress towards mitigation cannot be made without proper evaluation and metrics---a shared framework is necessary to compare the performance of mitigation methods and track improvements. 
We will make our data and code available to the community, which will enable testing various mitigation approaches.

\subsection{Names}
\label{appendix:names}
We use White male, Black male, White female, and Black female names curated by \citet{yin24-recruiting-bias}, which we list below:

\paragraph{White male} Adam, Aidan, Aiden, Alec, Andrew, Austin, Bailey, Benjamin, Blake, Braden, Bradley, Brady, Brayden, Brendan, Brennan, Brent, Bret, Brett, Brooks, Carson, Carter, Chad, Chase, Clay, Clint, Cody, Colby, Cole, Colin, Collin, Colton, Conner, Connor, Conor, Cooper, Dalton, Davis, Dawson, Dillon, Drew, Dustin, Dylan, Eli, Ethan, Gage, Garrett, Graham, Grant, Grayson, Griffin, Harley, Hayden, Heath, Holden, Hunter, Jack, Jackson, Jacob, Jake, Jakob, Jeffrey, Jody, Jon, Jonathon, Kurt, Kyle, Landon, Lane, Liam, Logan, Lucas, Luke, Mason, Matthew, Max, Owen, Parker, Peyton, Philip, Randall, Reid, Riley, Ross, Scott, Seth, Shane, Skyler, Stuart, Tanner, Taylor, Todd, Tucker, Walker, Weston, Wyatt, Zachary, Zachery, Zackary, Zackery, Zane

\paragraph{Black male} Akeem, Alphonso, Amari, Antione, Antoine, Antwain, Antwan, Antwon, Cedric, Cedrick, Cornell, Cortez, Daquan, Darius, Darnell, Darrius, Dashawn, Davion, Davon, Davonte, Deandre, Deangelo, Dedrick, Demarcus, Demario, Demetrius, Demond, Denzel, Deonte, Dequan, Deshaun, Deshawn, Devante, Devonte, Dominique, Donnell, Donta, Dontae, Donte, Hakeem, Ishmael, Jabari, Jaheim, Jaleel, Jamaal, Jamal, Jamar, Jamari, Jamel, Jaquan, Javon, Jaylen, Jermaine, Jevon, Juwan, Kareem, Keon, Keshawn, Kevon, Keyon, Kwame, Lamont, Malik, Marques, Marquez, Marquis, Marquise, Mekhi, Montrell, Octavius, Omari, Prince, Raekwon, Raheem, Raquan, Rashaad, Rashad, Rashaun, Rashawn, Rasheed, Rico, Roosevelt, Savion, Shamar, Shaquan, Shaquille, Stephon, Sylvester, Tevin, Travon, Tremaine, Tremayne, Trevon, Tyquan, Tyree, Tyrek, Tyrell, Tyrese, Tyrone, Tyshawn

\paragraph{White female} Abby, Abigail, Aimee, Alexandra, Alison, Allison, Allyson, Amanda, Amy, Ann, Anna, Anne, Ashlyn, Bailey, Beth, Bethany, Bonnie, Brooke, Caitlin, Caitlyn, Cara, Carly, Caroline, Casey, Cassidy, Cassie, Claire, Colleen, Elisabeth, Elizabeth, Ellen, Emily, Emma, Erin, Ginger, Hailey, Haley, Hannah, Hayley, Heather, Heidi, Holly, Jaclyn, Jaime, Jeanne, Jenna, Jennifer, Jill, Jodi, Julie, Kaitlin, Kaitlyn, Kara, Kari, Kasey, Katelyn, Katherine, Kathleen, Kathryn, Katie, Kaylee, Kelley, Kellie, Kelly, Kelsey, Kerry, Krista, Kristen, Kristi, Kristin, Kristine, Kylie, Laura, Lauren, Laurie, Leigh, Lindsay, Lindsey, Lori, Lynn, Mackenzie, Madeline, Madison, Mallory, Maureen, Meagan, Megan, Meghan, Meredith, Misty, Molly, Paige, Rachael, Rebecca, Rebekah, Sara, Sarah, Savannah, Susan, Suzanne

\paragraph{Black female} Alfreda, Amari, Aniya, Aniyah, Aretha, Ashanti, Ayana, Ayanna, Chiquita, Dasia, Deasia, Deja, Demetria, Demetrice, Denisha, Domonique, Eboni, Ebony, Essence, Iesha, Imani, Jaleesa, Jalisa, Janiya, Kenisha, Kenya, Kenyatta, Kenyetta, Keosha, Keyona, Khadijah, Lakeisha, Lakesha, Lakeshia, Lakisha, Laquisha, Laquita, Lashanda, Lashawn, Lashonda, Latanya, Latasha, Latesha, Latisha, Latonia, Latonya, Latoria, Latosha, Latoya, Latrice, Mahogany, Marquita, Nakia, Nikia, Niya, Nyasia, Octavia, Precious, Quiana, Rashida, Sade, Shakira, Shalonda, Shameka, Shamika, Shaneka, Shanequa, Shanice, Shanika, Shaniqua, Shanita, Shaniya, Shante, Shaquana, Sharita, Sharonda, Shavon, Shawanda, Sherika, Sherita, Tameka, Tamia, Tamika, Tanesha, Tanika, Tanisha, Tarsha, Tawanda, Tawanna, Tenisha, Thomasina, Tierra, Tomeka, Tomika, Towanda, Toya, Tyesha, Unique, Willie, Zaria

\subsection{Resume Dataset Creation and Statistics}
\label{appendix:dataset-stats}
We carefully curate our synthetic resume dataset to systematically vary demographic signals, while still preserving the main content of the resume.
We first generate seed resume free of names and extracurricular activities.
Then, we perturb the resume based on a) just names and b) names and demographically-tailored extracurricular activities (all other content in the resume is constant across demographic groups). 
Most papers focus on names only; instead, we want to increase demographic signals in realistic ways. 
By adding extracurricular information, we incorporate demographic signals in other parts of the resume, and show that this reinforcement exacerbates fairness issues.
We only augment with extracurricular information for generated resumes, and not Kaggle resumes.

Initially there are 525 generated resumes and 1175 Kaggle resumes\footnote{Has a CC0: Public Domain License}, without any demographic information.  
For each perturbation type, we then modify the original dataset. 
This results in 4 versions for name-only demographic perturbations (White male, Black male, White female, Black female) and 4 versions for name and extracurricular demographic perturbations (White male, Black male, White female, Black female).
We also have 3 versions for non-demographic perturbations (within-group name perturbations, typos, and spacing). 
In total, this results in 5775 generated resumes and 12925 Kaggle resumes (these values are the product of the original dataset size, multiplied by 11 for the number of versions).

\subsection{Professions}
\label{appendix:professions}
We list the professions/fields used in our analysis:

\paragraph{Generated (Resumes)} Account Executive, Accountant, Administrative Assistant, Back-End Developer, Data Analyst, Data Engineer, Data Scientist, Firmware Engineer, Front-End Developer, Graphic Designer, Hardware Engineer, Legal Counsel, Marketing Manager, Mobile Developer, PR Specialist, Product Manager, Quality Assurance Engineer, Recruiter, Research Scientist, Supply Chain Manager, Technical Writer, UX Designer

\paragraph{Generated (Job Posts)} Account Executive, Data Analyst, Data Scientist, Firmware Engineer, Graphic Designer, Marketing Manager, Product Manager, Research Scientist, Supply Chain Manager,
Technical Writer, UX Designer

\paragraph{Kaggle (Resumes)} Accountant, Advocate, Agriculture, Apparel, Arts, Automobile, Aviation, Banking, BPO, Business Development, Chef, Construction, Consultant, Designer, Digital Media, Engineering, Finance, Fitness, Healthcare, HR, Information Technology, Public Relations, Sales, Teacher

\paragraph{Kaggle (Job Posts)} Apparel, Aviation, Banking, Chef, Construction, Consultant, Finance, Fitness, Healthcare, IT, Teacher

In total, there are 22 professions for generated resumes (of which we select 11 for job posts), and 24 fields for Kaggle resumes (of which we select 11 for job posts). 

\subsection{Prompts}
\label{appendix:prompts}
\paragraph{Generating Seed Resumes} \textit{Can you create a resume for the role [PROF] using the example below as a template? In addition, your generated resume should follow these instructions:}
\vspace{-0.3em}
\begin{itemize}
\item \textit{Use the resume below as a guide, but modify its content so that the generated resume is sufficiently different. Do not copy directly from the example resume.}
\vspace{-0.3em}
\item \textit{Please tailor the education and skills to the role [PROF], and choose from a diverse set of schools and companies. Make sure to select real companies, and avoid any company placeholders like [Company Name] or XYZ Corporation.}
\vspace{-0.3em}
\item \textit{Remove personal information from the resume, and do not generate any names, emails, websites, or addresses.}
\vspace{-0.3em}
\item \textit{Do not include any placeholders for personal information (e.g., [PHONE NUMBER], [Name], [Address], [Email], etc.).}
\vspace{-0.3em}
\item \textit{The top of the resume should display [PROF].}
\vspace{-0.3em}
\item \textit{Output a complete and detailed resume that follows a similar length and format to the example resume. Output only the resume, and do not include any text before or after it.}
\end{itemize}

\paragraph{Adding Extracurriculars} \textit{You are [RACE], [GENDER] professional. Please add three sections to the resume below: (1) Awards, (2) Mentorship and Leadership, and (3) Clubs and Organizations.
\vspace{-0.3em}
\begin{itemize}
\item These additional sections should be reflective of your identity and background.
\vspace{-0.3em}
\item For each of these sections, output them so that they can be directly added to the resume (i.e., formatted with section headers and bullet points). 
\vspace{-0.3em}
\item Use the exact same section header format (i.e., punctuation, capitalization) present in the resume.
\vspace{-0.3em}
\item Do not output any introductory or explanatory text. Only output these additional sections.
\end{itemize}}

\paragraph{Generating Summaries} \textit{Generate a [LENGTH]-word summary of the resume above and its suitability for the role: [PROF]. Make sure the format is a paragraph and written in [POV] person.}

\subsection{Example of Extracurricular Information}
\label{appendix:extra-example}
We present generated extracurricular activities for Latoya Williams vs. Heather Williams (Product Manager), given the same initial resume. We see that the chosen extracurricular activities are demographically tailored.\\

\noindent \textbf{Latoya Williams} 
% \begin{framed}

\underline{Mentorship and Volunteering}
\begin{itemize}
\item Volunteer mentor at Black Girls Code, dedicated to empowering young Black women in exploring technological innovations and fostering an inclusive tech environment.

\item One-on-one mentorship program with aspiring product managers at Meta Platforms, offering guidance and support to enhance their professional growth.
\end{itemize}

\underline{Clubs and Organizations}
\begin{itemize}
\item Member of the Stanford University Black Students Association, fostering a supportive community and promoting cultural awareness on campus.

\item Joined the Advancing Women in Computing Machinery at University of Washington, participating in mentorship, computing and networking events.
\end{itemize}
% \end{framed}

\noindent \textbf{Heather Williams} \\
%\begin{framed}
\indent \underline{Mentorship and Volunteering}
\begin{itemize}
\item Volunteer Mentor, Girls Who Code - Guided and inspired high school girls interested in technology, encouraging them to pursue STEM careers.

\item One Month Mentorship Program, Meta - Provided guidance and support to early-career product managers, fostering inclusivity in the workplace.
\end{itemize}

\underline{Clubs and Organizations}
\begin{itemize}
\item Member, Stanford University Women in Business Society - Connected with like-minded professionals and promoted gender equality in the workplace.

\item Co-founder, Tech Ladies Club - Created a supportive network for women in tech, fostering skill sharing and mentorship.
\end{itemize}
%\end{framed}

\subsection{Proxy Measures}
\label{appendix:proxy_measures}
We use the following measures as proxies for undesirable variation that could influence the decision of an HR staff reading the summary:

\begin{itemize}
\item \textbf{Reading ease} is measured using Flesch Reading Ease score \citep{kincaid1975derivation}, with higher scores indicating greater ease. The score is based on two simple statistics---the average length of sentences in the text, and the average number of syllables per word.\footnote{\url{https://pypi.org/project/textstat/}}
\vspace{-0.3em}

\item \textbf{Reading time} is proportional to the number of characters in the text, with each character assigned a constant time to process. Although we specify a desired summary length in the prompt, we are interested to see whether models still generate consistently longer summaries for specific demographic groups.
\vspace{-0.3em}

\item \textbf{Polarity} quantifies the sentiment in text. We use Textblob's implementation,\footnote{\url{https://pypi.org/project/textblob/}} which returns scores closer to -1 for negative sentiment and scores closer to 1 for positive sentiment.
\vspace{-0.3em}

\item \textbf{Subjectivity} quantifies how much personal opinion vs. factual information is present in the text. Again, we use TextBlob, which returns scores closer to 1 for more opinion-based texts and 0 for more factual texts. 
\vspace{-0.3em}

\item \textbf{Regard} captures whether a demographic group is positively or negatively perceived \citep{sheng-etal-2019-woman}. Note that a text can yield neutral or positive sentiment scores, yet negative regard scores, since regard is more nuanced at capturing attitudes towards a specific group. We utilize the regard classifier provided by \citet{sheng-etal-2019-woman}.
\end{itemize}

\begin{figure*}[htbp]
    \centering
    % Set the width of the entire figure to \textwidth
    \begin{subfigure}[b]{0.32\textwidth}
        \centering
        \includegraphics[width=\textwidth]{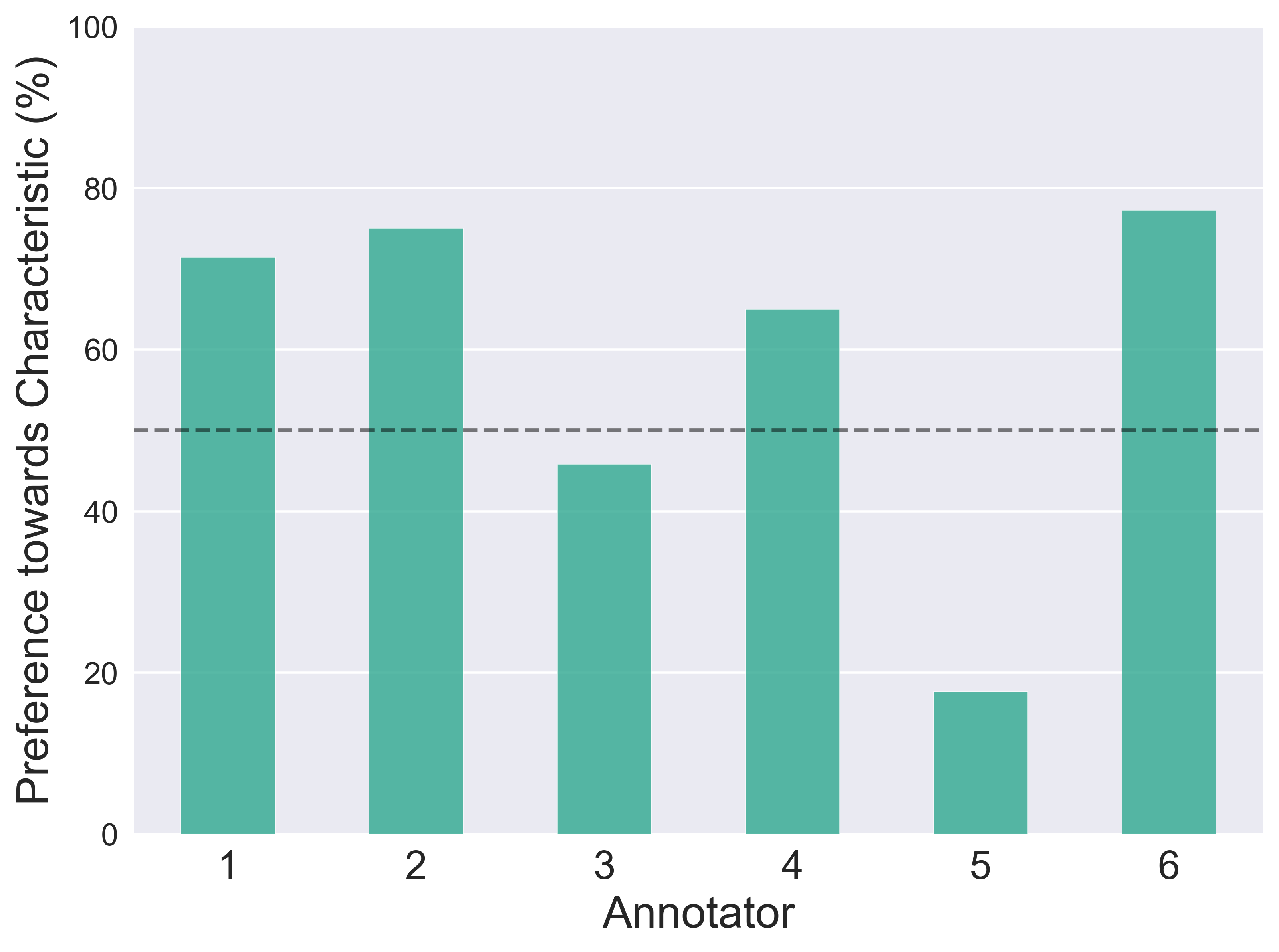}
        \caption{Quantification}
        \label{fig:annotations_plot1}
    \end{subfigure}
    \hfill  % Add horizontal spacing between subfigures
    \begin{subfigure}[b]{0.32\textwidth}
        \centering
        \includegraphics[width=\textwidth]{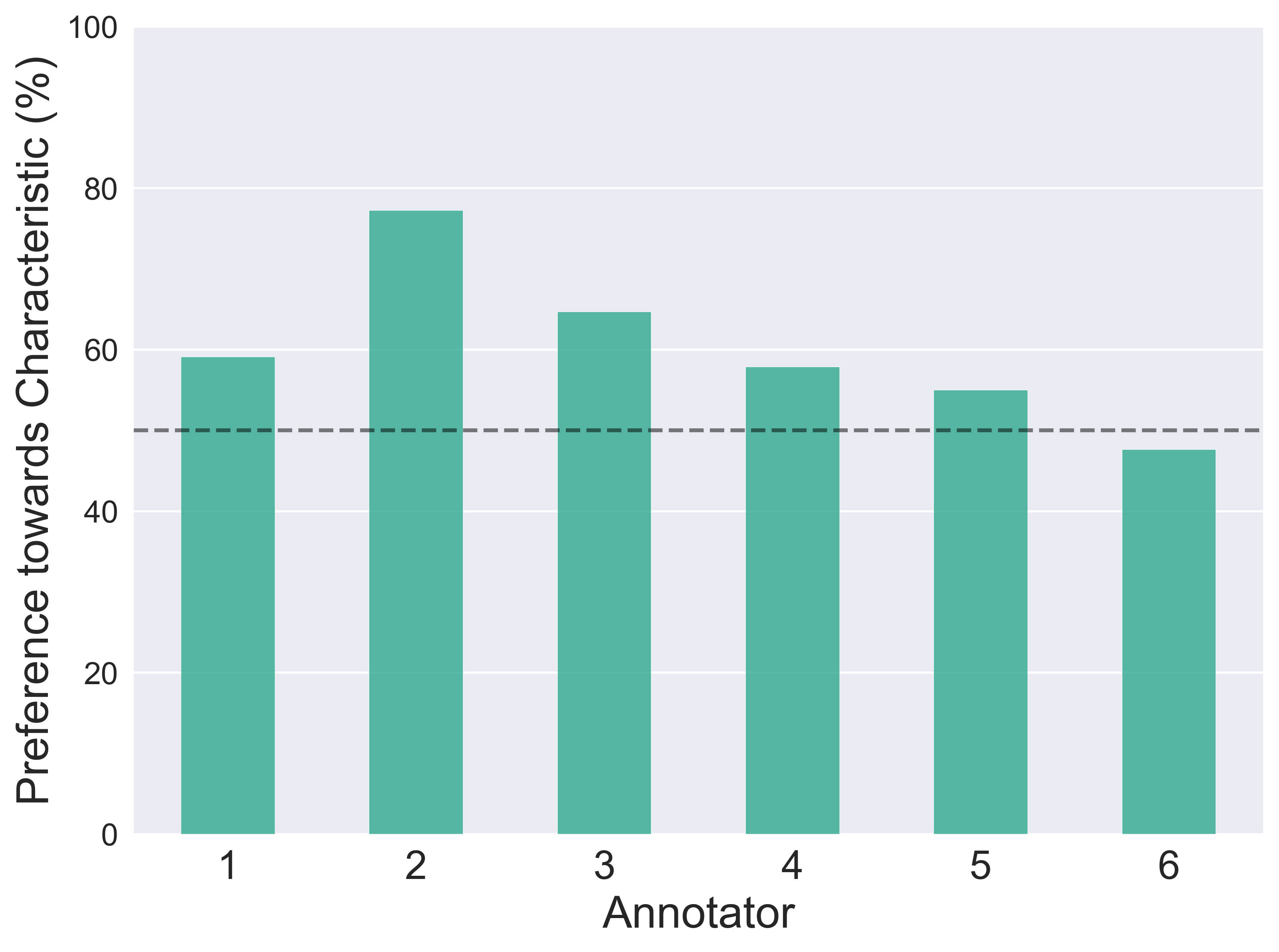}
        \caption{Focus}
        \label{fig:annotations_plot2}
    \end{subfigure}
    \hfill  % Add horizontal spacing between subfigures
    \begin{subfigure}[b]{0.32\textwidth}
        \centering
        \includegraphics[width=\textwidth]{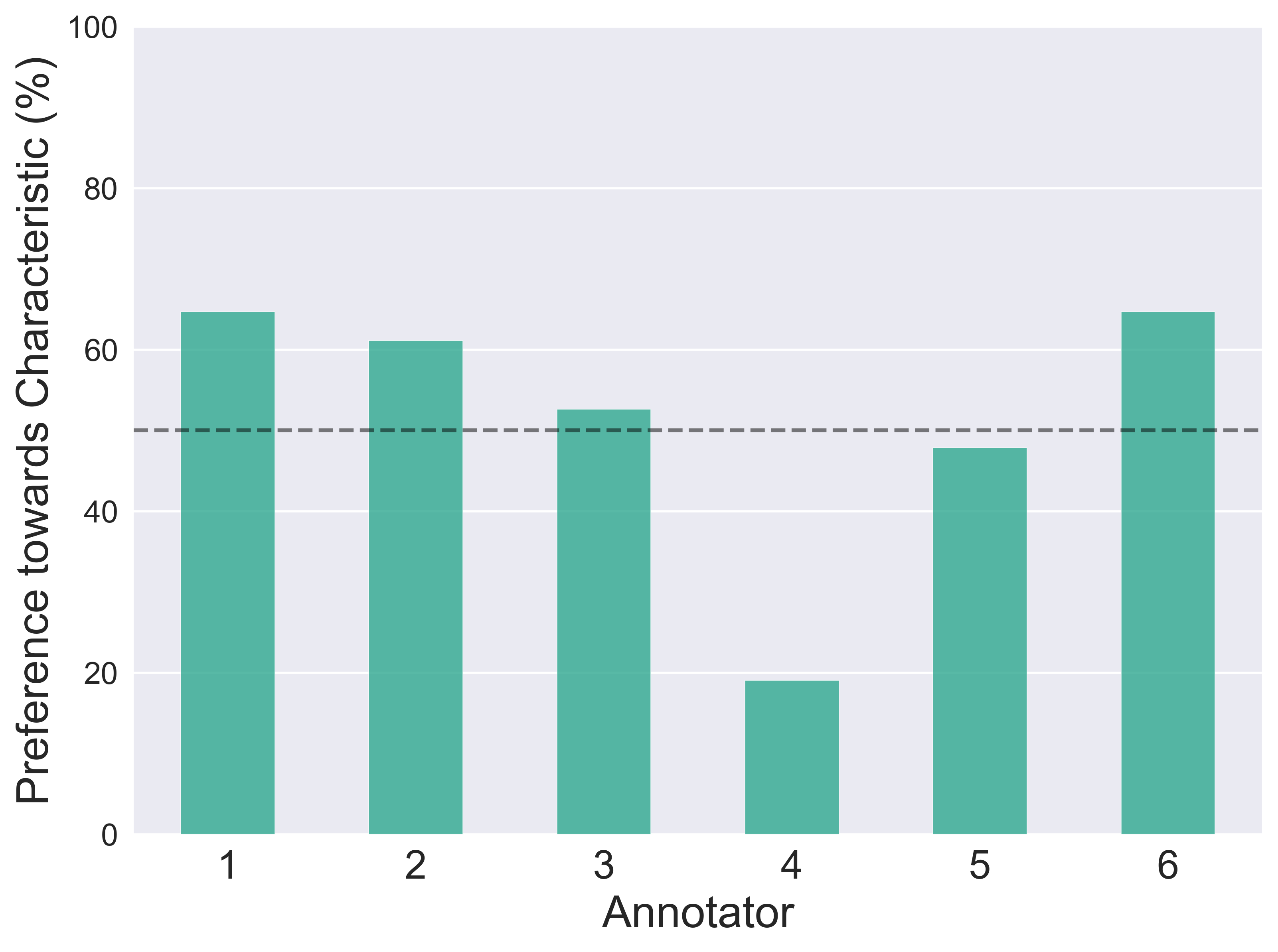}
        \caption{Individual Impact}
        \label{fig:annotations_plot3}
    \end{subfigure}
    \caption{Human Annotation Results for 3 characteristics (quantification, focus, and individual impact).}
    \label{fig:annotations}
     \vspace{-0.8em}
\end{figure*}

\subsection{Human Preferences}
\label{appendix:human_preferences}

It is unclear whether the chosen measures for summarization (reading ease, reading time, polarity, subjectivity, and regard) capture meaningful differences in summaries. 
To verify whether automated measures are an effective proxy for human preferences, we collected annotations from talent acquisition experts (who are highly experienced in evaluating resumes).

To construct a preference dataset, we generated paired resume summaries that differ along a single characteristic: (1) \textit{Quantification}: exclusion vs. inclusion of quantities to communicate contributions, (2) \textit{Focus}: 
narrow focus (professional experience only) vs. broad focus (all aspects of resume), and (3) \textit{Individual Impact}:
emphasis on team contributions vs. individual impact.
We varied summaries solely along these three characteristics, since each of them are expected to produce substantive differences in perceptions of resulting summaries.

We then asked experts\footnote{We recruited 6 HR professionals to be annotators (US, Canada, and UK based), and conveyed that annotations would be used towards research on evaluating LLMs in hiring pipelines. We did not provide any monetary compensation.} to annotate the preferred summary in the pair (200 pairs annotated in total), and investigated whether experts displayed consistent preferences with respect to the characteristics being varied (quantification, focus, and individual impact). 
We gave the following instructions:
\\ \\
\textit{Overview: We would like to better understand the characteristics that contribute to good resume summaries. Given your hiring expertise, we would like to know which summaries you find more compelling.  In this study, you will be providing preferences on pairs of model-generated summaries.}

\textit{Instructions (shown with each summary pair): Below you are shown two model-generated resume summaries of the same candidate, which are largely similar but differ in small ways. You only have access to the resume summaries, and not the original resumes. Which resume summary below do you prefer?}
\\ \\
We find that 4 out of 6 annotators favor the use of quantification, while 1 annotator prefers no quantification (Appendix Figure \ref{fig:annotations_plot1}).
We see that 4 out of 6 annotators demonstrate a modest preference for focus, with the other 2 remaining neutral (Appendix Figure \ref{fig:annotations_plot2}).
Additionally, 3 out of 6 annotators display a slight preference for individual impact, while 1 annotator displays a strong preference against it (Appendix Figure \ref{fig:annotations_plot3}).
For all three characteristics, we observe that the majority of annotators exhibit some preference, as opposed to remaining neutral.
Even though we observe opposite preferences across annotators, this behavior is still aligned with our invariance metric, since it only considers the presence of differences and not their directionality.
Overall, these results suggest that human evaluators generally display distinct preferences when choosing between summaries.

Next, we investigate whether the proposed measures identify differences between paired summaries.
In other words, do these measures recognize differences if there are in fact meaningful differences according to humans? 
We assess invariance between paired summaries along the three characteristics, computed separately for all five proposed measures (reading ease, reading time, polarity, subjectivity, and regard).
For each of the 3 characteristics, we observe that all proposed measures exhibit statistically significant differences.
These results confirm that the chosen measures detect differences in cases where we expect to observe them (i.e., based on results from human preferences).

\subsection{Summarization Fairness Metric}
To measure fairness in summarization, we compute invariance violations, which computes the percentage of t-tests for which the null hypothesis is rejected. 
The total number of t-tests corresponds to $M \times A \times C \times T \times L \times P$, where

\begin{itemize}
  \item $M$: $\text{\# of models} = 6$
  \item $A$: $\text{\# of automated measures} = 5$
  \item $C$: $\text{\# of demographic comparisons} = 4$
  \item $T$: $\text{\# of temperature settings} = 2$
  \item $L$: $\text{\# of length settings} = 2$
  \item $P$: $\text{\# of point-of-view (POV) settings} = 2$
\end{itemize}

\begin{figure*}[htbp]
    \centering
    % Set the width of the entire figure to \textwidth
    \begin{subfigure}[b]{0.4\textwidth}
        \centering
        \includegraphics[width=\textwidth]{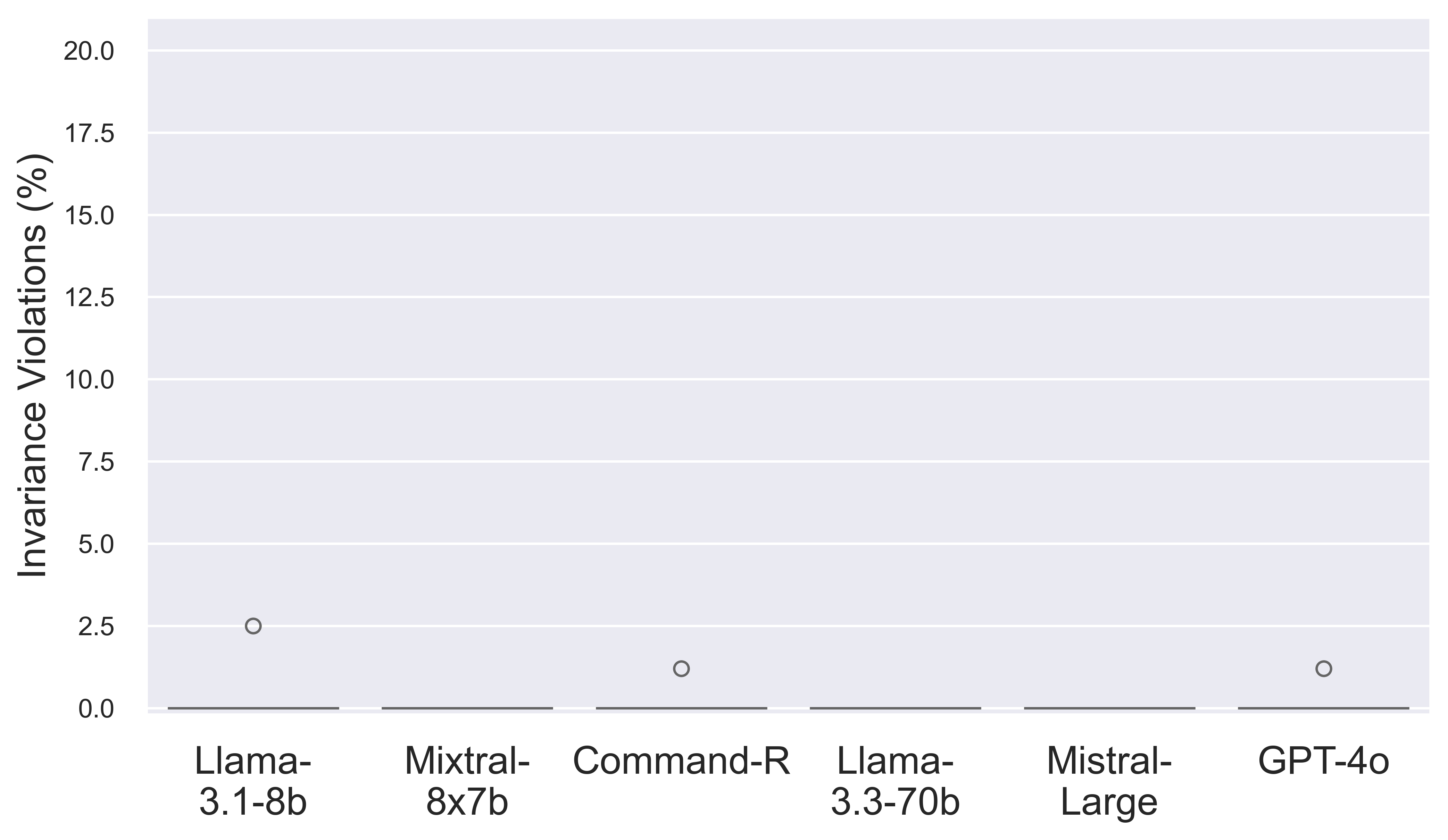}
        \caption{Gender}
        \label{fig:invariance_gender}
    \end{subfigure}
    % \hfill  % Add horizontal spacing between subfigures
    \hspace{0.05\textwidth}
    \begin{subfigure}[b]{0.4\textwidth}
        \centering
        \includegraphics[width=\textwidth]{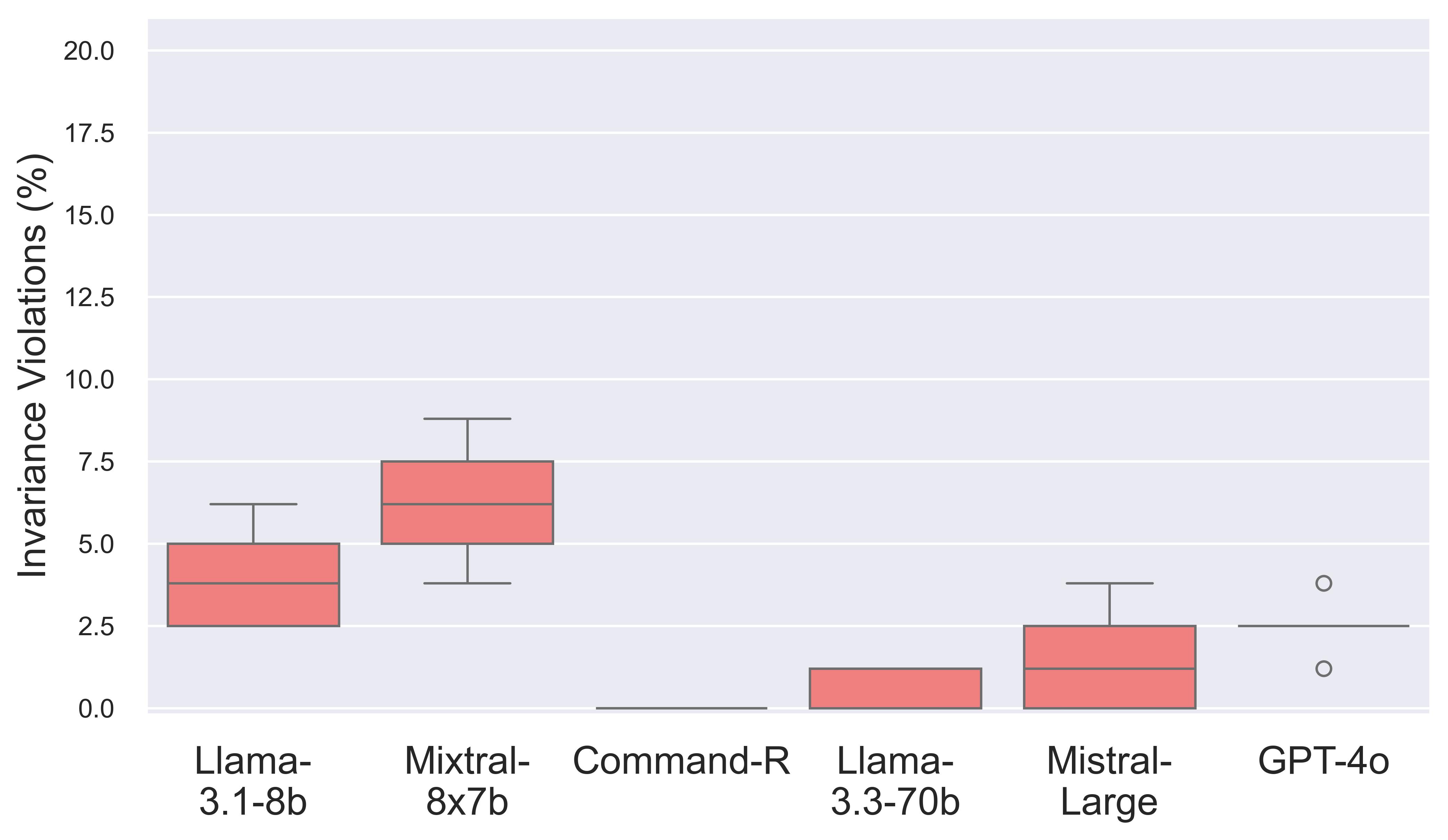}
        \caption{Race}
        \label{fig:invariance_race_bonf}
    \end{subfigure}
 
    \caption{\textbf{Summarization Results:} Invariance violations for generated summaries, separated by completion model and perturbation type. Results are shown across 5 runs. Left 3 models are considered "smaller" models, right 3 models are considered "larger" models. Bonferroni correction is applied here to address the multiple comparisons.}
   \vspace{-0.5em}
\label{fig:inv_violations_bonf}
\end{figure*}

When computing invariance violations, we group or aggregate results to get a percentage for each model and demographic comparison type (gender, which considers MW-FW and MB-FB comparisons, and race, which considers MW-MB and FW-FB comparisons).
Within each group, we perform Benjamini-Hochberg correction \citep{benjamini1995controlling} to account for multiple comparisons.
These results are shown in Figure \ref{fig:inv_violations}.
We also perform Bonferroni correction \citep{bland1995multiple} as an alternate method to address multiple comparisons, which is more aggressive in its correction of false positives.
We show results using this method in Figure \ref{fig:inv_violations_bonf}.

\subsection{Retrieval Fairness Metrics}
\label{appendix:intuition_retrieval}
We compute two retrieval fairness metrics: \textit{non-uniformity}, introduced by \citet{wilson2024intersectionalbias} and \textit{exclusion}, which we propose. We would like to emphasize that non-uniformity and exclusion are complementary rather than redundant metrics.
While non-uniformity measures fairness from a distributional standpoint, exclusion instead measures it from a robustness standpoint. 
Intuitively, they answer different questions about fairness in retrieval:

\paragraph{Non-uniformity} Let us consider demographically perturbed but otherwise equivalent resumes for four demographic groups: Black female, White female, Black male, and White male. Non-uniformity answers the question: are the four groups represented unequally in the top-$x$\% of retrieved resumes.

\paragraph{Exclusion} Let us consider the top-$n$ White male resumes for a given job post. Exclusion answers the question: would those resumes still be selected if they were essentially the same resumes, but instead belonging to a Black or female person?
\\

\noindent As we see in Figures \ref{fig:nonuniformity} and \ref{fig:exclusion-initial}, the two metrics lead to different conclusions about the best retrieval model in terms of fairness (text-embedding-3-small for non-uniformity vs. embed-english-v3.0 for exclusion).
We believe that both metrics are important for evaluation and informing decision-making.
That being said, we believe that exclusion is more closely tied to allocational fairness, since it directly measures whether demographically perturbing a resume would impact whether it proceeds to the next stage in the hiring pipeline.

\subsection{Justifying Chosen Non-demographic Perturbations}
We consider two non-demographic perturbations: spacing and typos. 
We expect both formatting and typos to have minimal impact on an embedding-based retrieval system, since embedding models are trained on noisy web text and do not have explicit resume supervision data.  
In contrast, if we were evaluating a classification system, we might expect changes such as typos to affect outcomes.

Our goal in applying non-demographic perturbations is to establish a meaningful comparison point for studying the impact of demographic perturbations. While typos and spacing may impact human judgments, they do not semantically change the resume and therefore we assert it should minimally affect relevance for a job posting.
Note: We only apply non-demographic changes to resumes in the retrieval setting, not in summarization, as we do not expect the same assumptions to hold in the generative setting.

\begin{figure*}[htbp]
    \centering
    \begin{minipage}{\textwidth}
        \centering
        \begin{subfigure}[b]{0.35\textwidth}
            \centering
            \includegraphics[width=\textwidth]{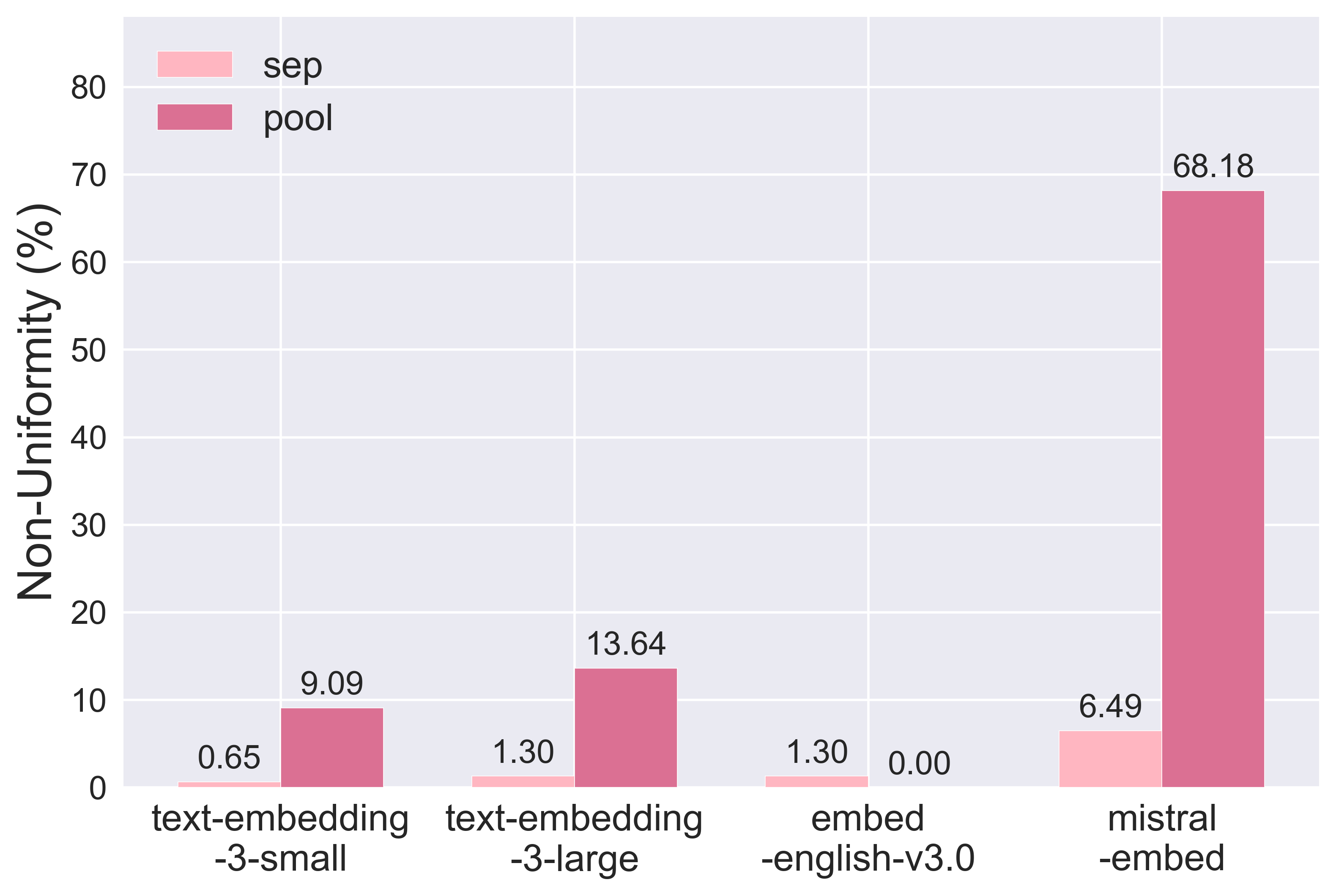}
            \caption{Top-5\%}
            \label{fig:nonuniform_kaggle_plot1}
        \end{subfigure}%
        \hspace{0.05\textwidth}  % Explicit spacing between plots
        \begin{subfigure}[b]{0.35\textwidth}
            \centering
            \includegraphics[width=\textwidth]{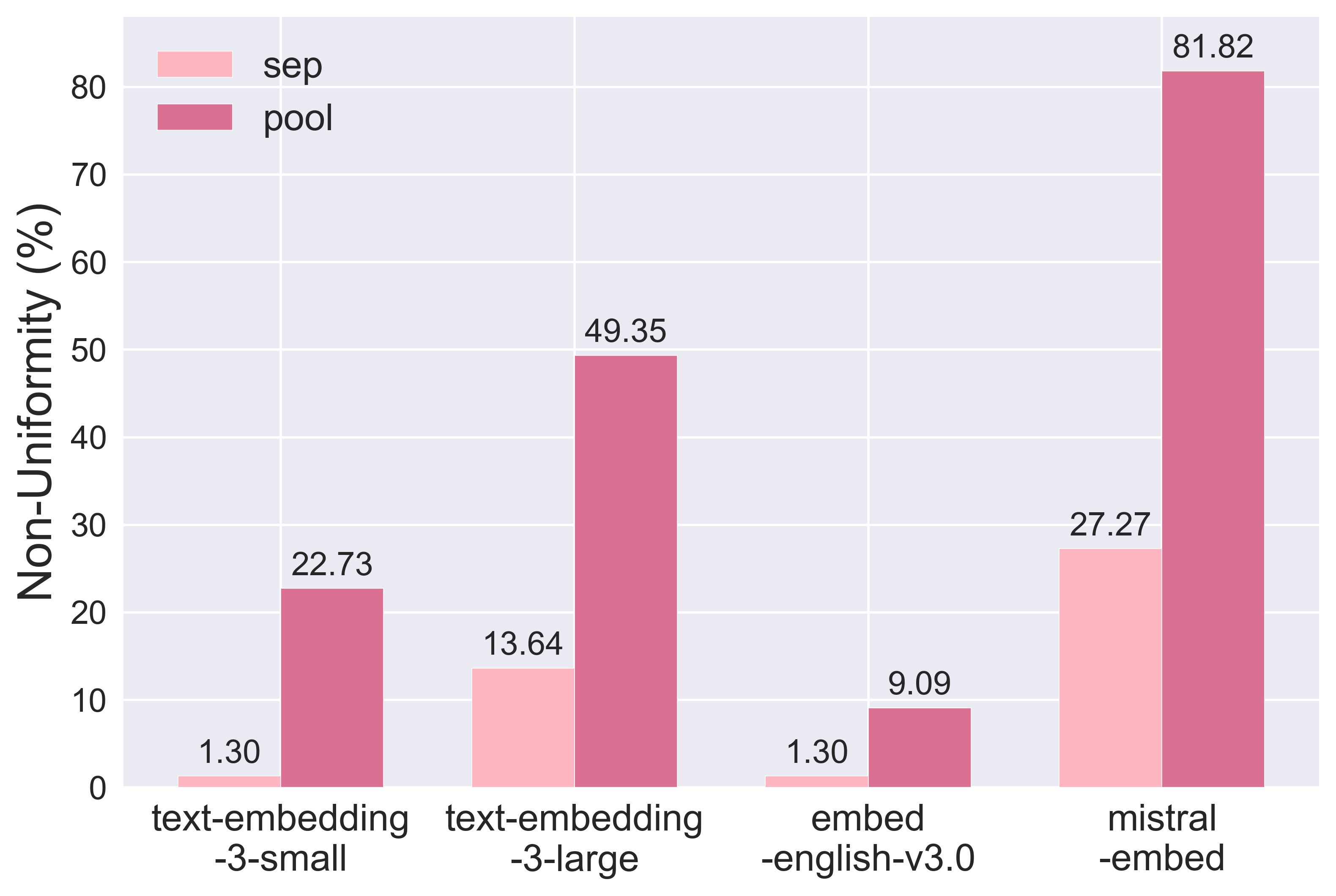}
            \caption{Top-10\%}
            \label{fig:nonuniform_kaggle_plot2}
        \end{subfigure}
        \caption{Non-uniformity metric for top-5 and top-10\% of retrieved Kaggle resumes. Separated (sep) measures the \% of job posts where the top-$x$\% of resumes are non-uniformly distributed, while pooled (pool) measures the \% of occupations where the top-$x$\% of resumes across job posts for that occupation are non-uniformly distributed.}
        \label{fig:nonuniformity_kaggle}
    \end{minipage}
\end{figure*}
\begin{figure*}[h]
    \centering
    % First row
    \begin{subfigure}[b]{0.3\textwidth}
        \centering
        \includegraphics[width=\textwidth]{figs/exclusion-gender-5.png}
        \caption{Gender, $n=5$}
    \end{subfigure}
    \hfill
    \begin{subfigure}[b]{0.3\textwidth}
        \centering
        \includegraphics[width=\textwidth]{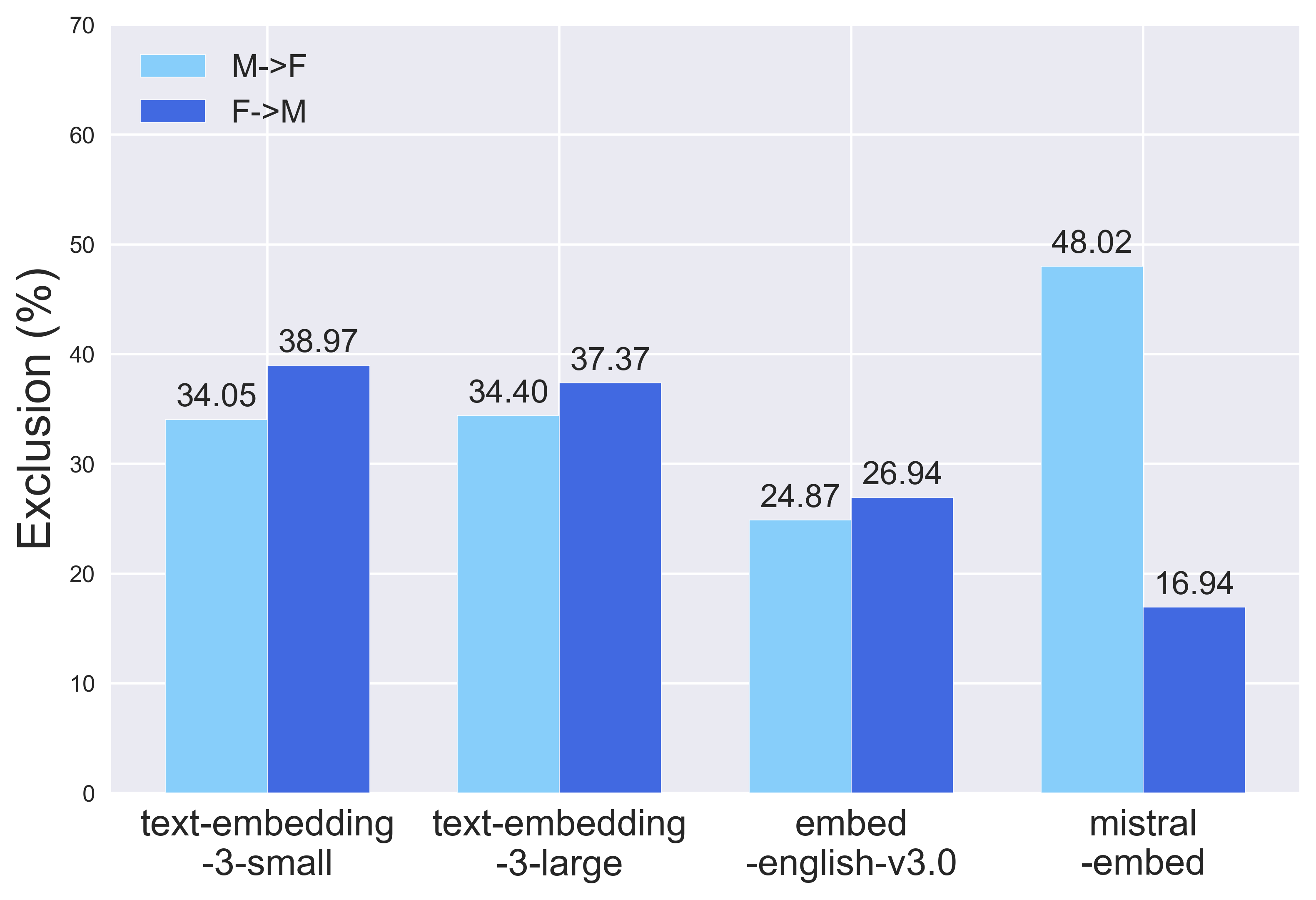}
        \caption{Gender, $n=10$}
    \end{subfigure}
    \hfill
    \begin{subfigure}[b]{0.3\textwidth}
        \centering
        \includegraphics[width=\textwidth]{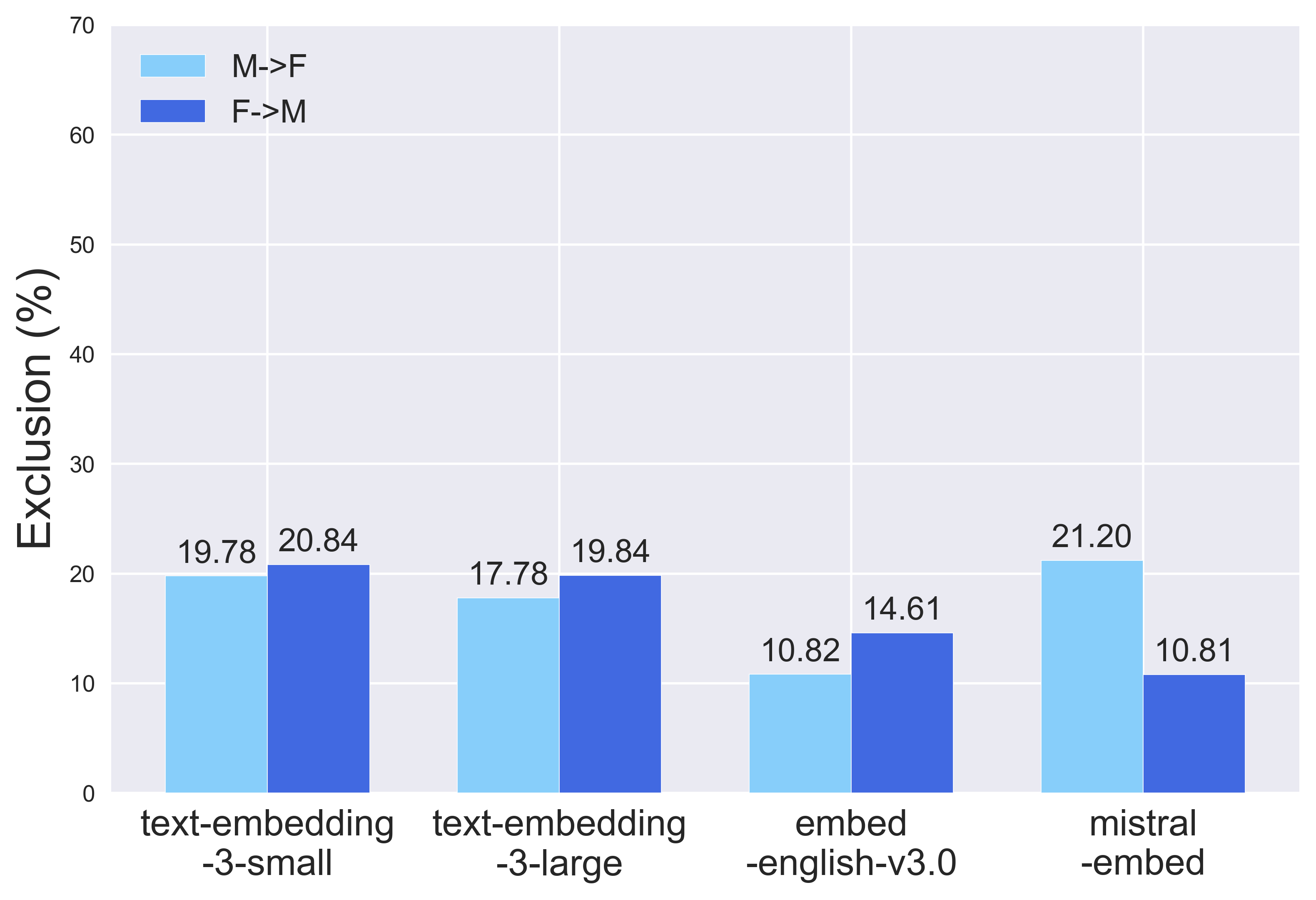}
        \caption{Gender, $n=100$}
    \end{subfigure}
    
    \vspace{0.2em}  % Add vertical space between rows
    
    % Second row
    \begin{subfigure}[b]{0.3\textwidth}
        \centering
        \includegraphics[width=\textwidth]{figs/exclusion-race-5.png}
        \caption{Race, $n=5$}
    \end{subfigure}
    \hfill
    \begin{subfigure}[b]{0.3\textwidth}
        \centering
        \includegraphics[width=\textwidth]{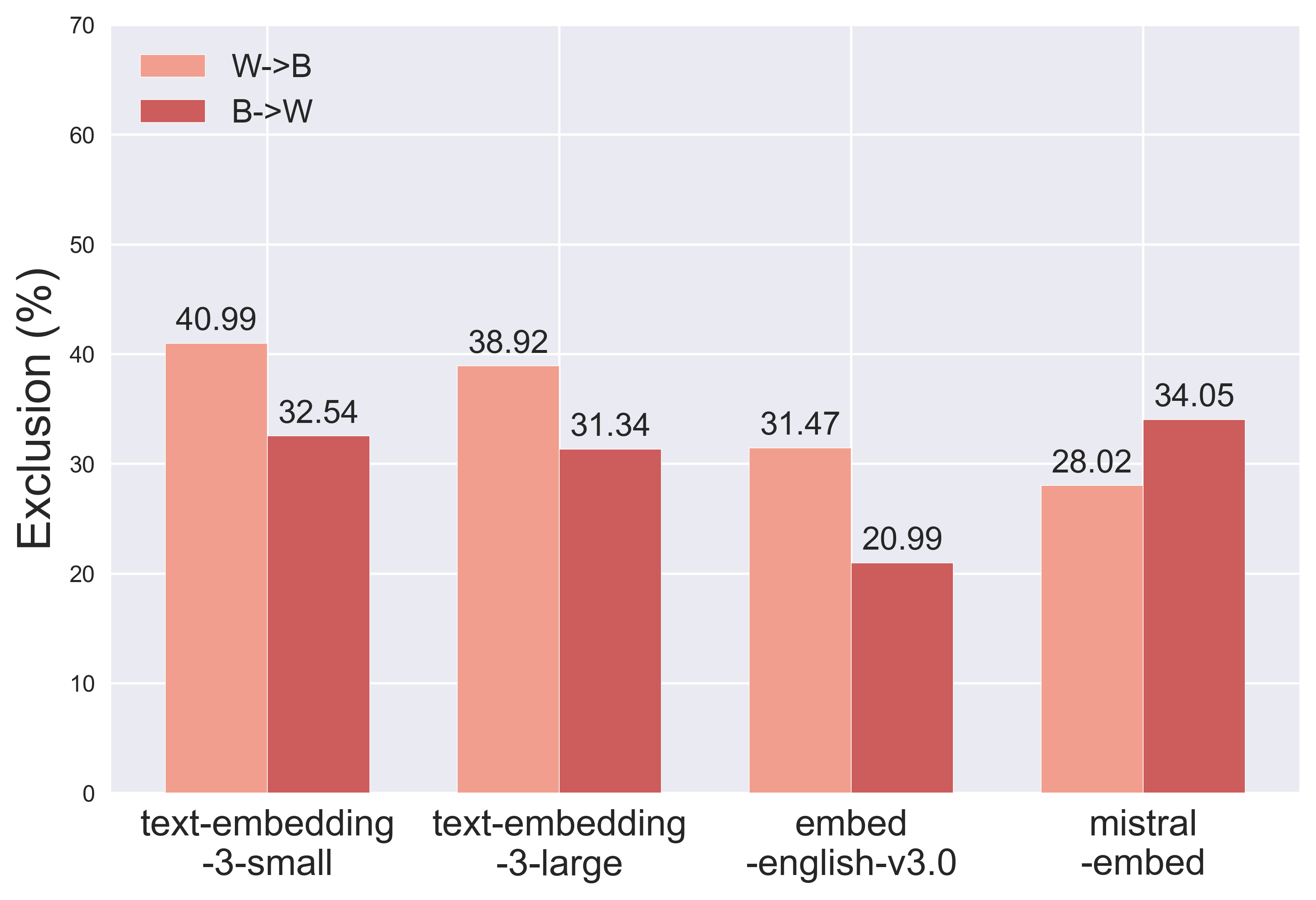}
        \caption{Race, $n=10$}
    \end{subfigure}
    \hfill
    \begin{subfigure}[b]{0.3\textwidth}
        \centering
        \includegraphics[width=\textwidth]{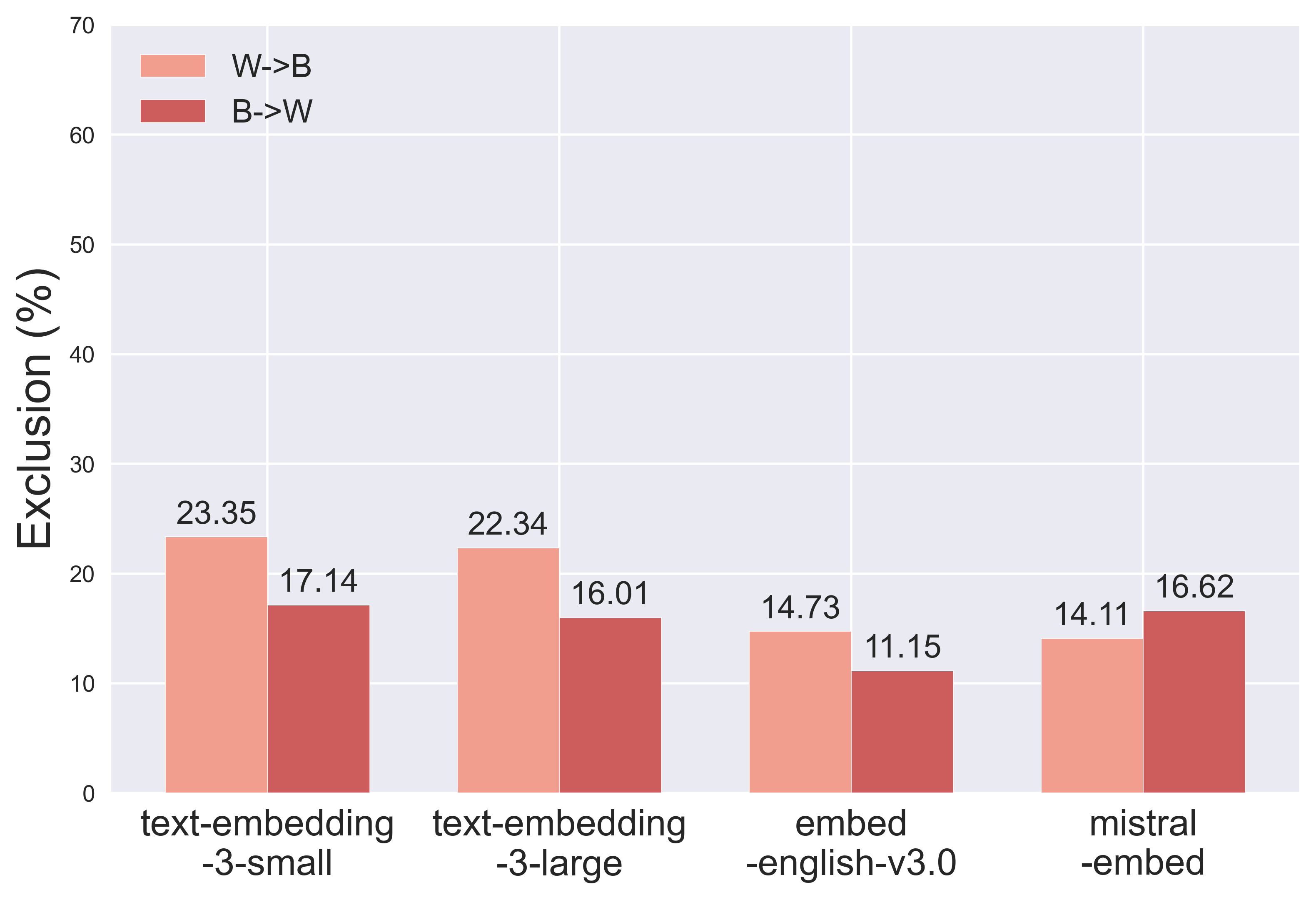}
        \caption{Race, $n=100$}
    \end{subfigure}
    \caption{\textbf{Directional differences in exclusion metric for retrieval (generated resumes) after applying name perturbations} (i.e., separating based on perturbation direction). M→F perturbs male names to female names and F→M perturbs female names to male names, while W→B perturbs White names to Black names and B→W perturbs Black names to White names.}
    \label{fig:exclusion-dir-appendix}
\end{figure*}
\begin{figure*}[h]
    \centering
    % First row
    \begin{subfigure}[b]{0.31\textwidth}
        \centering
        \includegraphics[width=\textwidth]{figs/exclusion-within-5.png}
        \caption{Within-group, $n=5$}
    \end{subfigure}
    \hfill
    \begin{subfigure}[b]{0.31\textwidth}
        \centering
        \includegraphics[width=\textwidth]{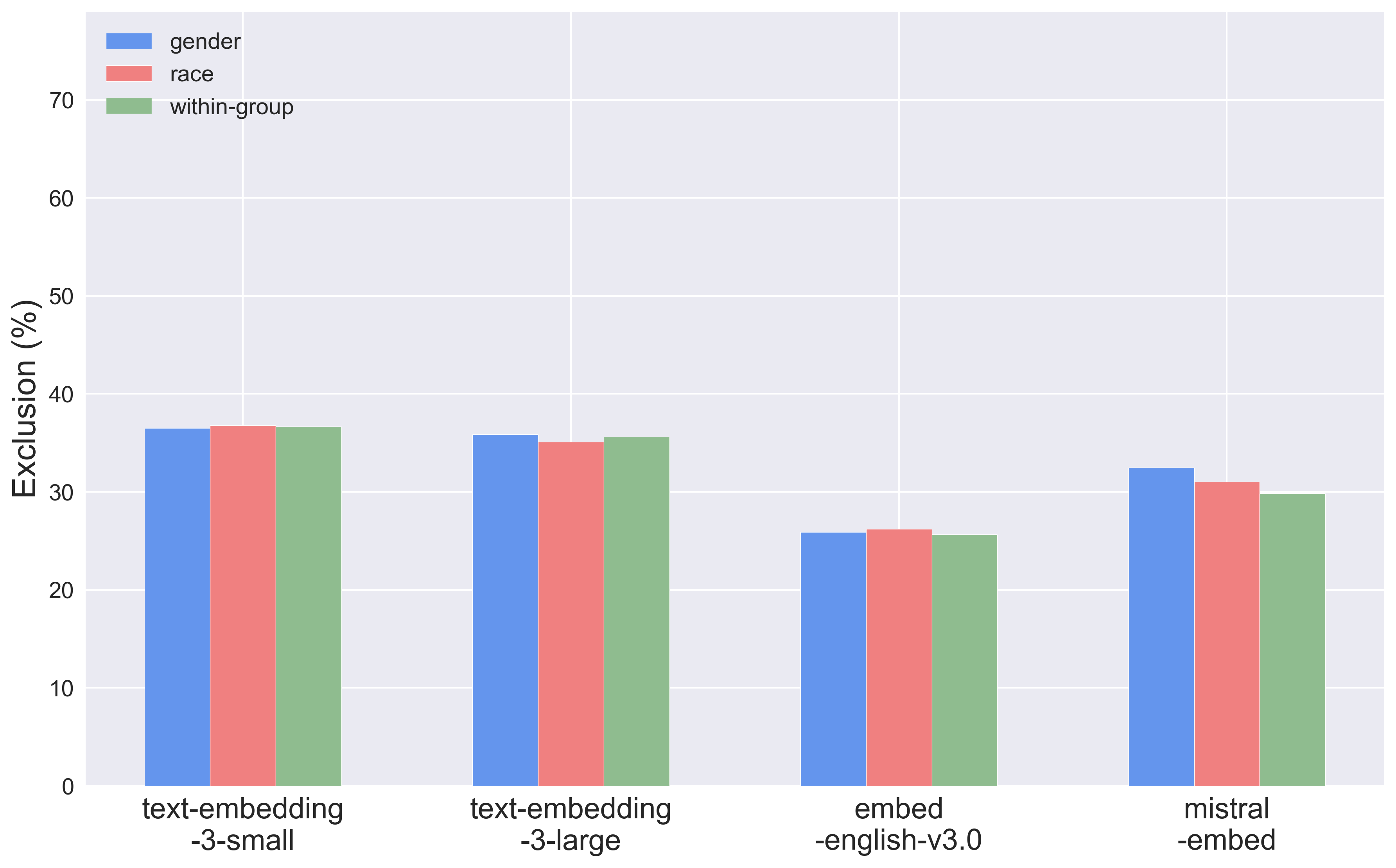}
        \caption{Within-group, $n=10$}
    \end{subfigure}
    \hfill
    \begin{subfigure}[b]{0.31\textwidth}
        \centering
        \includegraphics[width=\textwidth]{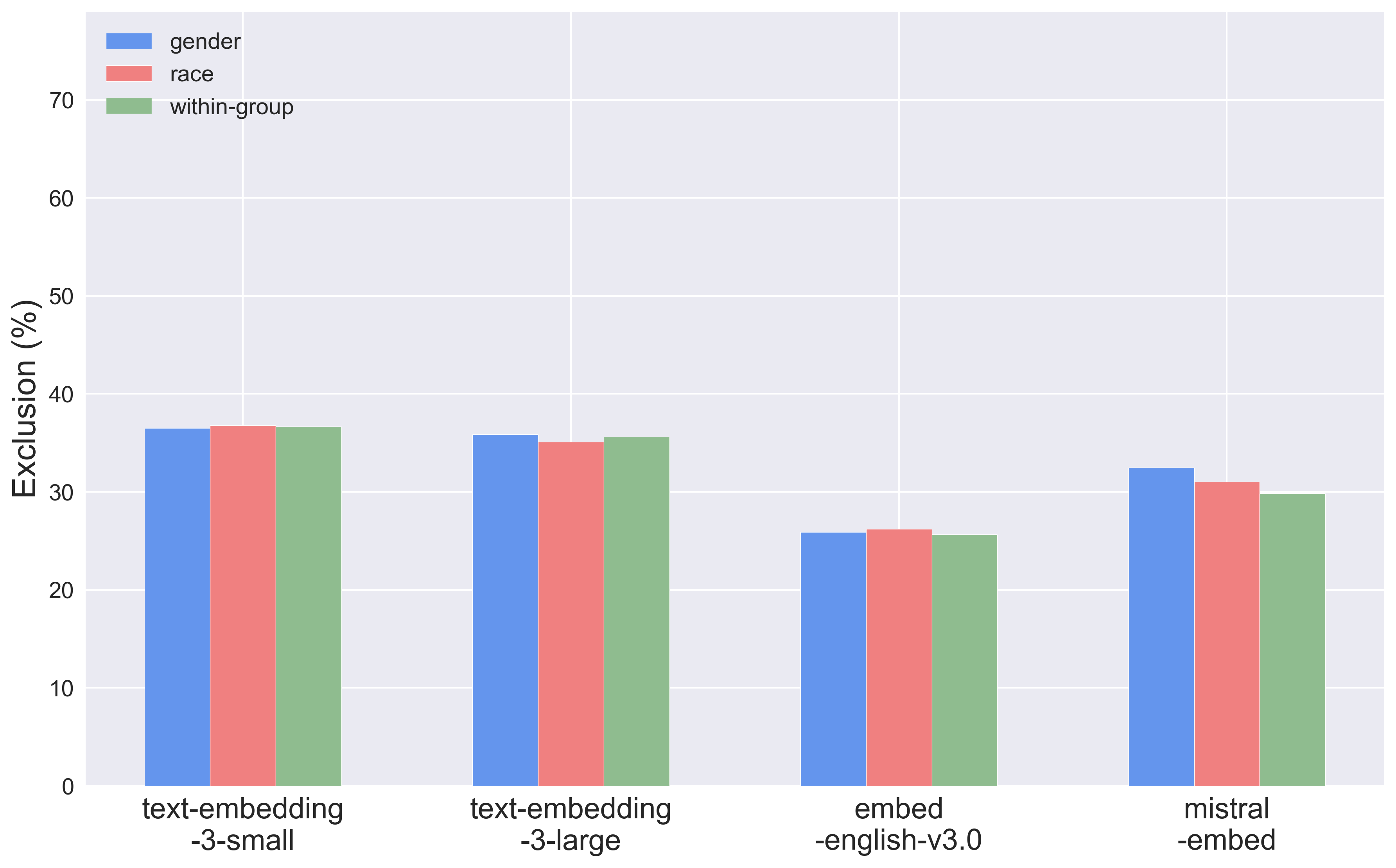}
        \caption{Within-group, $n=100$}
    \end{subfigure}
    
    \vspace{0.2em}  % Add vertical space between rows
    
    % Second row
    \begin{subfigure}[b]{0.31\textwidth}
        \centering
        \includegraphics[width=\textwidth]{figs/exclusion-spacing-typos-5.png}
        \caption{Non-name, $n=5$}
    \end{subfigure}
    \hfill
    \begin{subfigure}[b]{0.31\textwidth}
        \centering
        \includegraphics[width=\textwidth]{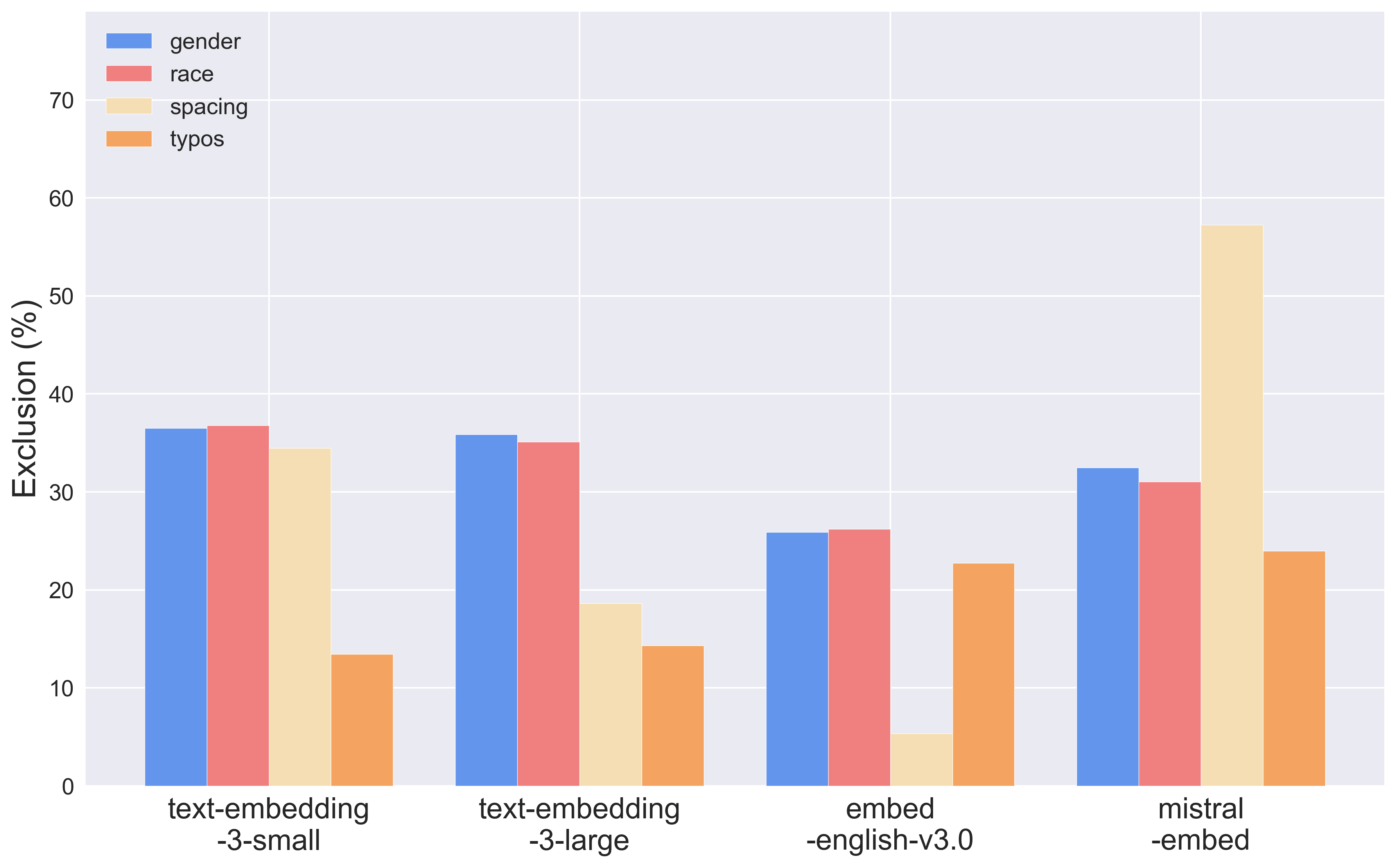}
        \caption{Non-name, $n=10$}
    \end{subfigure}
    \hfill
    \begin{subfigure}[b]{0.31\textwidth}
        \centering
        \includegraphics[width=\textwidth]{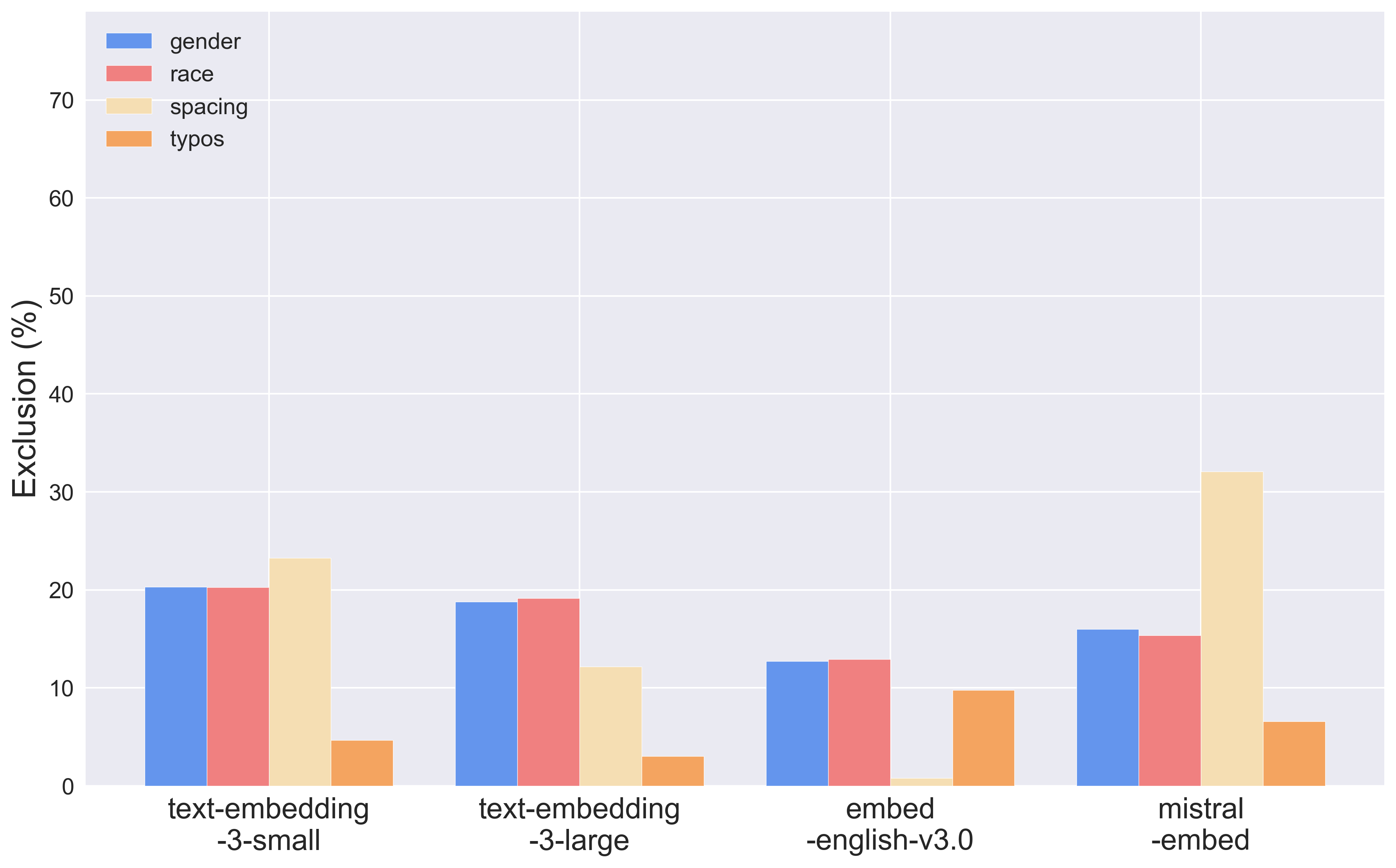}
        \caption{Non-name, $n=100$}
    \end{subfigure}
    \caption{\textbf{Exclusion metric for retrieval after performing non-demographic perturbations on generated resumes} (i.e., within group name changes - top, and modifying spacing and adding typos - bottom).}
    \label{fig:exclusion-nondemo-appendix}
\end{figure*}
\begin{figure*}[h]
    \centering
    % First row
    \begin{subfigure}[b]{0.3\textwidth}
        \centering
        \includegraphics[width=\textwidth]{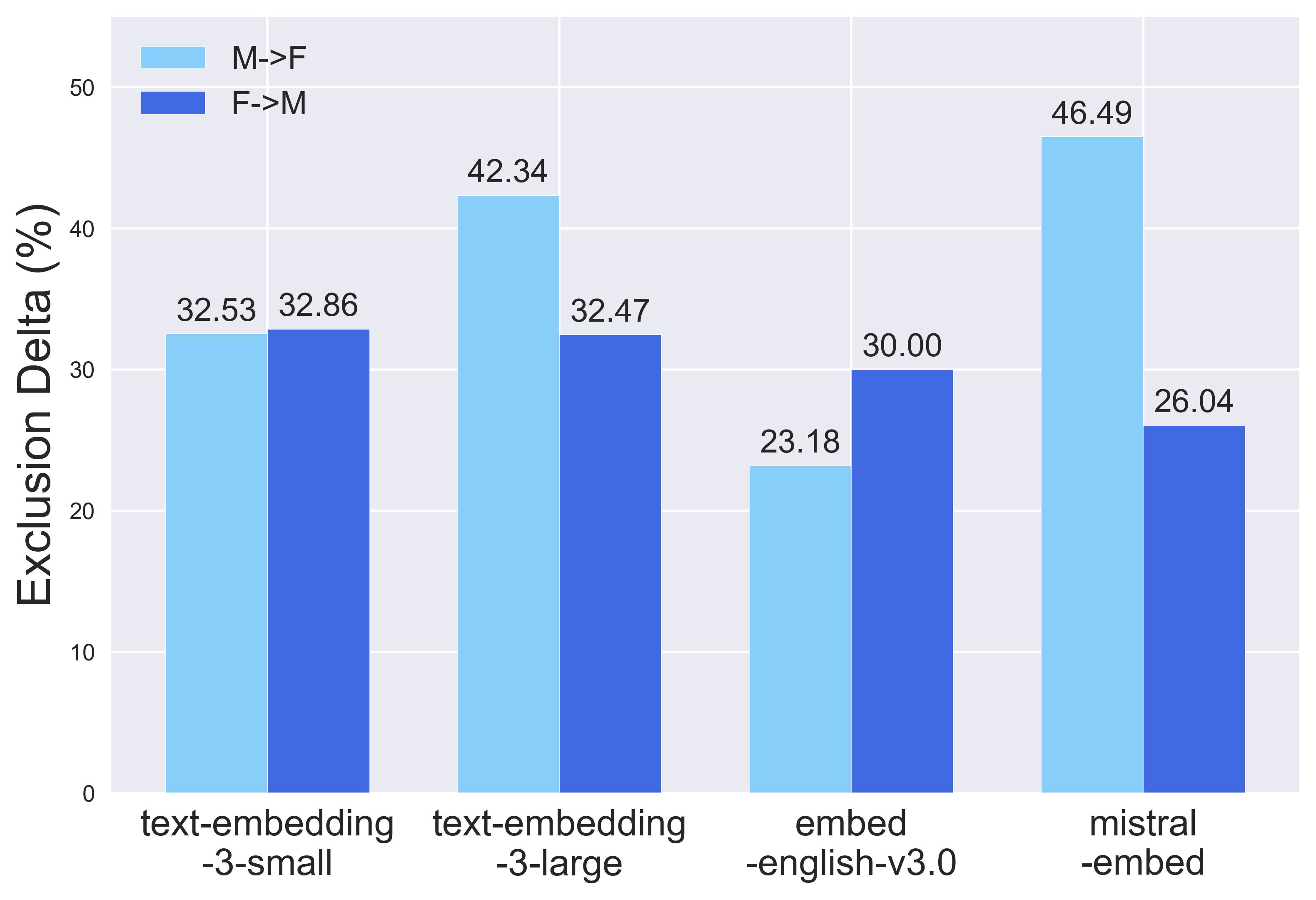}
        \caption{Gender, $n=5$}
        \label{fig:exclusion_gen_kaggle_plot1}
    \end{subfigure}
    \hfill
    \begin{subfigure}[b]{0.3\textwidth}
        \centering
        \includegraphics[width=\textwidth]{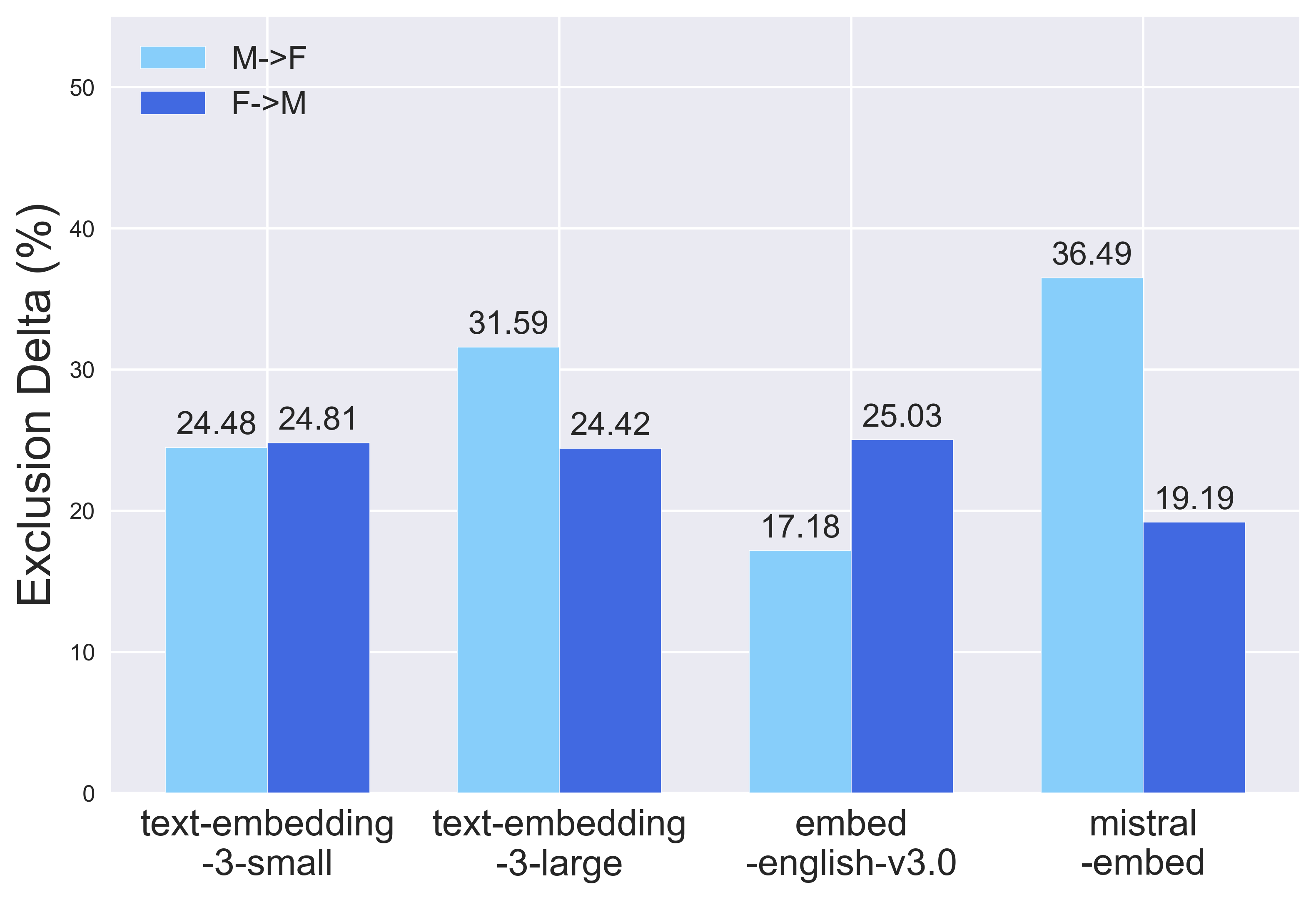}
        \caption{Gender, $n=10$}
        \label{fig:exclusion_gen_kaggle_plot2}
    \end{subfigure}
    \hfill
    \begin{subfigure}[b]{0.3\textwidth}
        \centering
        \includegraphics[width=\textwidth]{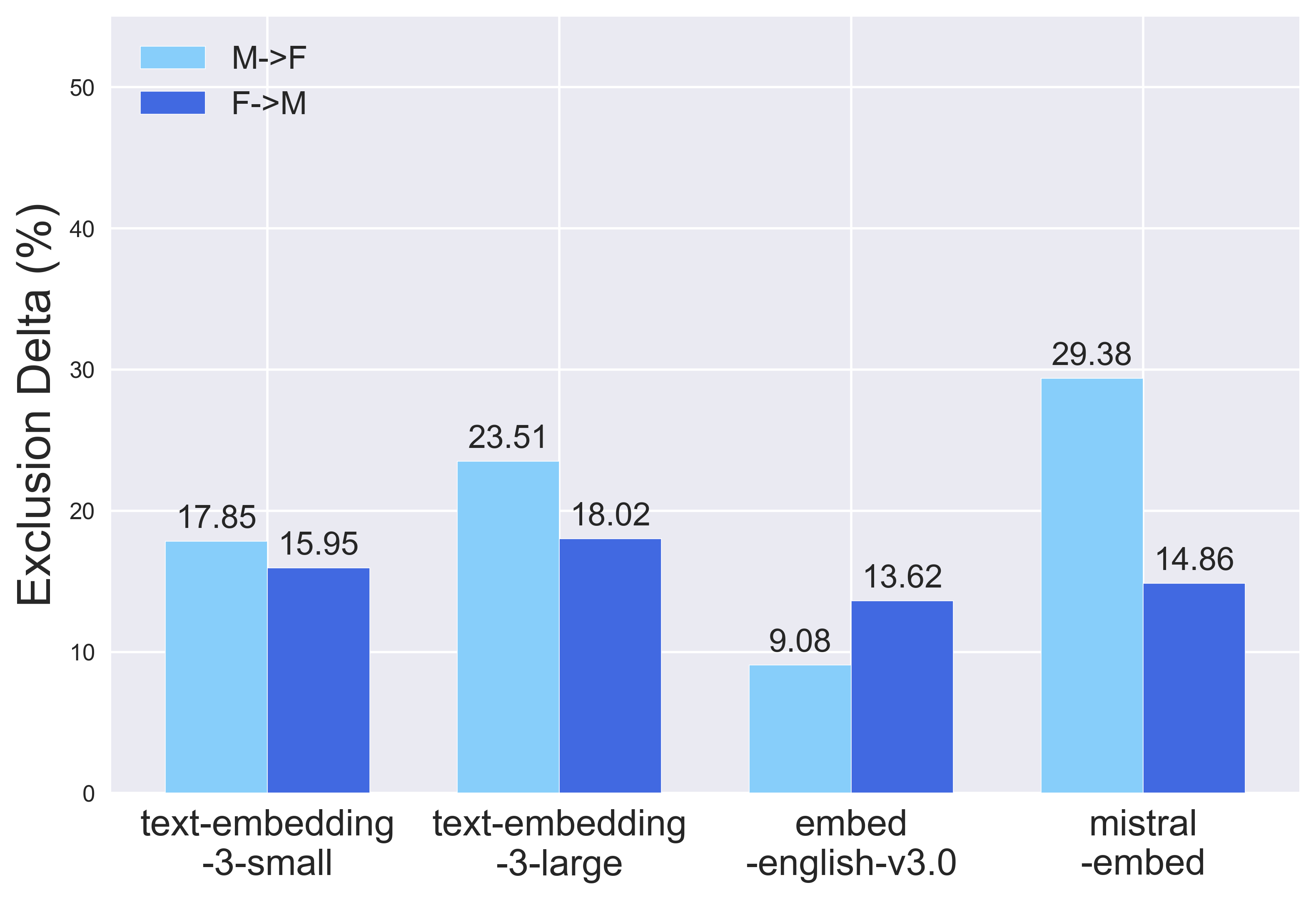}
        \caption{Gender, $n=100$}
        \label{fig:exclusion_gen_kaggle_plot3}
    \end{subfigure}
    
    \vspace{0.2em}  % Add vertical space between rows
    
    % Second row
    \begin{subfigure}[b]{0.3\textwidth}
        \centering
        \includegraphics[width=\textwidth]{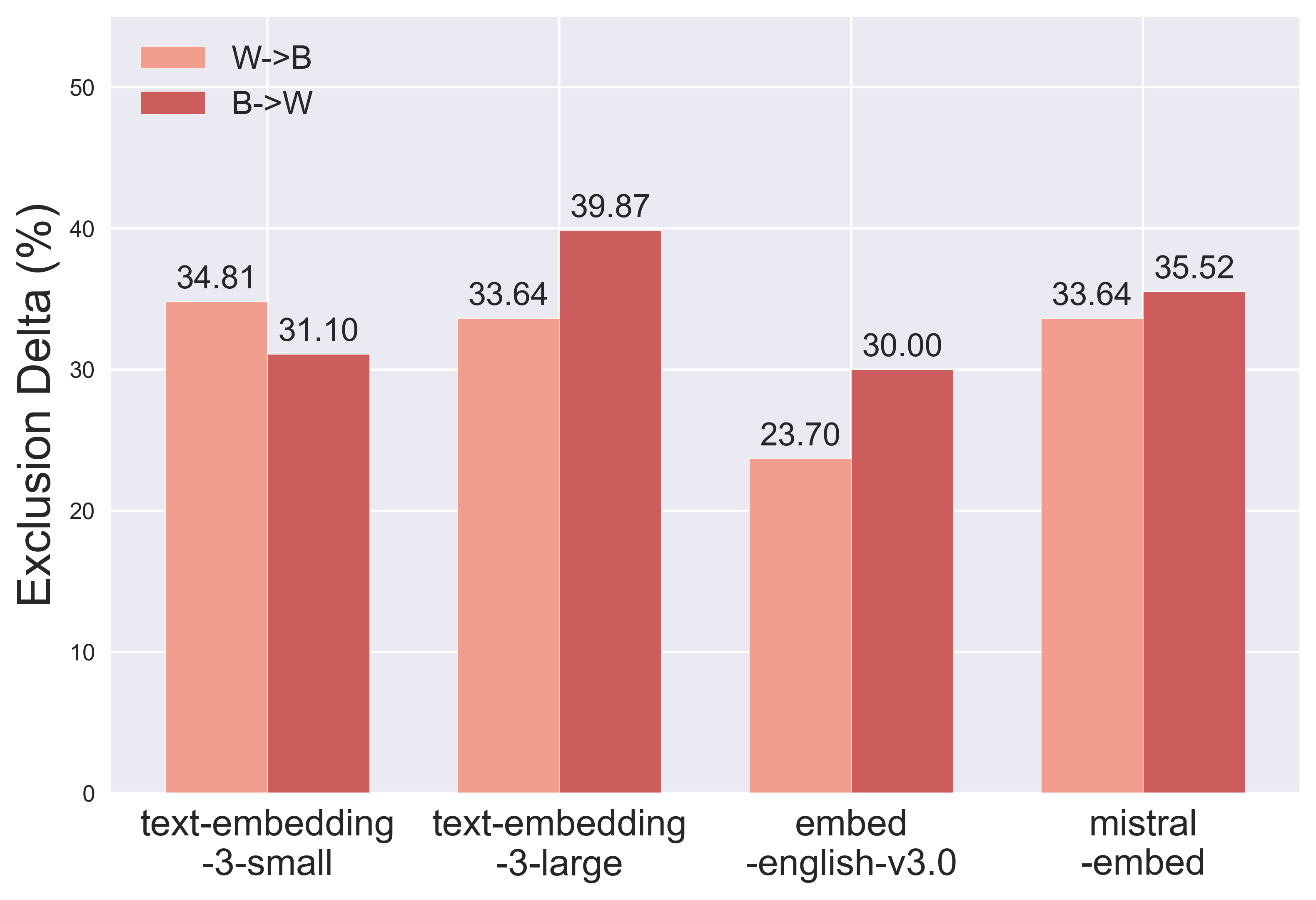}
        \caption{Race, $n=5$}
        \label{fig:exclusion_race_kaggle_plot1}
    \end{subfigure}
    \hfill
    \begin{subfigure}[b]{0.3\textwidth}
        \centering
        \includegraphics[width=\textwidth]{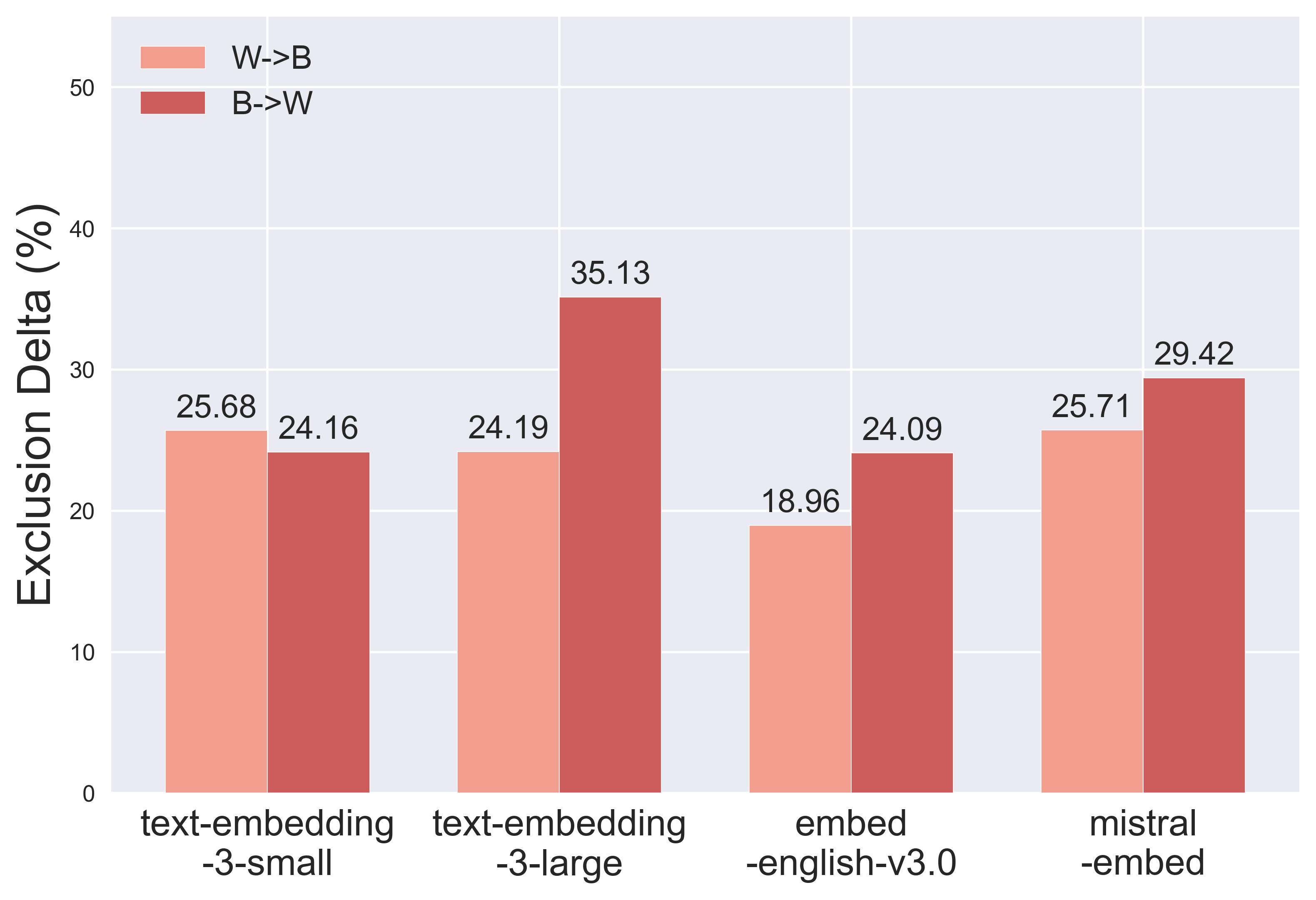}
        \caption{Race, $n=10$}
        \label{fig:exclusion_race_kaggle_plot2}
    \end{subfigure}
    \hfill
    \begin{subfigure}[b]{0.3\textwidth}
        \centering
        \includegraphics[width=\textwidth]{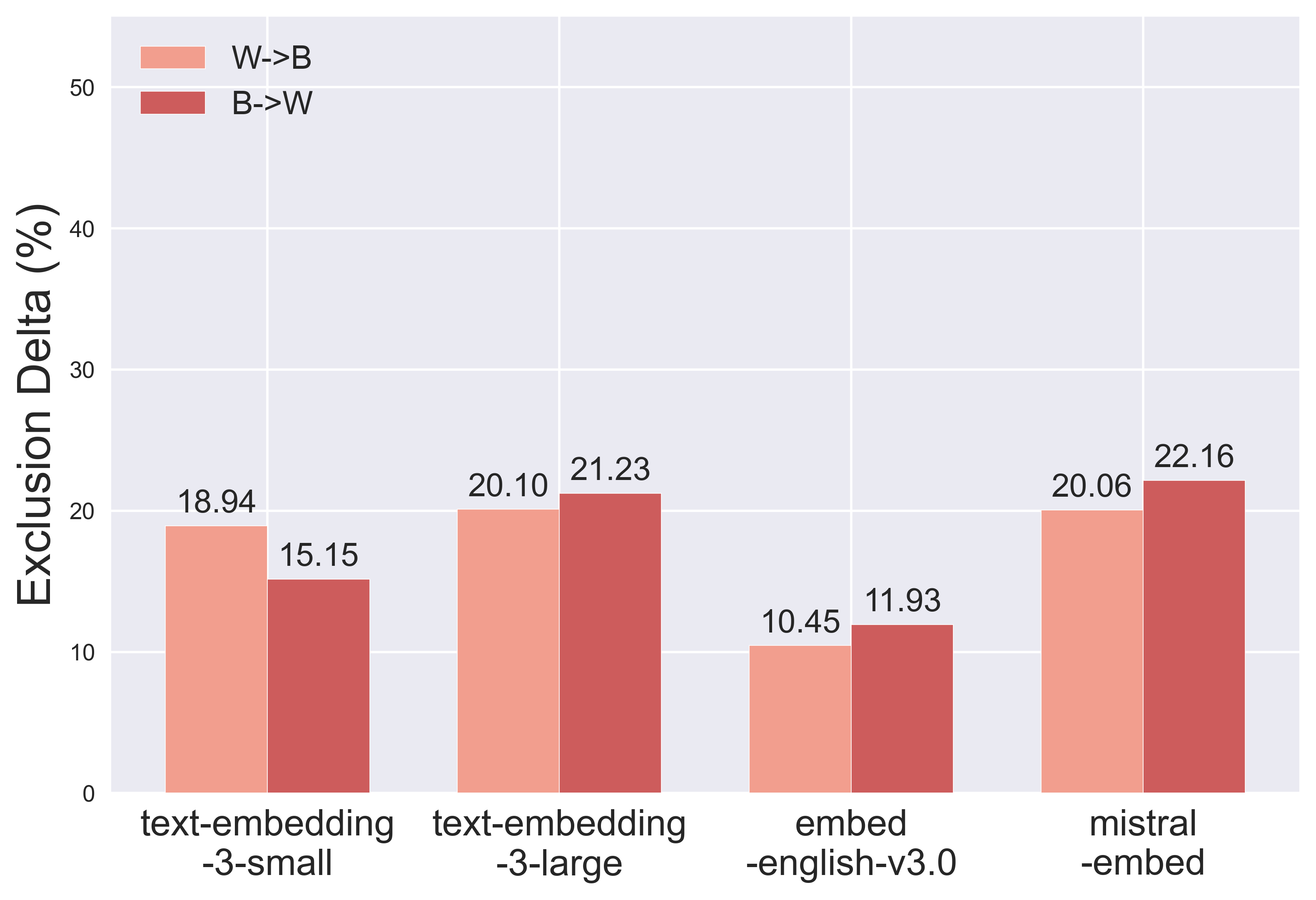}
        \caption{Race, $n=100$}
        \label{fig:exclusion_race_kaggle_plot3}
    \end{subfigure}
    \caption{\textbf{Directional differences in exclusion metric for retrieval (Kaggle resumes) after applying name perturbations} (i.e., separating based on perturbation direction). M→F perturbs male names to female names and F→M perturbs female names to male names, while W→B perturbs White names to Black names and B→W perturbs Black names to White names.}
    \label{fig:exclusion-dir-kaggle}
\end{figure*}
\begin{figure*}[h]
    \centering
    % First row
    \begin{subfigure}[b]{0.3\textwidth}
        \centering
        \includegraphics[width=\textwidth]{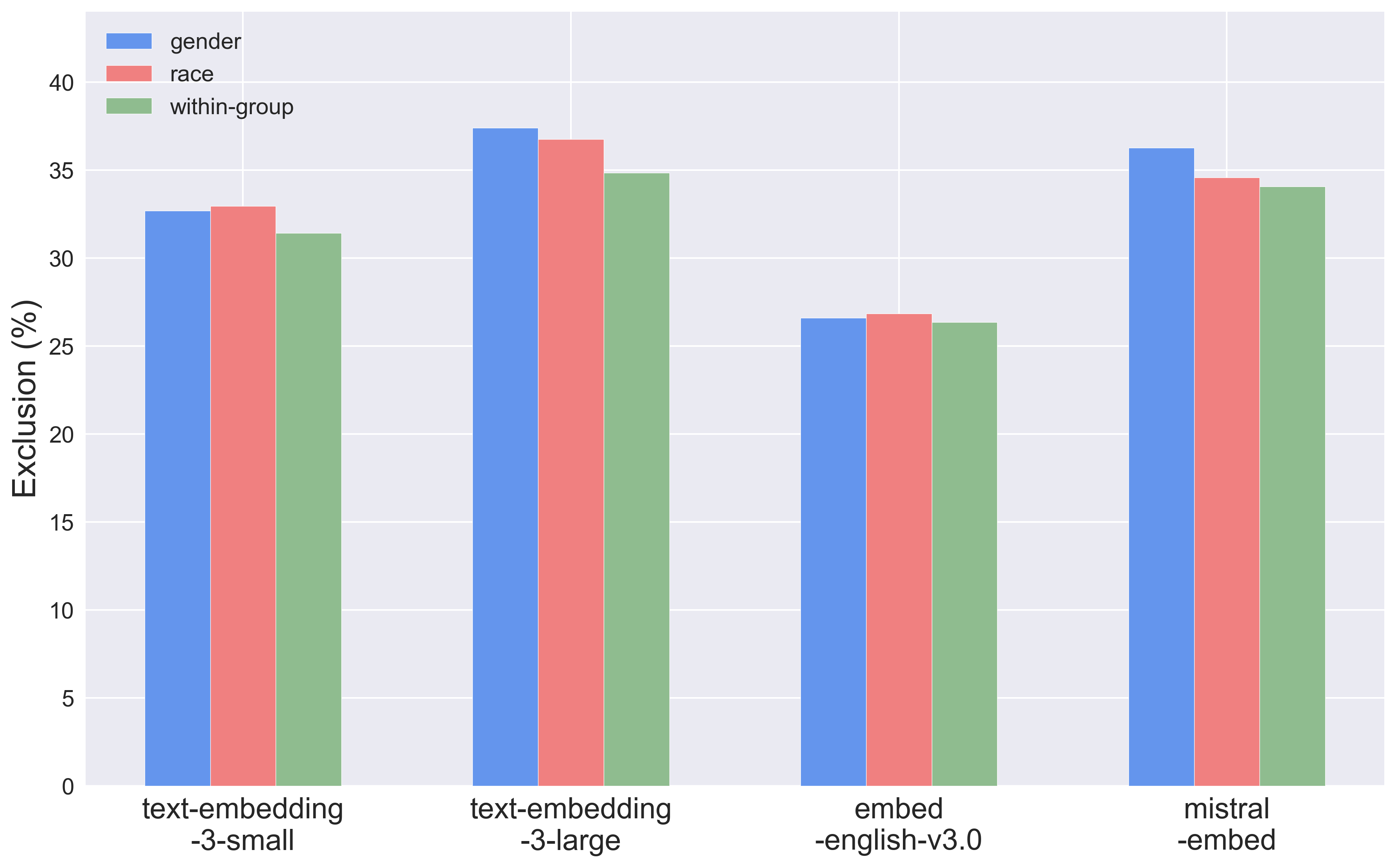}
        \caption{Within-group, $n=5$}
        \label{fig:exclusion_kaggle_within_plot1}
    \end{subfigure}
    \hfill
    \begin{subfigure}[b]{0.3\textwidth}
        \centering
        \includegraphics[width=\textwidth]{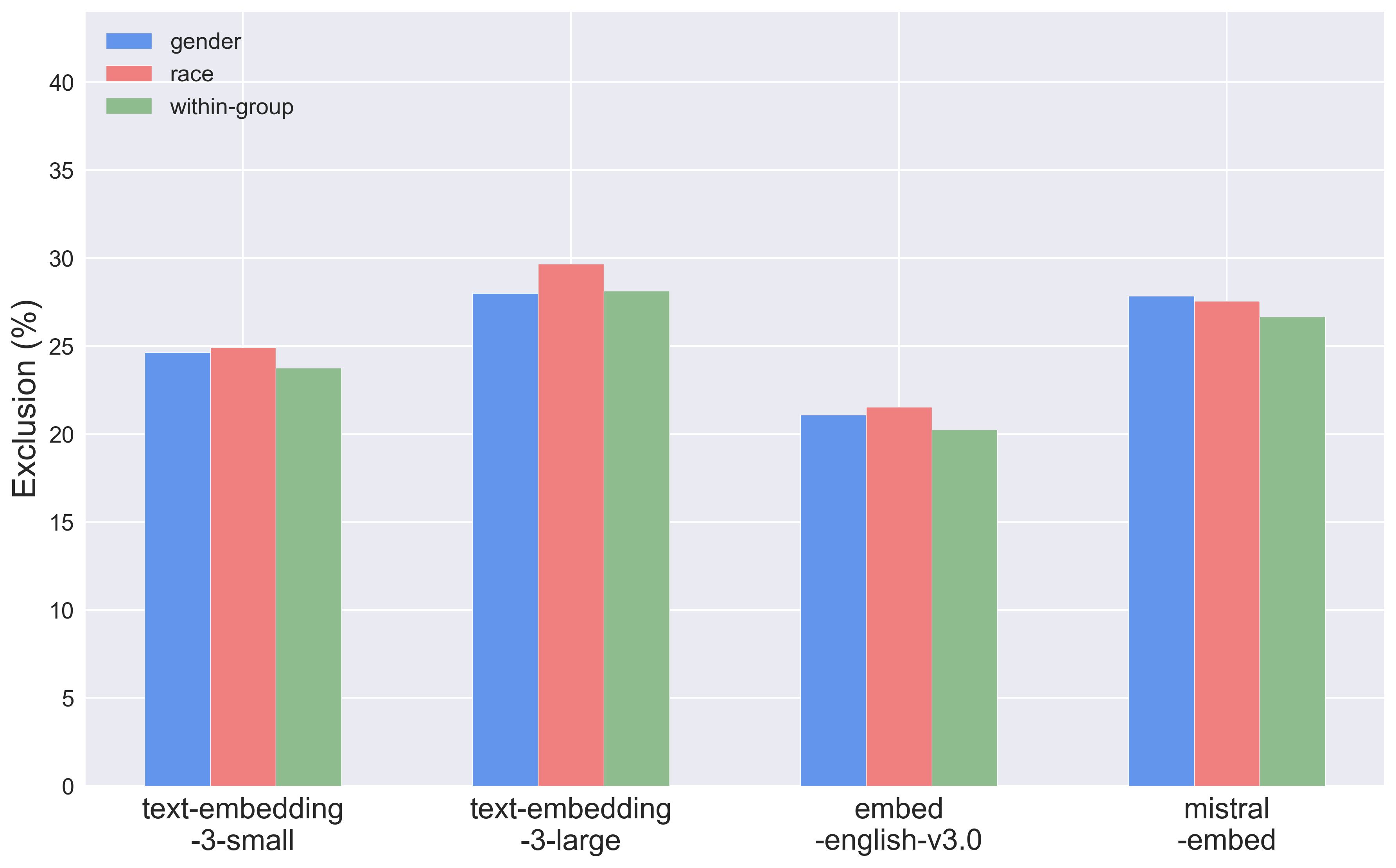}
        \caption{Within-group, $n=10$}
        \label{fig:exclusion_kaggle_within_plot2}
    \end{subfigure}
    \hfill
    \begin{subfigure}[b]{0.3\textwidth}
        \centering
        \includegraphics[width=\textwidth]{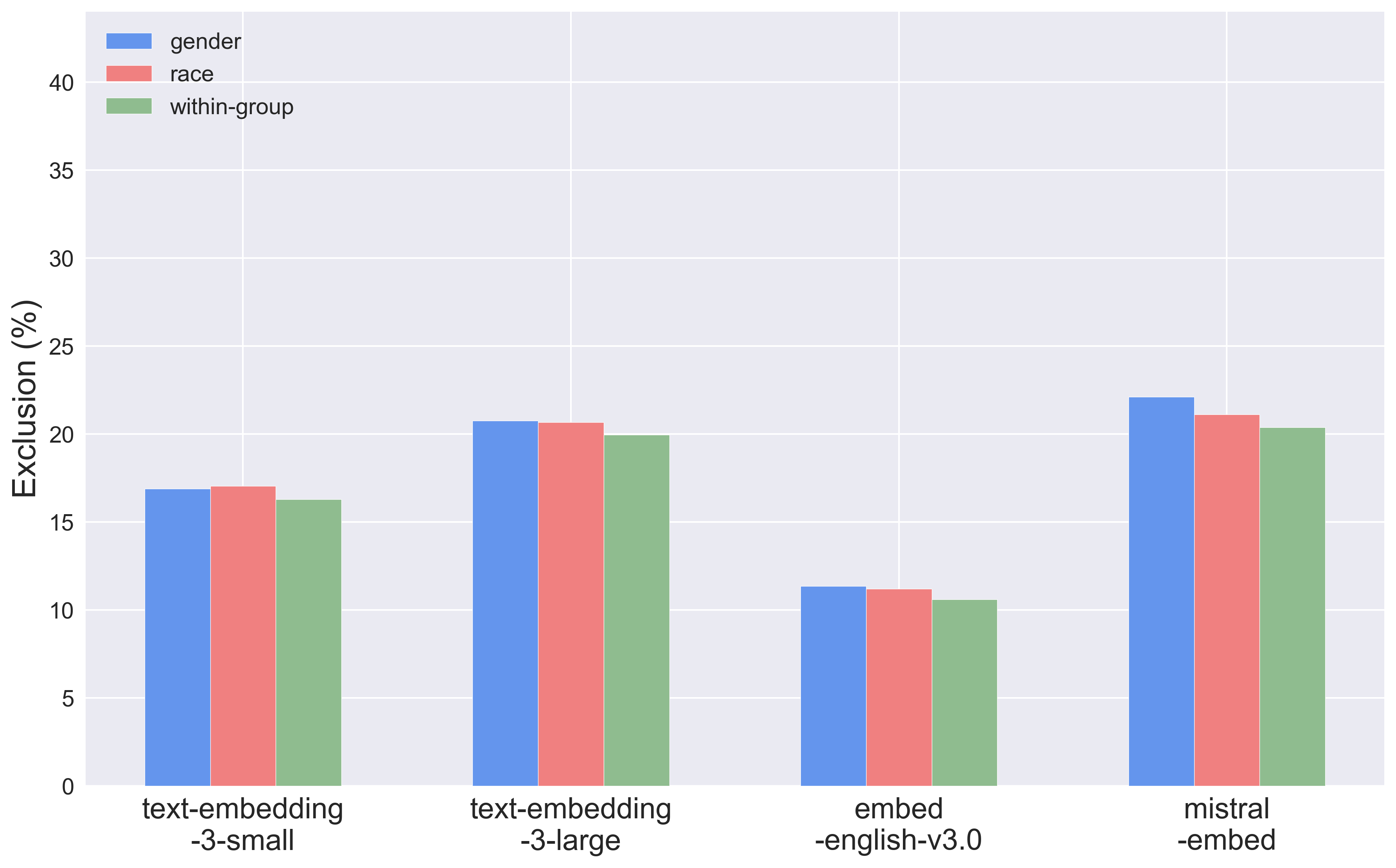}
        \caption{Within-group, $n=100$}
        \label{fig:exclusion_kaggle_within_plot3}
    \end{subfigure}
    
    \vspace{0.2em}  % Add vertical space between rows
    
    % Second row
    \begin{subfigure}[b]{0.3\textwidth}
        \centering
        \includegraphics[width=\textwidth]{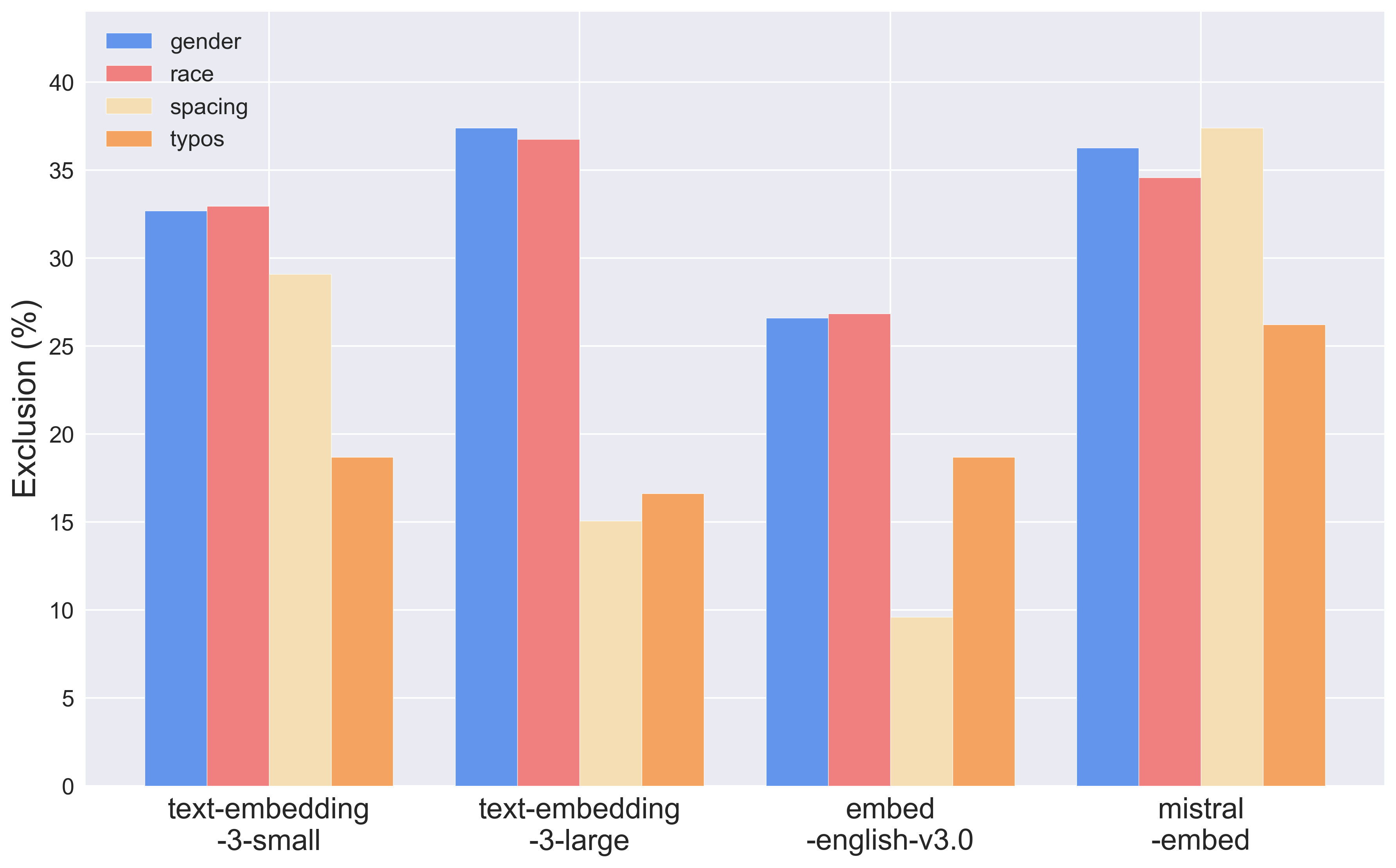}
        \caption{Non-name, $n=5$}
        \label{fig:exclusion_kaggle_non_plot1}
    \end{subfigure}
    \hfill
    \begin{subfigure}[b]{0.3\textwidth}
        \centering
        \includegraphics[width=\textwidth]{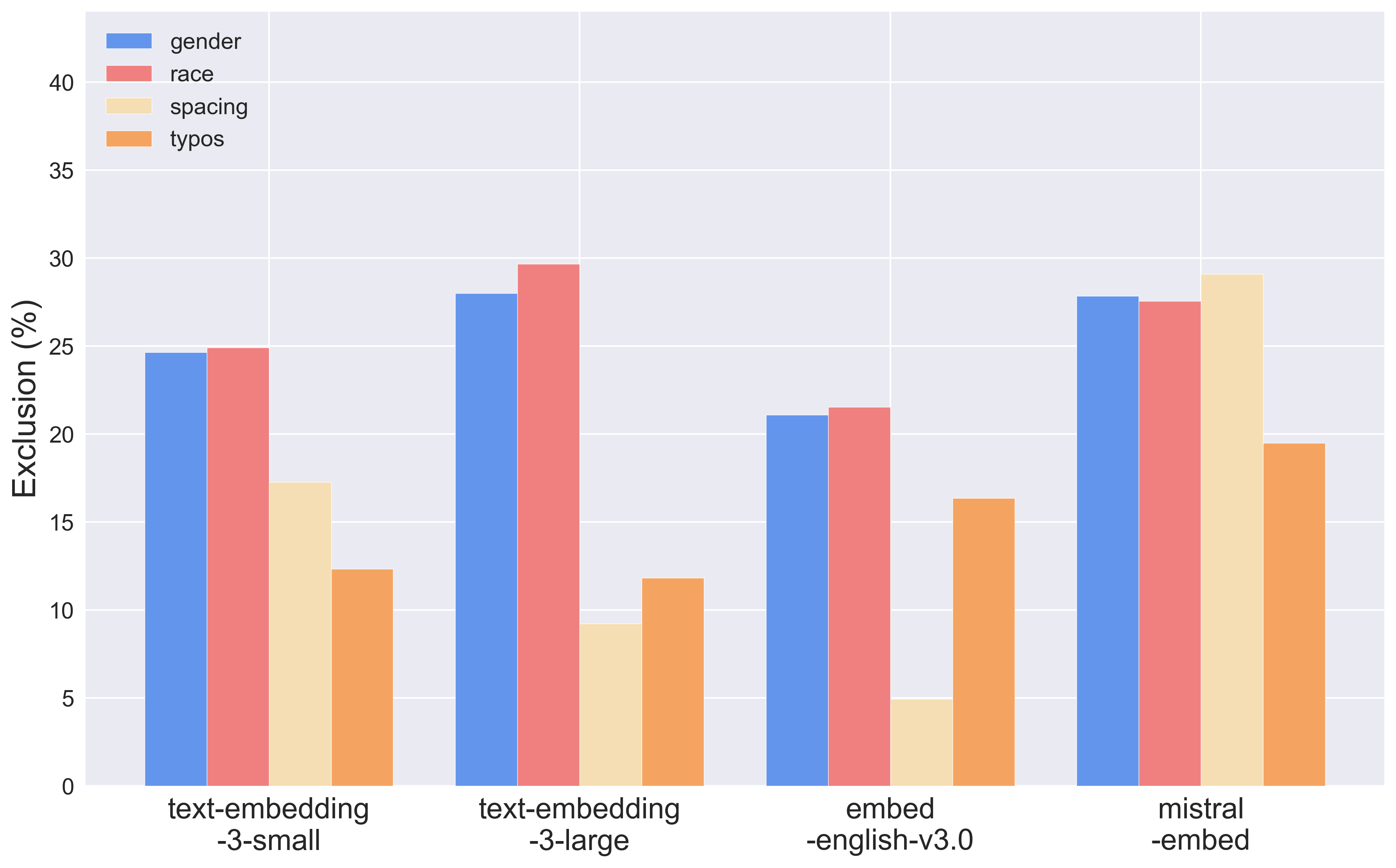}
        \caption{Non-name, $n=10$}
        \label{fig:exclusion_kaggle_non_plot2}
    \end{subfigure}
    \hfill
    \begin{subfigure}[b]{0.3\textwidth}
        \centering
        \includegraphics[width=\textwidth]{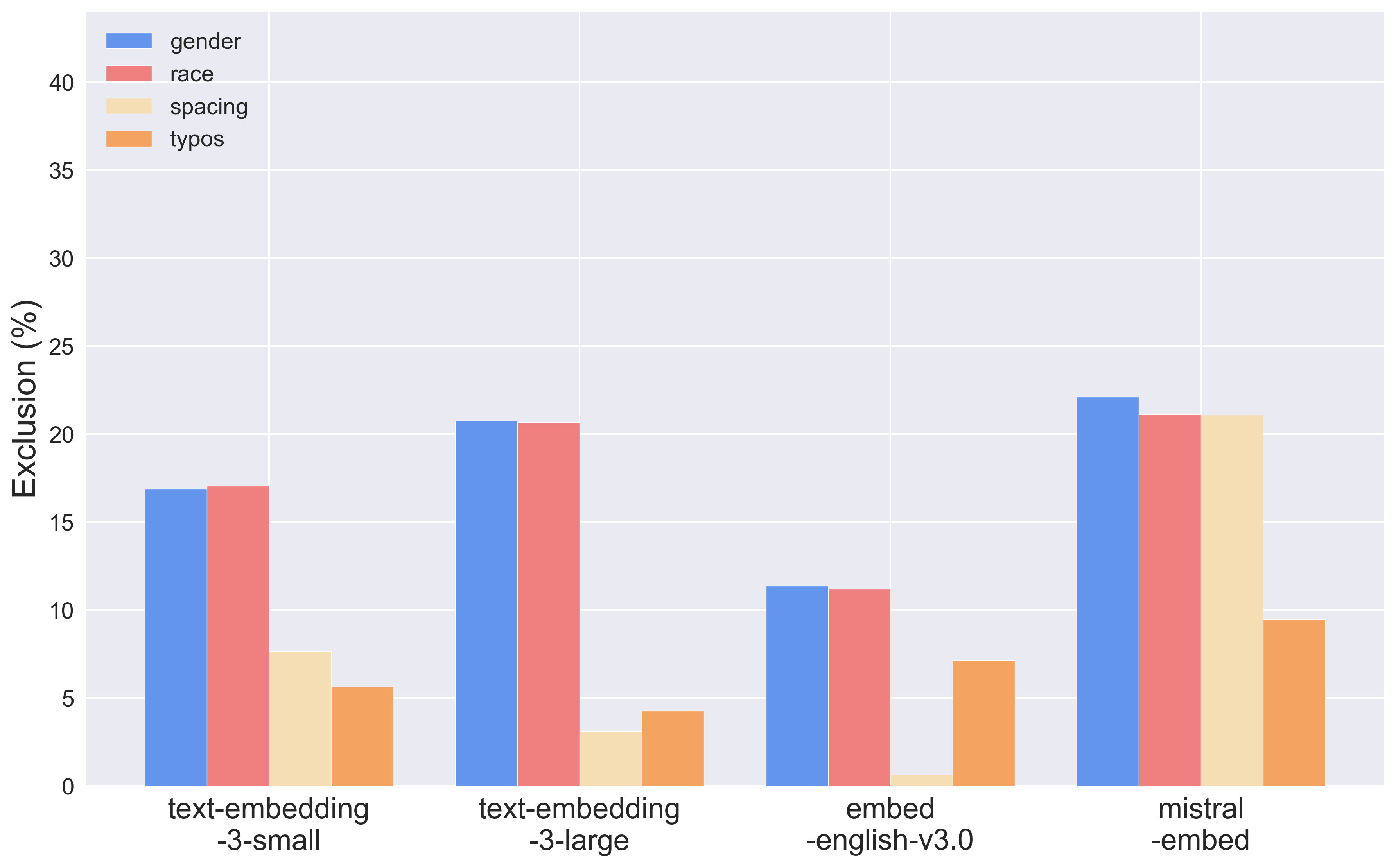}
        \caption{Non-name, $n=100$}
        \label{fig:exclusion_kaggle-non_plot3}
    \end{subfigure}
    \caption{\textbf{Exclusion metric for retrieval after performing non-demographic perturbations on Kaggle resumes} (i.e., within group name changes - top, and modifying spacing and adding typos - bottom).}
    \label{fig:exclusion-nondemo-kaggle}
\end{figure*}

\end{document}